\documentclass{article}

\usepackage{microtype}
\usepackage{graphicx}
\graphicspath{ {./figures/} }
\usepackage{booktabs} 
\usepackage{subcaption}
\usepackage{makecell}

\usepackage[dvipsnames]{xcolor}
\usepackage{soul}
 
\usepackage{hyperref}

\usepackage[accepted]{icml2023}

\usepackage{amsmath}
\usepackage{amssymb}
\usepackage{mathtools}
\usepackage{amsthm}

\usepackage[capitalize,noabbrev]{cleveref}

\theoremstyle{plain}

\theoremstyle{definition}

\theoremstyle{remark}

\usepackage[textsize=tiny]{todonotes}

\icmltitlerunning{Trompt: Towards a Better Deep Neural Network for Tabular Data}

\begin{document}

\twocolumn[
\icmltitle{Trompt: Towards a Better Deep Neural Network for Tabular Data}




\begin{icmlauthorlist}
\icmlauthor{Kuan-Yu Chen}{comp}
\icmlauthor{Ping-Han Chiang}{comp}
\icmlauthor{Hsin-Rung Chou}{comp}
\icmlauthor{Ting-Wei Chen}{comp}
\icmlauthor{Darby Tien-Hao Chang}{comp,yyy}
\end{icmlauthorlist}

\icmlaffiliation{yyy}{Department of Electronic Engineering, National Cheng Kung University, Tainan, Taiwan}
\icmlaffiliation{comp}{SinoPac Holdings, Taipei, Taiwan}

\icmlcorrespondingauthor{Kuan-Yu Chen}{lavamore@sinopac.com}

\icmlkeywords{Tabular Data, Prompt Learning, Model Architecture Design}

\vskip 0.3in
]



\printAffiliationsAndNotice{} 

\begin{abstract}
Tabular data is arguably one of the most commonly used data structures in various practical domains, including finance, healthcare and e-commerce.
However, based on a recently published tabular benchmark, we can see deep neural networks still fall behind tree-based models on tabular datasets \cite{grinsztajn2022why}.
In this paper, we propose \emph{Trompt}--which stands for \textbf{T}abular P\textbf{rompt}--a novel architecture inspired by prompt learning of language models.
The essence of prompt learning is to adjust a large pre-trained model through a set of prompts outside the model without directly modifying the model.
Based on this idea, Trompt separates the learning strategy of tabular data into two parts for the intrinsic information of a table and the varied information among samples.
Trompt is evaluated with the benchmark mentioned above.
The experimental results demonstrate that Trompt outperforms state-of-the-art deep neural networks and is comparable to tree-based models (\cref{fig:overall-eval}).
\end{abstract}

\begin{figure}[t]
    \centering
    \begin{subfigure}{.5\columnwidth}
        \centering
        \includegraphics[width=.95\linewidth]{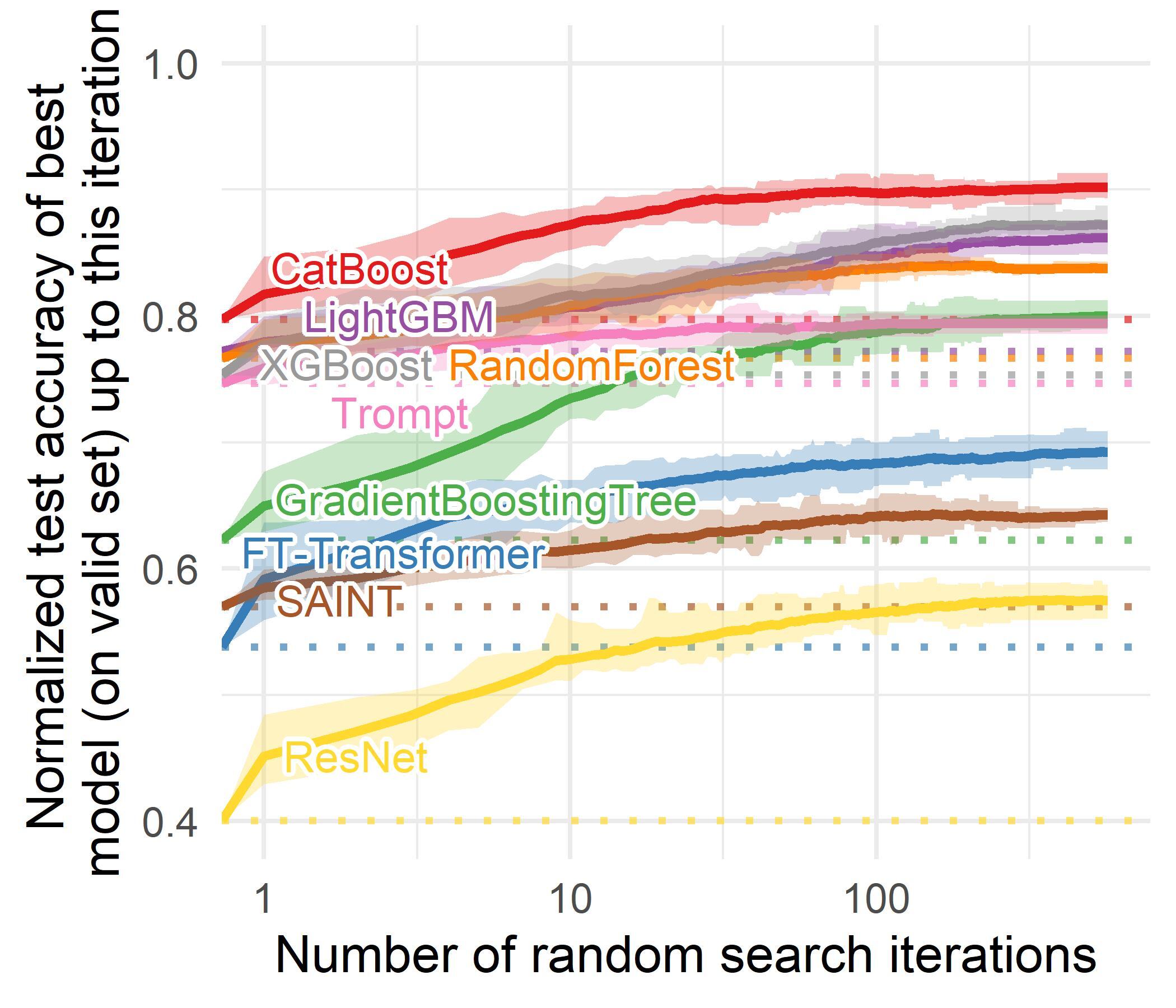}
        \caption{Medium-sized classification task.}
    \end{subfigure}%
    \begin{subfigure}{.5\columnwidth}
        \centering
        \includegraphics[width=.95\linewidth]{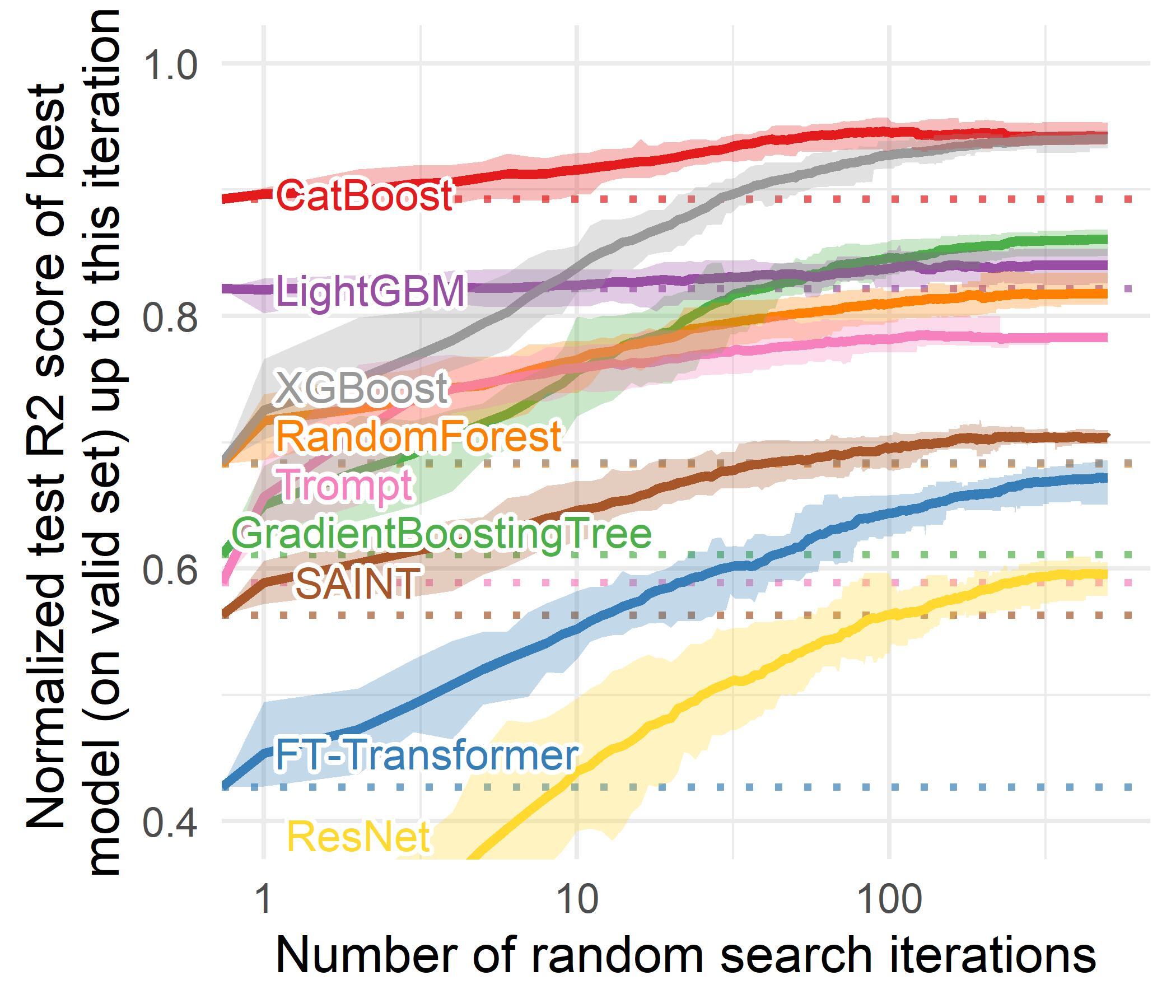}
        \caption{Medium-sized regression task.}
    \end{subfigure}
    \begin{subfigure}{.5\columnwidth}
        \centering
        \includegraphics[width=.95\linewidth]{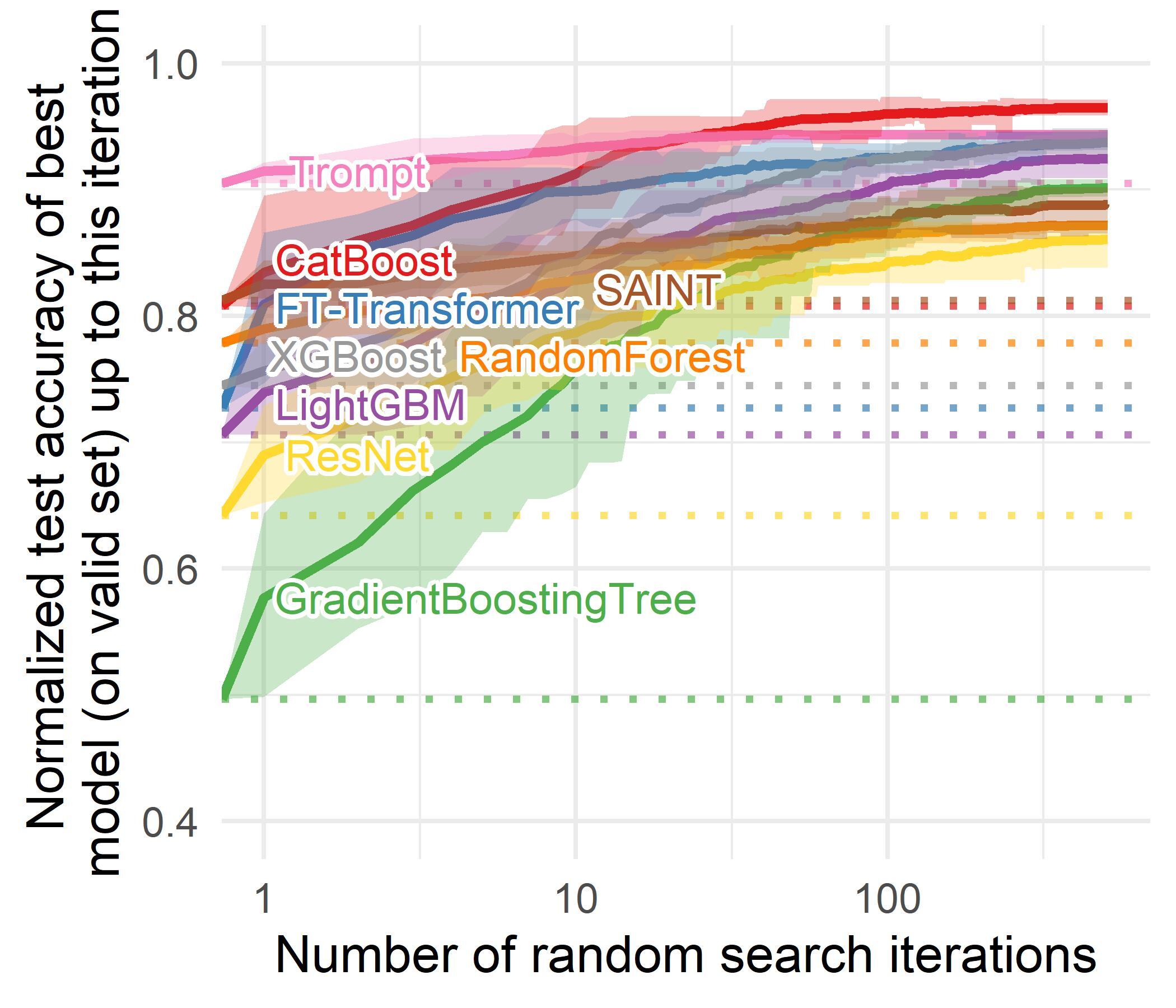}
        \caption{Large-sized classification task.}
    \end{subfigure}%
    \begin{subfigure}{.5\columnwidth}
        \centering
        \includegraphics[width=.95\linewidth]{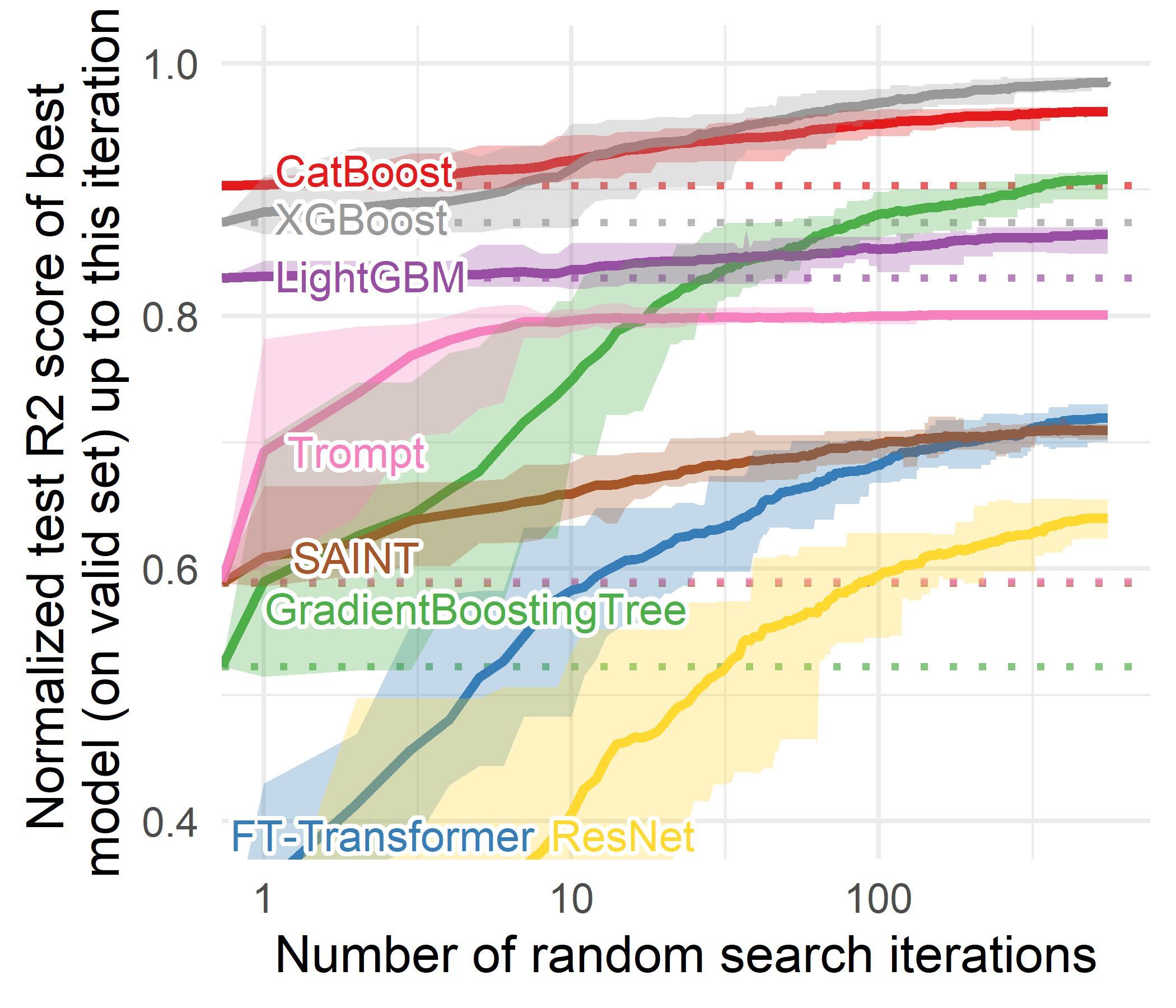}
        \caption{Large-sized regression task.}
    \end{subfigure}
    \caption{Benchmark results.}
    \label{fig:overall-eval}
\end{figure}

\section{Introduction}
\label{submission}
Tabular data plays a vital role in many real world applications, such as financial statements for banks to evaluate the credibility of a company, diagnostic reports for doctors to identify the aetiology of a patient, and customer records for e-commerce platforms to discover the potential interest of a customer.
In general, tabular data can be used to record activities consisting of heterogeneous features and has many practical usages.

On the other hand, deep learning has achieved a great success in various domains, including computer vision, natural language processing (NLP) and robotics \cite{he2016deep,redmon2016you,gu2017deep,devlin2018bert}.
Besides extraordinary performance, there are numerous benefits of the end-to-end optimization nature of deep learning, including (i) online learning with streaming data \cite{sahoo2017online}, (ii) multi-model integration that incorporates different types of input, e.g., image and text \cite{ramachandram2017deep} and (iii) representation learning that realizes semi-supervised learning and generative modeling \cite{van2020survey,goodfellow2020generative}.

Consequently, researchers have been dedicated to apply deep learning on tabular data, either through (i) transformer \cite{huang2020tabtransformer,somepalli2021saint,gorishniy2021revisiting} or (ii) inductive bias investigation \cite{katzir2020net,arik2021tabnet}. 

Though many of the previous publications claimed that they have achieved the state of the art, further researches pointed that previous works were evaluated on favorable datasets and tree-based models still show superior performances in the realm of tabular data \cite{borisov2021deep,gorishniy2021revisiting,SHWARTZZIV202284}.
For a fair comparison between different algorithms, a standard benchmark for tabular data was proposed by \cite{grinsztajn2022why}.
The benchmark, denoted as \emph{Grinsztajn45} in this work, consists of 45 curated datasets from various domains.

In this paper, we propose a novel prompt-inspired architecture, \emph{Trompt}, which abbreviates \textbf{T}abular P\textbf{rompt}.
Prompt learning has played an important role in the recent development of language models.
For example, GPT-3 can well handle a wide range of tasks with an appropriate prompt engineering \cite{radford2018improving,brown2020language}.
In Trompt, prompt is utilized to derive feature importances that vary in different samples.
Trompt consists of multiple \emph{Trompt Cell}s and a shared \emph{Trompt Downstream} as \cref{fig:overall}.
Each Trompt Cell is responsible for feature extraction, while the Trompt Downstream is for prediction.

The performance of Trompt is evaluated on the Grinsztajn45 benchmark and compared with three deep learning models and five tree-based models.
\cref{fig:overall-eval} illustrates the overall evaluation results on Grinsztajn45.
The x-axis is the number of hyperparameter search iterations and y-axis is the normalized performance.
In \cref{fig:overall-eval}, Trompt is consistently better than state-of-the-art deep learning models (SAINT and FT-Transformer) and the gap between deep learning models and tree-based models is narrowed.

Our key contributions are summarized as follows:
\begin{itemize}
\item The experiments are conducted on a recognized tabular benchmark, Grinsztajn45.
Additionally, we add two well-performed tree-based models, LightGBM \cite{ke2017lightgbm} and CatBoost \cite{prokhorenkova2018catboost} to baselines.
\item Trompt achieves state-of-the-art performance among deep learning models and narrows the performance gap between deep learning models and tree-based models.
\item Thorough empirical studies and ablation tests were conducted to verify the design of Trompt.
The results further shed light on future research directions of the architecture design of tabular neural network.
\end{itemize}

\section{Related Work}
In this section, we first discuss the prompt learning of language models.
Secondly, we discuss two research branches of tabular neural networks, transformer and inductive bias investigation.
Lastly, we discuss the differences between Trompt and the related works and highlight the uniqueness of our work.

\subsection{Prompt Learning}
The purpose of prompt learning is to transform the input and output of downstream tasks to the original task used to build a pre-trained model.
Unlike fine-tuning that changes the task and usually involves updating model weights, a pre-train model with prompts can dedicate itself to one task.
With prompt learning, a small amount of data or even zero-shot can achieve good results \cite{radford2018improving,brown2020language}.
The emergence of prompt learning substantially improves the application versatility of pre-trained models that are too large for common users to fine-tune.

To prompt a language model, one can insert a task-specific prompt before a sentence and hint the model to adjust its responses for different tasks \cite{brown2020language}.
Prompts can either be discrete or soft.
The former are composed of discrete tokens from the vocabulary of natural languages \cite{radford2018improving,brown2020language}, while the latter are learned representations \cite{li2021prefix,lester2021power}.

\subsection{Tabular Neural Network}
\label{sec:tabular-nn}
\textbf{Transformer.} Self-attention has revolutionized NLP since 2017 \cite{vaswani2017attention}, and soon been adopted by other domains, such as computer vision, reinforcement learning and speech recognition \cite{dosovitskiy2020image,chen2021decision,zhang2020transformer}.
The intention of transformer blocks is to capture the relationships among features, which can be applied on tabular data as well.

TabTransformer \cite{huang2020tabtransformer} is the first transformer-based tabular neural network.
However, TabTransformer only fed categorical features to transformer blocks and ignored the potential relationships among categorical and numerical features.
FT-Transformer \cite{gorishniy2021revisiting} fixed this issue through feeding both categorical and numerical features to transformer blocks.
SAINT \cite{somepalli2021saint} further improved FT-Transformer through applying attentions on not only the feature dimensions but also the sample dimensions.

\begin{figure*}[t]
    \centering
    \includegraphics[width=0.7\textwidth]{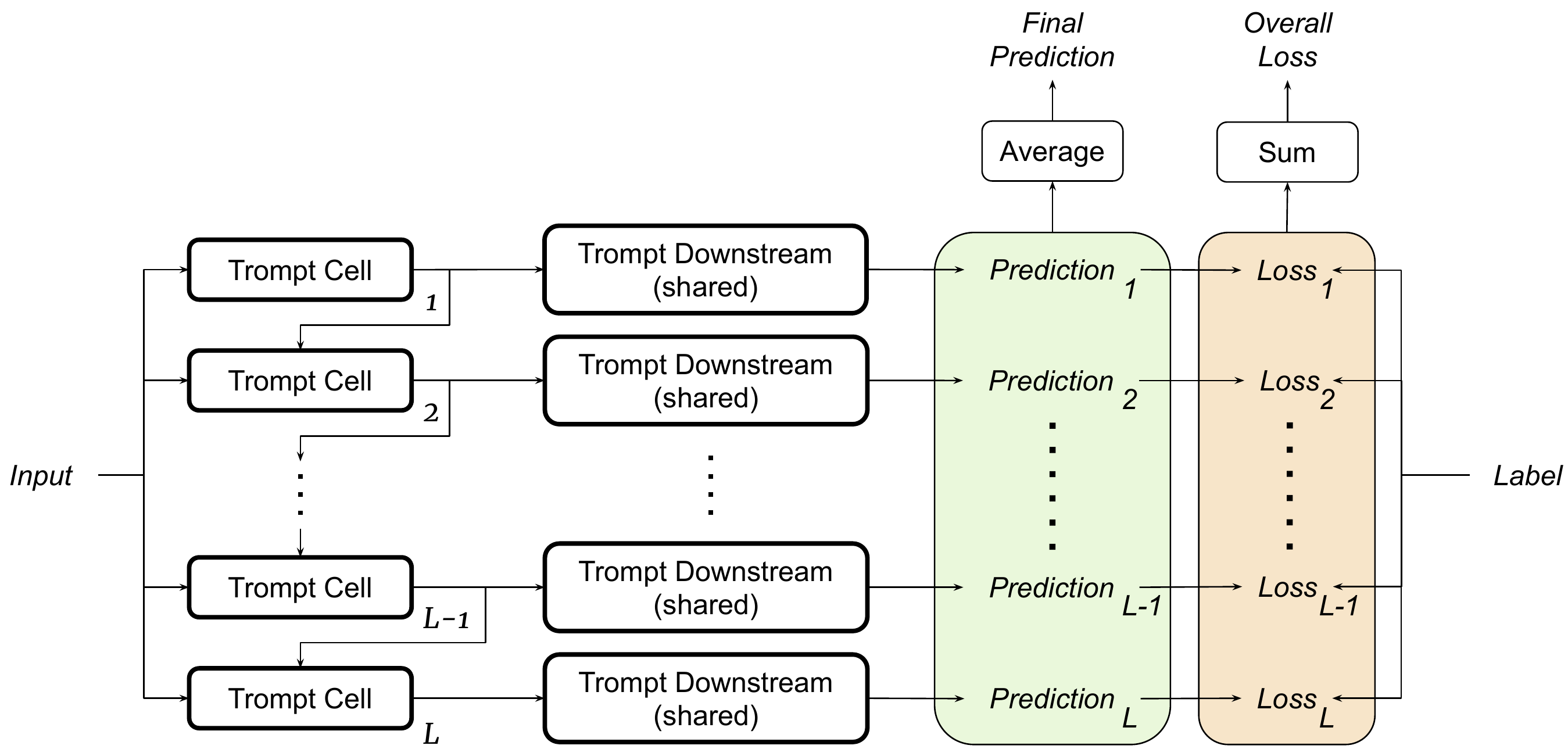}
    \caption{Overall architecture of the proposed Trompt.}
    \label{fig:overall}
\end{figure*}

\textbf{Inductive Bias Investigation.} Deep neural networks perform well on tasks with clear inductive bias.
For example, Convolutional Neural Network (CNN) works well on images. The kernel of CNN is designed to capture local patterns since neighboring pixels usually relate to each other \cite{lecun1995convolutional}.
Recurrent Neural Networks (RNN) is widely used in language understanding because the causal relationship among words is well encapsulated through recurrent units \cite{rumelhart1986learning}.
However, unlike other popular tasks, the inductive bias of tabular data has not been well discovered.

Given the fact that tree-based model has been the solid state of the art for tabular data \cite{borisov2021deep,gorishniy2021revisiting,SHWARTZZIV202284}, Net-DNF \cite{katzir2020net} and TabNet \cite{arik2021tabnet} hypothesized that the inductive bias for tabular data might be the learning strategy of tree-based model.
The strategy is to find the optimal root-to-leaf decision paths by selecting a portion of the features and deriving the optimal split from the selected features in non-leaf nodes.
To emulate the learning strategy, TabNet utilized sequential attention and sparsity regularization.
On the other hand, Net-DNF theoretically proved that decision tree is equivalent to some disjunctive normal form (DNF) and proposed disjunctive neural normal form to emulate a DNF formula.

\subsection{The Uniqueness of Trompt}
We argue that the column importances of tabular data are not invariant for all samples and can be grouped into multiple modalities.
Since prompt learning is born to adapt a model to multiple tasks, the concept is used in Trompt to handle multiple modalities.
To this end, Trompt separates the learning strategy of tabular data into two parts.
The first part, analogous to pre-trained models, focus on learning the intrinsic column information of a table.
The second part, analogous to prompts, focus on diversifying the feature importances of different samples.

As far as our understanding, Trompt is the first prompt-inspired tabular neural network.
Compared to transformer-based models, Trompt learns separated column importances instead of focusing on the interactions among columns.
Compared to TabNet and Net-DNF, Trompt handle multiple modalities by emulating prompt learning instead of the branch split of decision tree.






\begin{figure*}[t]
    \centering
    \includegraphics[width=.95\textwidth]{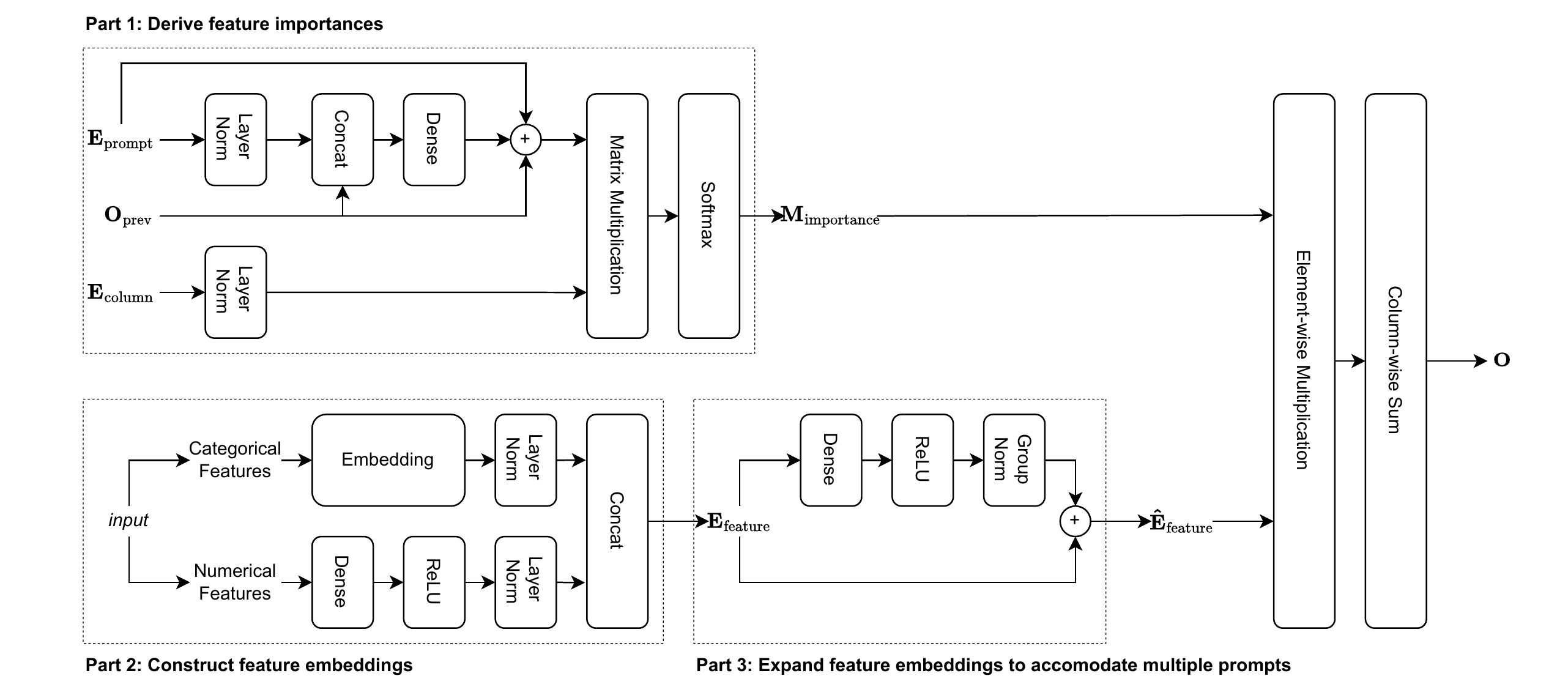}
    \caption{Architecture of a Trompt Cell.}
    \label{fig:cell}
\end{figure*}

\section{Trompt}
In this section, we elaborate on the architecture design of Trompt.
As \cref{fig:overall} shows, Trompt consists of multiple Trompt Cells and a shared Trompt Downstream.
Each Trompt Cell is responsible for feature extraction and providing diverse representations, while the Trompt Downstream is for prediction.
The details of Trompt Cell and Trompt Downstream are discussed in \cref{sec:cell} and \cref{sec:downstream}, respectively.
In \cref{sec:discuss}, we further discuss the prompt learning of Trompt.

\subsection{Trompt Cell}
\label{sec:cell}
\cref{fig:cell} illustrates the architecture of a Trompt Cell, which can be divided into three parts.
The first part derives feature importances ($\mathbf{M}_\text{importance}$) based on column embeddings ($\mathbf{E}_\text{column}$), the previous cell's output ($\mathbf{O}_\text{prev}$) and prompt embeddings ($\mathbf{E}_\text{prompt}$).
The second part transforms the input into feature embeddings ($\mathbf{E}_\text{feature}$) with two paths for categorical and numerical columns, respectively.
The third part expands $\mathbf{E}_\text{feature}$ for the later multiplication.

The details of the first part are illustrated in \cref{sec:feature-importances} and the details of the second and third parts are illustrated in \cref{sec:input-transform}.
Lastly, the generation of the output of a Trompt Cell is illustrated in \cref{sec:cell-output}.

\subsubsection{Derive Feature Importances}
\label{sec:feature-importances}
Let $\mathbf{E}_\text{column}\in\mathbb{R}^{C\times{d}}$ be column embeddings and $\mathbf{E}_\text{prompt}\in\mathbb{R}^{P\times{d}}$ be prompt embeddings.
$C$ is the number of columns of a table defined by the dataset, while $P$ and $d$ are hyperparameters for the number of prompts and the hidden dimension, respectively.
Both $\mathbf{E}_\text{column}$ and $\mathbf{E}_\text{prompt}$ are input independent and trainable.
Let $\mathbf{O}_\text{prev}\in\mathbb{R}^{B\times{P}\times{d}}$ be the previous cell's output and $B$ be the batch size.

$\mathbf{O}_\text{prev}$ is fused with the prompt embeddings as \cref{eq:prompt-stack,eq:prompt-concat}.
Since $\mathbf{E}_\text{prompt}$ is input independent and lack a batch dimension, $\mathbf{E}_\text{prompt}$ is expanded to $\mathbf{SE}_\text{prompt}$ through the stack operation as \cref{eq:prompt-stack}.
Later, we concatenate $\mathbf{SE}_\text{prompt}$ and $\mathbf{O}_\text{prev}$ and then reduce the dimension of the concatenated tensor back to $\mathbb{R}^{B\times{P}\times{d}}$ for the final addition as \cref{eq:prompt-concat}.

For the same reason as $\mathbf{E}_\text{prompt}$, the $\mathbf{E}_\text{column}$ is expanded to $\mathbf{SE}_\text{column}$ as \cref{eq:column-stack}.
Subsequently, feature importances are derived through \cref{eq:soft-select}, where $\otimes$ is the batch matrix multiplication, $\intercal$ is the batch transpose, and the $\mathtt{softmax}$ is applied to the column axis.

\begin{equation}
\label{eq:prompt-stack}
    \mathbf{SE}_\text{prompt}=\mathtt{stack}(\mathbf{E}_\text{prompt})\in\mathbb{R}^{B\times{P}\times{d}}
\end{equation}
\begin{equation}
\label{eq:prompt-concat}
    \begin{split}
    \mathbf{\hat{SE}}_\text{prompt} = & \mathtt{dense}(\mathtt{concat}(\mathbf{SE}_\text{prompt},\mathbf{O}_\text{prev})) \\
    & + \mathbf{SE}_\text{prompt} \\
    & + \mathbf{O}_\text{prev} \\
    & \in\mathbb{R}^{B\times{P}\times{d}}
    \end{split}
\end{equation}
\begin{equation}
\label{eq:column-stack}
    \mathbf{SE}_\text{column}= \mathtt{stack}(\mathbf{E}_\text{column})\in\mathbb{R}^{B\times{C}\times{d}}
\end{equation}

\begin{equation}
\label{eq:soft-select}
    \mathbf{M}_\text{importance}=\mathtt{softmax}(\mathbf{\hat{SE}}_\text{prompt}\otimes\mathbf{SE}_\text{column}^\intercal)\in\mathbb{R}^{B\times{P}\times{C}}
\end{equation}

The output of the first part is $\mathbf{M}_\text{importance}\in\mathbb{R}^{B\times{P}\times{C}}$, which accommodates the feature importances yielded by $P$ prompts.
Notice that the column embeddings are not connected to the input and the prompt embeddings are fused with the previous cell's output.
In \cref{sec:discuss}, we further discuss these designs and their connections to the prompt learning of NLP.

\subsubsection{Construct and Expand Feature Embeddings}
\label{sec:input-transform}
In Trompt, categorical features are embedded through a embedding layer and numerical features are embedded through a dense layer as previous works \cite{somepalli2021saint,gorishniy2021revisiting}.
The embedding construction procedure is illustrated in part two of \cref{fig:cell}, where $\mathbf{E}_\text{feature}\in\mathbb{R}^{B\times{C}\times{d}}$ is the feature embeddings of the batch.

The shapes of $\mathbf{M}_\text{importance}$ and $\mathbf{E}_\text{feature}$ are $\mathbb{R}^{B\times{P}\times{C}}$ and $\mathbb{R}^{B\times{C}\times{d}}$, respectively.
Since $\mathbf{E}_\text{feature}$ lacks the prompt dimension, Trompt expands $\mathbf{E}_\text{feature}$ into $\mathbf{\hat{E}}_\text{feature}\in\mathbb{R}^{B\times{P}\times{C}\times{d}}$ to accommodate the $P$ prompts by a dense layer in part three of \cref{fig:cell}.

\subsubsection{Generate Output}
\label{sec:cell-output}
The output of Trompt Cell is the column-wise sum of the element-wise multiplication of $\mathbf{\hat{E}}_\text{feature}$ and $\mathbf{M}_\text{importance}$ as \cref{eq:output}, where $\odot$ is element-wise multiplication.
Notice that, during element-wise multiplication, the shape of $\mathbf{M}_\text{importance}$ is considered $\mathbb{R}^{B\times{P}\times{C}\times{1}}$.
In addition, since column is the third axis, the shape is reduced from $\mathbb{R}^{B\times{P}\times{C}\times{d}}$ to $\mathbb{R}^{B\times{P}\times{d}}$ after column-wise summation.
\begin{equation}
\label{eq:output}
    \mathbf{O}=\sum_{i=1}^C({\mathbf{\hat{E}}_\text{feature}\odot\mathbf{M}_\text{importance})_{:,:,i,:}}\in\mathbb{R}^{B\times{P}\times{d}}
\end{equation}

\begin{figure}[t]
    \centering
    \includegraphics[width=0.25\columnwidth]{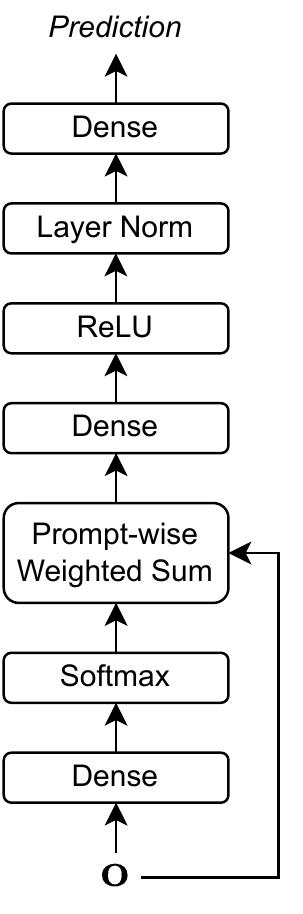}
    \caption{Architecture of a Trompt Downstream.}
    \label{fig:downstream}
\end{figure}

\subsection{Trompt Downstream}
\label{sec:downstream}
A Trompt Downstream makes a prediction based on a Trompt Cell's output, which contains representations corresponding to $P$ prompt embeddings.
To aggregate these representations, the weight for each prompt is first derived through a dense layer and a softmax activation function as \cref{eq:prompt-weight}.
Afterwards, the weighted sum is calculated as \cref{eq:prompt-aggregate}.

The prediction is subsequently made through two dense layers as \cref{eq:prediction}, where $T$  is the  target dimension.
For classification tasks, $T$ is the number of target classes.
For regression tasks, $T$ is set to 1.
As \cref{fig:overall} shows, a sample gets a prediction through a Trompt Cell and thus multiple predictions through all cells.
During training, the loss of each prediction is separately calculated and the loss is summed up to update model weights.
During inference, on the other hand, predictions through all cells are simply averaged as the final prediction as \cref{eq:loss-predict}, where $L$ is the number of Trompt Cells.

\begin{equation}
\label{eq:prompt-weight}
    \mathbf{W}_\mathtt{prompt}=\mathtt{softmax}(\mathtt{dense}(\mathbf{O}))\in\mathbb{R}^{B\times{P}}
\end{equation}
\begin{equation}
\label{eq:prompt-aggregate}
    \mathbf{\hat{O}}=\sum_{i=1}^P(\mathbf{W}_\text{prompt}\odot\mathbf{O})_{:,i,:}\in\mathbb{R}^{B\times{d}}
\end{equation}
\begin{equation}
\label{eq:prediction}
    \mathbf{P}=\mathtt{dense}(\mathtt{relu}(\mathtt{dense}(\mathbf{\hat{O}})))\in\mathbb{R}^{B\times{T}}
\end{equation}
\begin{equation}
\label{eq:loss-predict}
\begin{split}
    & loss=\sum_{i=1}^L{\mathtt{loss\char`_fn}(\mathbf{P}_i}, y) \\
    & pred=\sum_{i=1}^L{\mathbf{P}_i}/L
\end{split}
\end{equation}

\begin{table}
\caption{Analogy of the prompt learning of Trompt to that of NLP.}
\label{tab:prompt-learning}
\vskip 0.15in
\begin{center}
\begin{small}
\begin{tabular}{c | c | c}
\toprule
\thead{Problem\\Identification} & \thead{Implemented by} & \thead{Inspired by} \\
\midrule
\makecell[c]{Sample-invariant\\Intrinsic Properties} & $\mathbf{E}_\text{column}$ & \makecell[c]{Fixed Large\\Language Model} \\
\midrule
\makecell[c]{Sample-specific\\Feature Importances} & $\mathbf{M}_\text{importance}$ & \makecell[c]{Task-specific\\Predictions} \\
\bottomrule
\end{tabular}
\end{small}
\end{center}
\vskip -0.1in
\end{table}

\subsection{Prompt Learning of Trompt}
\label{sec:discuss}
Trompt's architecture is specifically designed for tabular data, taking into account the unique characteristics of this type of data and the impressive performance of tree-based models.
Unlike conventional operations, the design may appear unconventional and detached from tabular data features.
In this section, we explain the rationale behind Trompt's network design and how we adapted prompt learning to a tabular neural network.

Tabular data is structured, with each column representing a specific dataset property that remains constant across individual samples.
The success of tree-based models relies on assigning feature importances to individual samples.
This concept has been explored in models such as TabNet \cite{arik2021tabnet} and Net-DNF \cite{katzir2020net}.
However, tree-based algorithms do not explicitly assign feature importances to individual samples.
Instead, importances vary implicitly along the path from the root to a leaf node.
Only the columns involved in this path are considered important features for the samples reaching the corresponding leaf node, representing sample-specific feature importances.

Given the fundamental characteristic of tabular data and the learning strategy of tree-based models, Trompt aims to combine the intrinsic properties of columns with sample-specific feature importances using a prompt learning-inspired architecture from NLP \cite{radford2018improving,brown2020language}.
Trompt employs column embeddings to represent the intrinsic properties of each column and prompt embeddings to prompt column embeddings, generating feature importances for given prompts.
Both column embeddings and prompt embeddings are invariant across samples.
However, before prompting column embeddings with prompt embeddings, the prompt embeddings are fused with the output of the previous Trompt Cell as shown in \cref{eq:prompt-concat}, enabling input-related representations to flow through and derive sample-specific feature importances.
The "prompt" mechanism in Trompt is implemented as a matrix multiplication in \cref{eq:soft-select}.

A conceptual analogy of Trompt's prompt learning approach to NLP is presented in \cref{tab:prompt-learning}.
It's important to note that the implementation details of prompt learning differ substantially between tabular data and NLP tasks due to the fundamental differences between the two domains.
Therefore, appropriate adjustments must be made to bridge these two domains.

\section{Experiments}
\label{sec:experiments}
In this section, the experimental results and analyses are presented.
First, we elaborate on the settings of experiments and the configurations of Trompt in \cref{sec:setup}.
Second, the performance of Trompt on Grinsztajn45 is reported in \cref{sec:eval}.
Third, ablation studies regarding the hyperparameters and the architecture of Trompt are studied in \cref{sec:ablation}.
Lastly, the interpretability of Trompt is investigated using synthetic and real-world datasets in \cref{sec:interpretability}.

\subsection{Setup}
\label{sec:setup}

The performance and ablation study of Trompt primarily focus on the Grinsztajn45 benchmark \cite{grinsztajn2022why} \footnote{\href{https://github.com/LeoGrin/tabular-benchmark}{https://github.com/LeoGrin/tabular-benchmark}}. This benchmark comprises datasets from various domains and follows a unified methodology for evaluating different models, providing a fair and comprehensive assessment.
Furthermore, we evaluate the performance of Trompt on datasets selected by FT-Transformer and SAINT to compare it with state-of-the-art tabular neural networks.

For interpretability analysis, we follow the experimental settings of TabNet \cite{arik2021tabnet}.
This involves using two synthetic datasets (Syn2 and Syn4) and a real-world dataset (mushroom) to visualize attention masks.

The settings of Grinsztajn45 are presented in \cref{sec:benchmark-settings} and the implementation details of Trompt are presented in \cref{sec:implement-details}. 
Furthermore, the settings of datasets chosen by FT-Transformer and SAINT are provided in \cref{sec:more-eval-ft} and \cref{sec:more-eval-saint}, respectively.

\subsubsection{Settings of Grinsztajn45}
\label{sec:benchmark-settings}
To fairly evaluate the performance, we follow the configurations of Grinsztajn45, including train test data split, data preprocessing and evaluation metric.
Grinsztajn45 comprises two kinds of tasks, classification tasks and regression tasks.
Please see \cref{sec:g45-dataset-selection} and \cref{sec:g45-dataset-normalization} for the dataset selection criteria and dataset normalization process of Grinsztajn45.
The tasks are further grouped according to (i) the size of datasets (medium-sized and large-sized) and (ii) the inclusion of categorical features (numerical only and heterogeneous).

In addition, we make the following adjustments: (i) models with incomplete experimental results in \cite{grinsztajn2022why} are omitted, (ii) two well-performed tree-based models are added for comparison, and (iii) Trompt used a hyperparameter search space smaller than its opponents.
The details of the adjustments are described in \cref{sec:g45-models} and \cref{sec:g45-hpo}.

\subsubsection{Implementation Details}
\label{sec:implement-details}
Trompt is implemented using PyTorch.
The default hyperparameters are shown in \cref{tab:params}.
The size of embeddings and the hidden dimension of dense layers are configured $d$.
Note that only the size of column and prompt embeddings must be the same by the architecture design.
The hidden dimension of dense layers is set as $d$ to reduce hyperparameters and save computing resources.
On the other hand, the number of prompts and the number of Trompt Cells are set to $P$ and $L$.
Please refer to \cref{sec:hyper} for the hyperparameter search spaces for all baselines and Trompt.

\begin{table}[h]
\caption{Default hyperparameters of Trompt.}
\label{tab:params}
\vskip 0.15in
\begin{center}
\begin{small}
\begin{tabular}{l | c | c}
\toprule
\thead{Hyperparameter} & \thead{Symbol} & \thead{Value} \\
\midrule
\makecell[cl]{Feature Embeddings \\ Prompt/Column Embeddings \\ Hidden Dimension} & $d$ & 128 \\
\midrule
Prompts & $P$ & 128 \\
\midrule
Layer & $L$ & 6 \\
\bottomrule
\end{tabular}
\end{small}
\end{center}
\vskip -0.1in
\end{table}

\subsection{Evaluation Results}
\label{sec:eval}

The results of classification tasks are discussed in \cref{sec:eval-clf} and the results of regression tasks are discussed in \cref{sec:eval-reg}.
The evaluation metrics are accuracy and r2-score for classification and regression tasks, respectively.
In this section, we report an overall result and leave results of individual datasets in \cref{sec:more-eval-g45}.
In addition, the evaluation results on datasets chosen by FT-Transformer and SAINT are provided in \cref{sec:more-eval-ft} and \cref{sec:more-eval-saint}, respectively.

\subsubsection{Classification}
\label{sec:eval-clf}
\begin{figure}
    \centering
    \begin{subfigure}{.5\columnwidth}
        \centering
        \includegraphics[width=.95\linewidth]{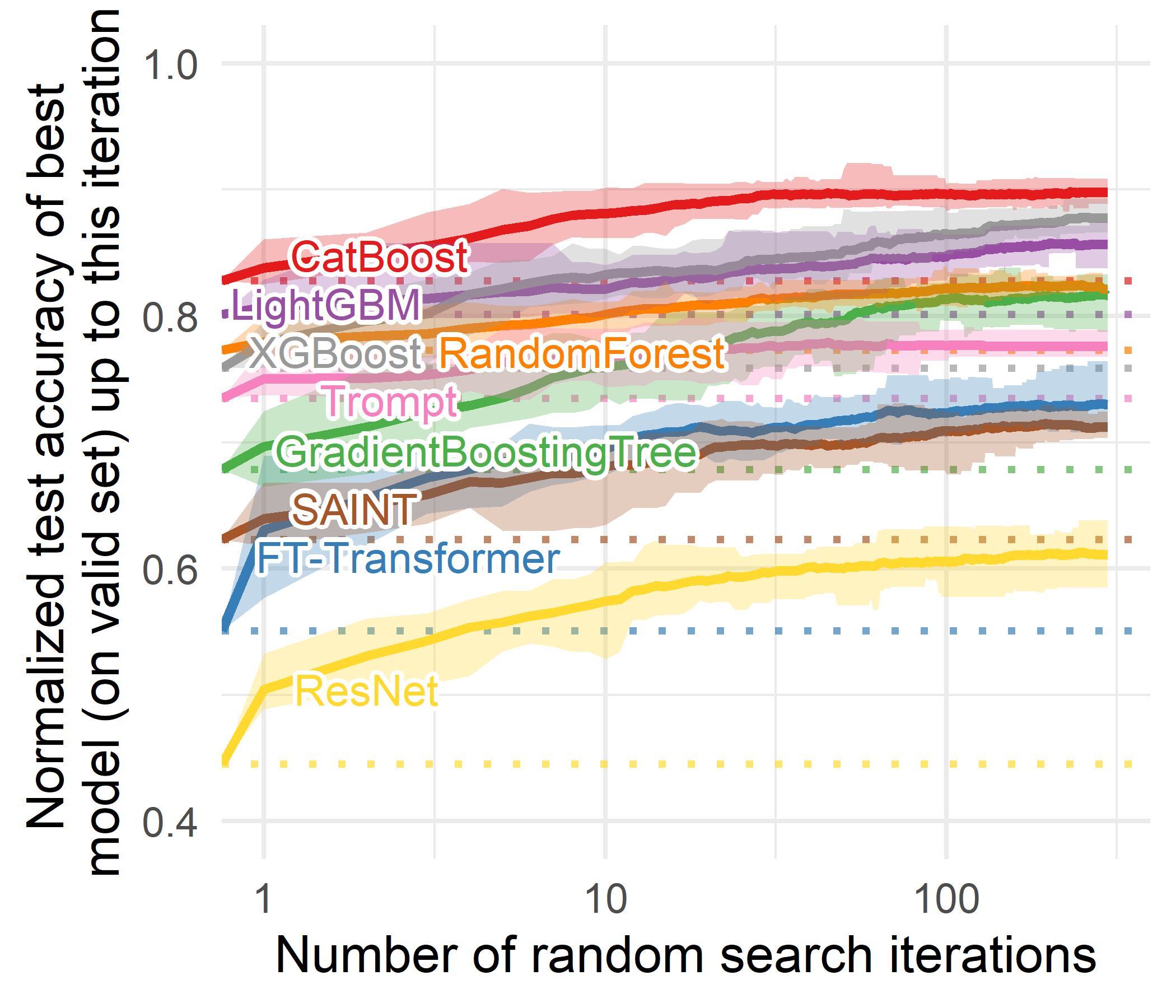}
        \caption{Numerical features only.}
        \label{fig:num-cls}
    \end{subfigure}%
    \begin{subfigure}{.5\columnwidth}
        \centering
        \includegraphics[width=.95\linewidth]{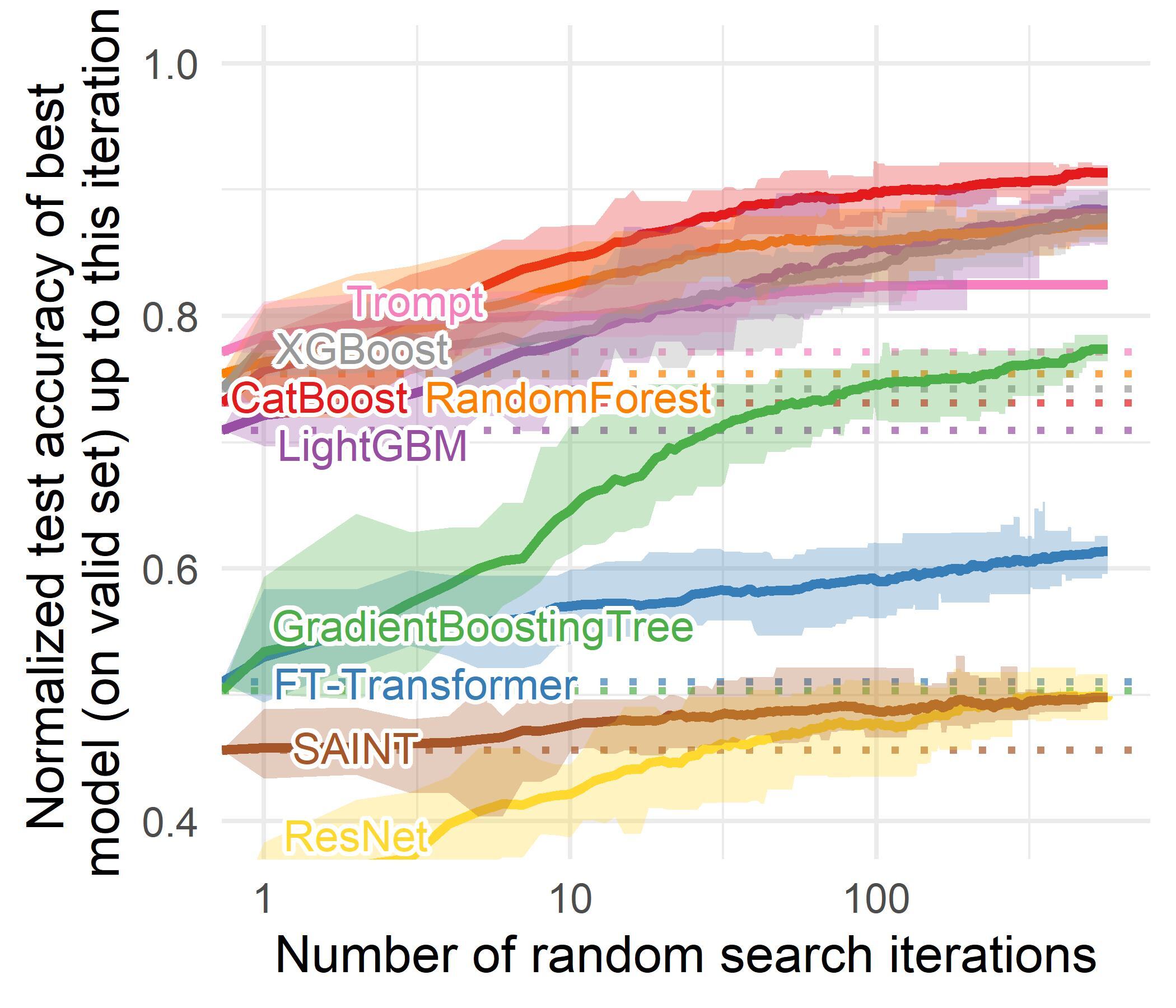}
        \caption{Heterogeneous features.}
        \label{fig:het-cls}
    \end{subfigure}
    \caption{Benchmark on \textbf{medium-sized classification} datasets.}
    \label{fig:eval-cls}
\end{figure}

On the medium-sized classification tasks, \cref{fig:eval-cls} shows that Trompt outperforms DNN models.
The curve of Trompt is consistently above deep neural networks (SAINT, FT-Transformer and ResNet) on tasks with and without categorical  features.
Additionally, Trompt narrows the gap between deep neural networks and tree-based models, especially on the tasks with heterogeneous features.
In \cref{fig:het-cls}, Trompt seems to be a member of the leading cluster with four tree-based models. The GradientBoostingTree starts slow but catches up the leading cluster in the end of search. The other deep neural networks forms the second cluster and have a gap to the leading one.

\begin{figure}
    \centering
    \begin{subfigure}{.5\columnwidth}
        \centering
        \includegraphics[width=.95\linewidth]{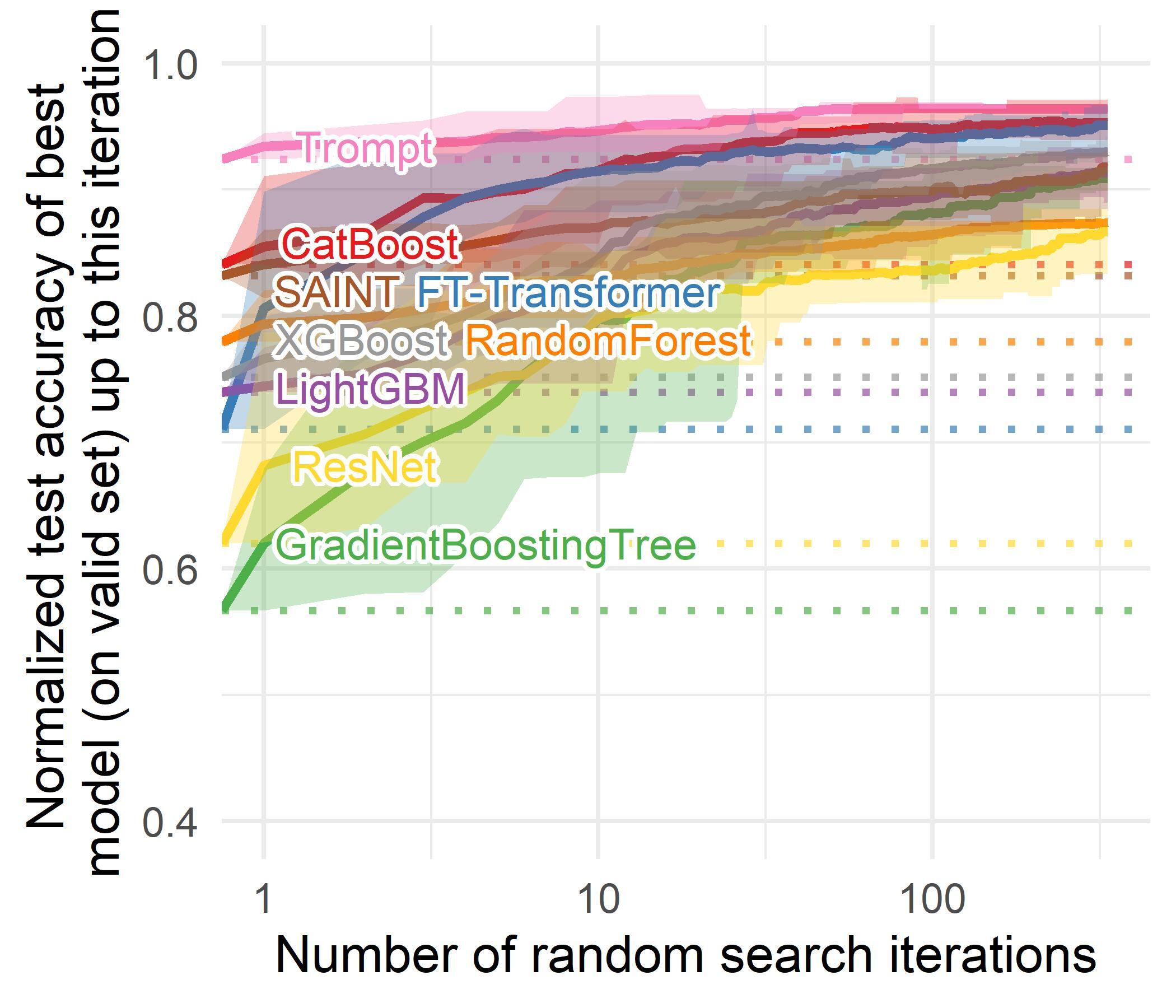}
        \caption{Numerical features only.}
        \label{fig:num-cls-large}
    \end{subfigure}%
    \begin{subfigure}{.5\columnwidth}
        \centering
        \includegraphics[width=.95\linewidth]{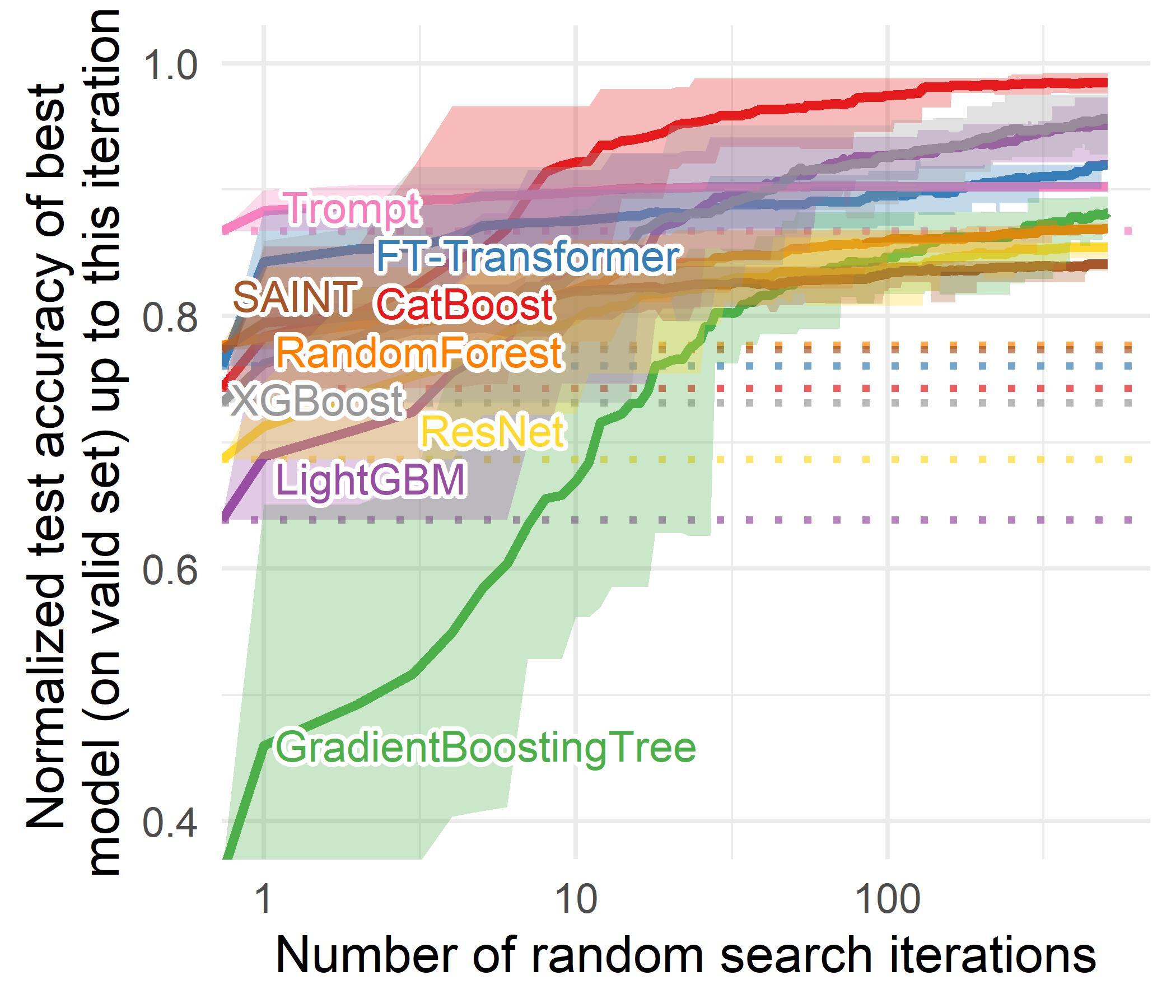}
        \caption{Heterogeneous features.}
        \label{fig:het-cls-large}
    \end{subfigure}
    \caption{Benchmark on \textbf{large-sized classification} datasets.}
    \label{fig:eval-cls-large}
\end{figure}

On the large-sized classification tasks, tree-based models remain the leading positions but the gap to deep neural networks is obscure.
This echoes that deep neural networks requires more samples for training \cite{lecun2015deep}.
\cref{fig:num-cls-large} shows that Trompt outperforms ALL models on the task with numerical features and \cref{fig:het-cls-large} shows that Trompt achieves a comparable performance to FT-Transformer on tasks with heterogeneous features.

With the small hyperparameter search space, the curve of Trompt is relatively flat.
The flat curve also suggests that Trompt performs well with its default hyperparameters.
Its performance after an exhausted search is worthy of future exploring.

\subsubsection{Regression}
\label{sec:eval-reg}
\begin{figure}
    \centering
    \begin{subfigure}{.5\columnwidth}
        \centering
        \includegraphics[width=.95\linewidth]{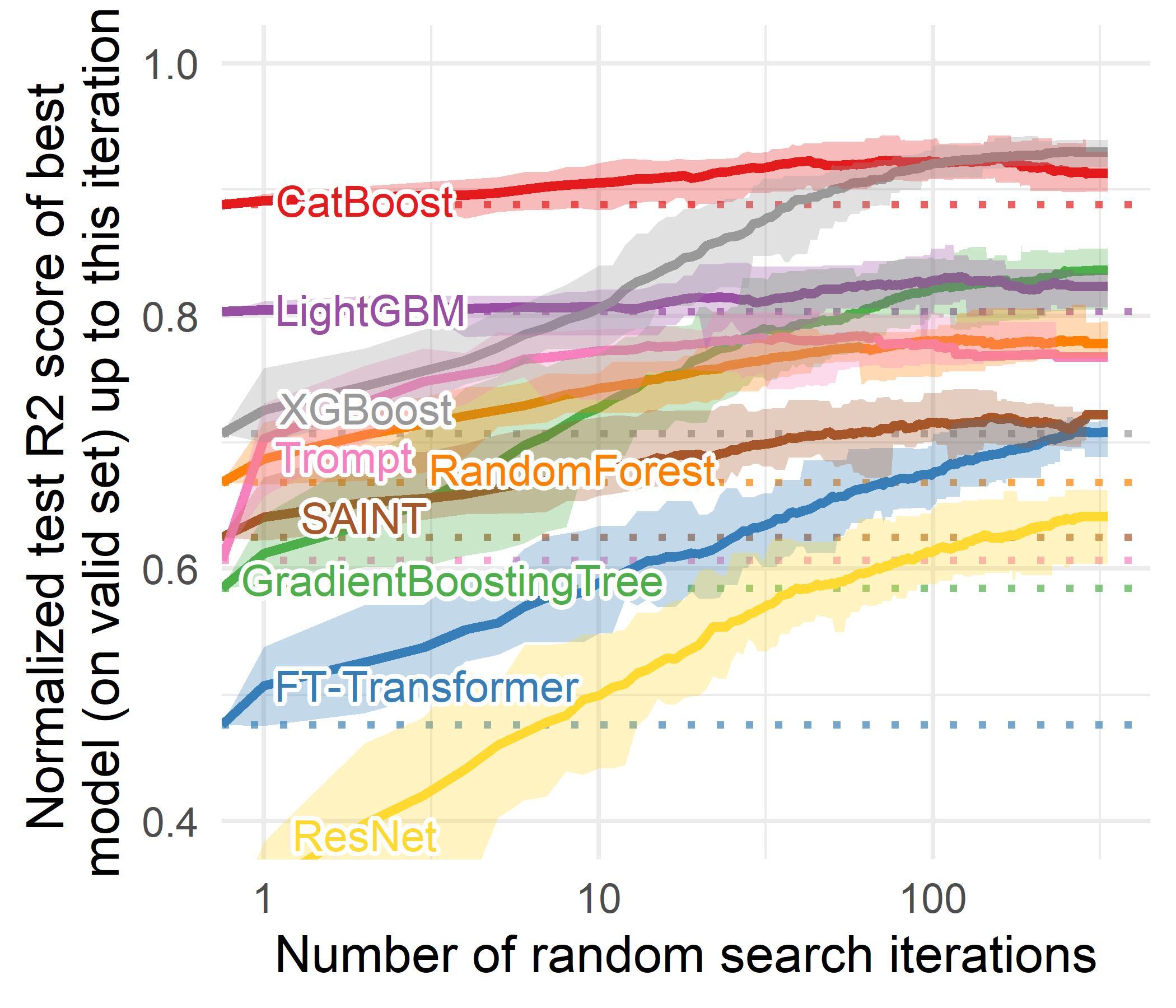}
        \caption{Numerical features only.}
        \label{fig:num-reg-quan}
        \vspace{.1cm}
    \end{subfigure}%
    \begin{subfigure}{.5\columnwidth}
        \centering
        \includegraphics[width=.95\linewidth]{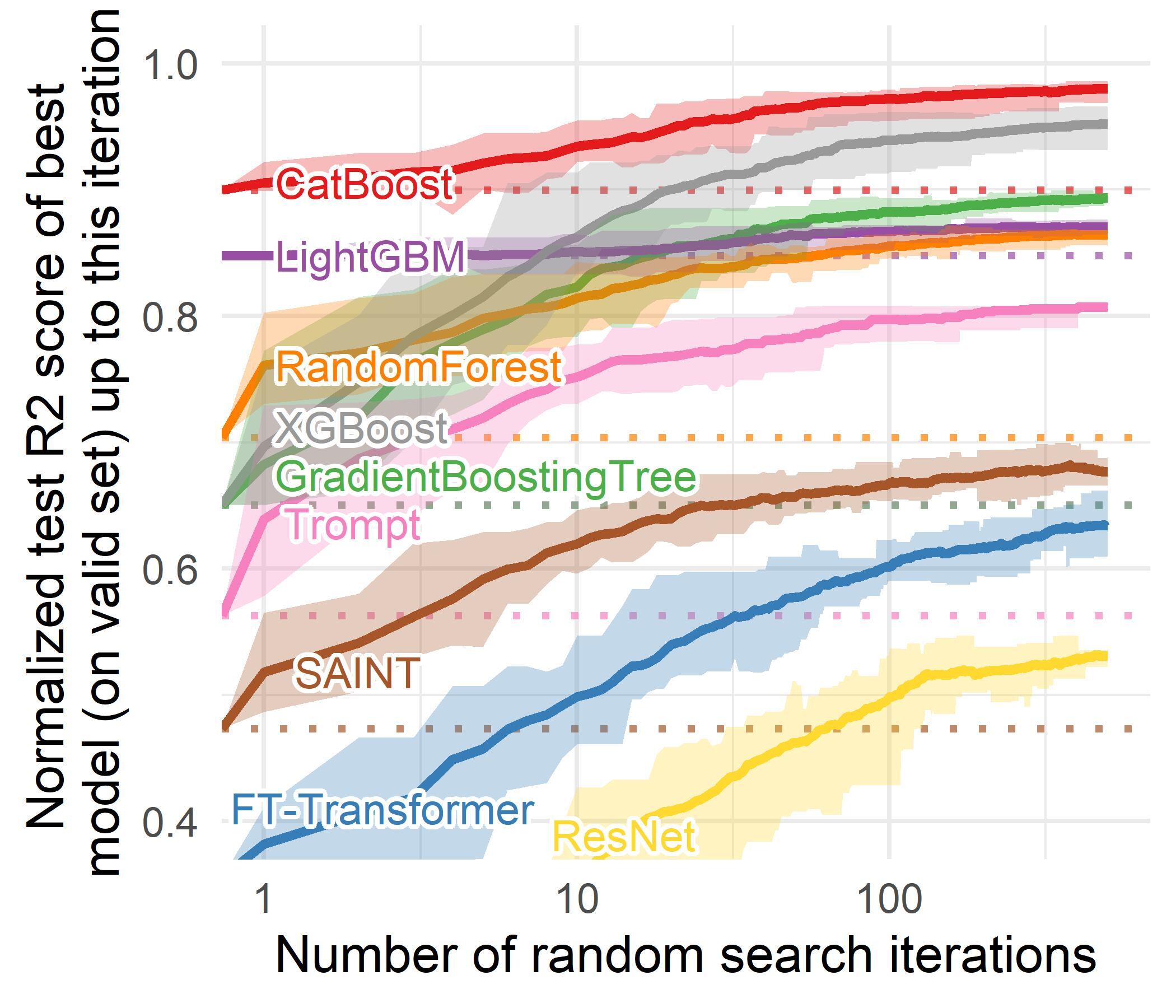}
        \caption{Heterogeneous features.}
        \label{fig:het-reg-quantile}
    \end{subfigure}
    \caption{Benchmark on \textbf{medium-sized regression} datasets.}
    \label{fig:eval-reg}
\end{figure}

On the medium-sized regression tasks, \cref{fig:eval-reg} shows that Trompt outperforms deep neural networks as the curves of Trompt are consistently higher than SAINT, FT-Transformer and ResNet on tasks with and without categorical features.
The gap between deep neural networks and tree-based models is less obvious in \cref{fig:num-reg-quan} than that in \cref{fig:het-reg-quantile}.
On the tasks with numerical features only, Trompt achieves a comparable performance with random forest.
On the tasks with heterogeneous features, Trompt narrows the gap but is below all the tree-based models.

\begin{figure}
    \centering
    \begin{subfigure}{.5\columnwidth}
        \centering
        \includegraphics[width=.95\linewidth]{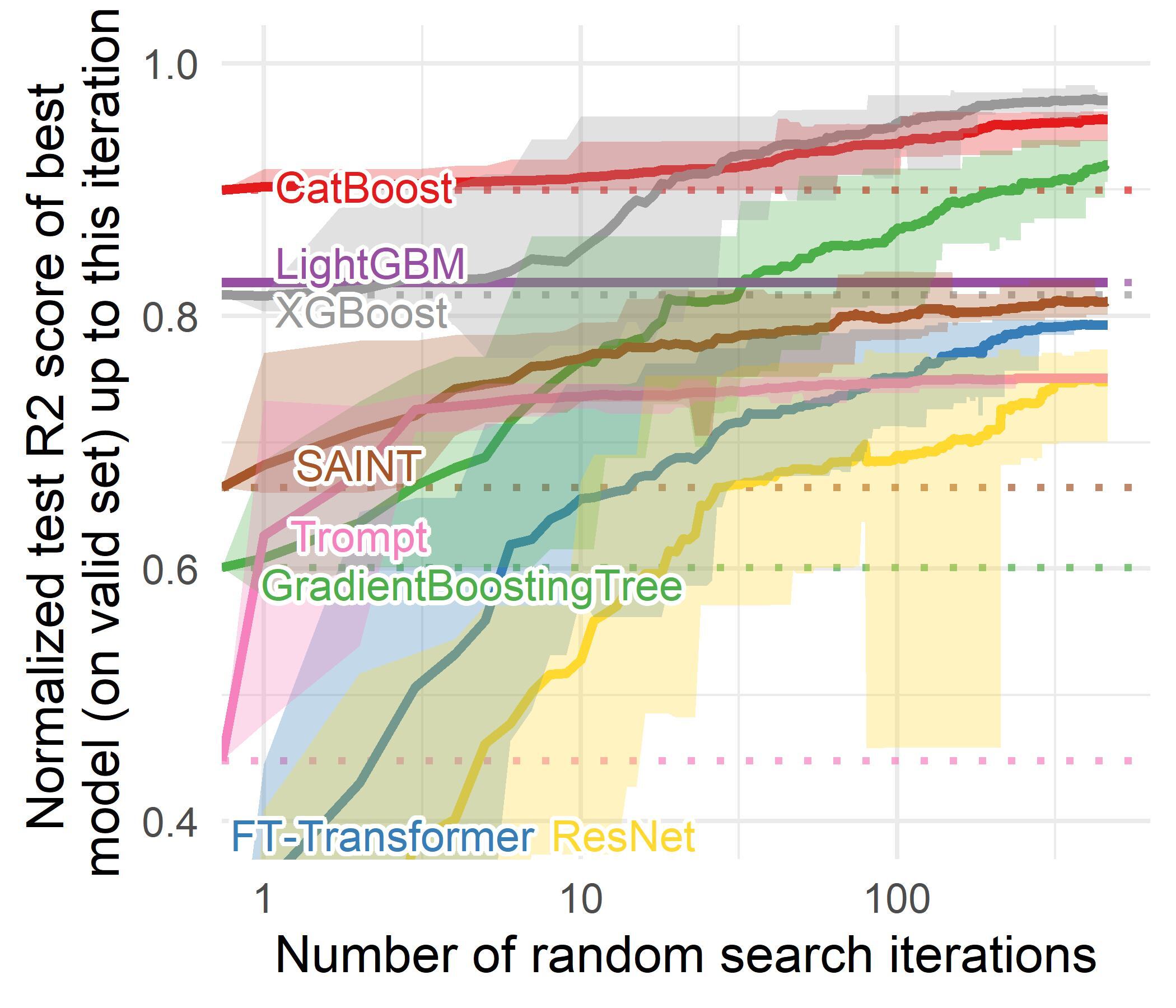}
        \caption{Numerical features only.}
        \label{fig:num-reg-quan-large}
        \vspace{.1cm}
    \end{subfigure}%
    \begin{subfigure}{.5\columnwidth}
        \centering
        \includegraphics[width=.95\linewidth]{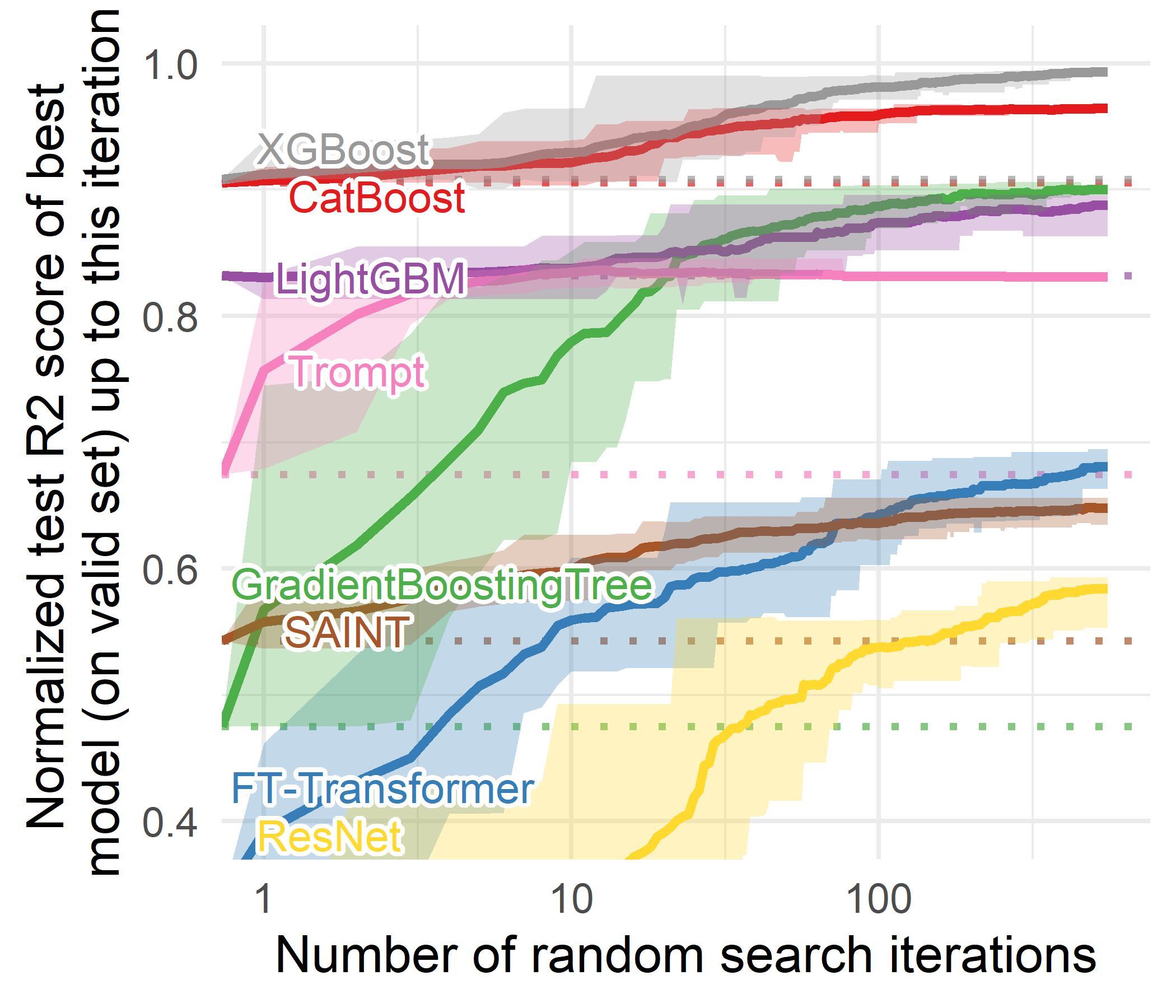}
        \caption{Heterogeneous features.}
        \label{fig:het-reg-quantile-large}
    \end{subfigure}
    \caption{Benchmark on \textbf{large-sized regression} datasets.}
    \label{fig:eval-reg-large}
\end{figure}

On the large-sized regression tasks with numerical features only, \cref{fig:num-reg-quan-large} shows that Trompt is slightly worse than SAINT and FT-Transformer in the end of search.
On the large-sized regression tasks with heterogeneous features, \cref{fig:het-reg-quantile-large} shows that Trompt outperforms deep neural networks with a large margin.

In general, deep learning models are not good at handling categorical features.
Trompt alleviates this weakness as shown in all tasks with heterogeneous features in \cref{fig:eval-cls}--\cref{fig:eval-reg-large}.
Trompt achieves superior performance over state-of-the-art deep neural networks except on the large-sized regression tasks with numerical features only.

\subsection{Ablation Study}
\label{sec:ablation}
In this subsection, we discuss the ablation study results of Trompt regarding hyperparameters and architecture design.
Please refer to \cref{sec:ablation-setting} for the settings of the ablation study.
In the main article, we report two major ablations on (i) the number of prompts and (ii) the necessity of expanding feature embeddings by a dense layer.
Other ablations can be found in \cref{sec:more-ablation}.

\textbf{Ablations on the number of prompts.}
Prompt embeddings ($\mathbf{E}_\text{prompt}$) stand a vital role to derive the feature importances.
Here we discuss the effectiveness of adjusting the number of prompts.

As shown in \cref{tab:ablation-prompts}, setting the number of prompts to one results in the worse results.
However, halving and doubling the default number (128) do not have much effect on the performance.
The results demonstrate that Trompt is not sensitive to the number of prompts, as long as the number of prompts is enough to accommodate the modalities of the dataset.
\begin{table}[h]
\caption{The performance of different number of prompts.}
\vskip 0.15in
\label{tab:ablation-prompts}
\begin{center}
\begin{small}
\begin{tabular}{l | c | c | c | c}
\toprule
 & \thead{1} & \thead{64} & \thead{128 (default)} & \thead{256}\\
\midrule
Classification & $79.74\%$ & $81.76\%$ & $81.81\%$ & $81.85\%$ \\
\midrule
Regression & $72.07\%$ & $74.11\%$ & $74.15\%$ & $74.14\%$ \\
\bottomrule
\end{tabular}
\end{small}
\end{center}
\vskip -0.1in
\end{table}

\textbf{Ablations on expanding feature embeddings by a dense layer.}
Part three of \cref{fig:cell} uses a dense layer to expand feature embeddings to accommodate $P$ prompts.
Here we discuss the necessity of the dense layer.

As you can see in \cref{tab:ablation-input-transform}, adding a dense layer really leads to better results and is a one of the key architecture designs of Trompt.
By design, adding the dense layer enables Trompt to generate different feature embeddings for each prompt.
Without the dense layer, Trompt is degraded to a simplified situation where each prompt uses the same feature embeddings.
The results of \cref{tab:ablation-prompts} and \cref{tab:ablation-input-transform} suggest that the variation of feature importances, which comes from both the prompt embedding and the expansion dense layer, is the key to the excellent performance of Trompt.

\begin{table}[h]
\caption{The performance of with and without applying feature transformation on Input Transform.}
\vskip 0.15in
\label{tab:ablation-input-transform}
\begin{center}
\begin{small}
\begin{tabular}{l | c | c}
\toprule
 & \thead{w (default)} & \thead{w/o}\\
\midrule
Classification & $81.81\%$ & $80.76\%$ \\
\midrule
Regression & $74.15\%$ & $73.73\%$ \\
\bottomrule
\end{tabular}
\end{small}
\end{center}
\vskip -0.1in
\end{table}

\begin{table*}[t]
\caption{The top-3 importance score ratio on the mushroom dataset.}
\vskip 0.15in
\label{tab:mushroom}
\begin{center}
\begin{small}
\begin{tabular}{l | c | c | c}
\toprule
 & \thead{1st} & \thead{2nd} & \thead{3rd} \\
\midrule
RandomForest & odor ($15.11\%$) & gill-size ($12.37\%$) & gill-color ($10.42\%$) \\
\midrule
XGBoost & spore-print-color ($29.43\%$) & odor ($22.71\%$) & cap-color ($14.07\%$) \\
\midrule
LightGBM & spore-print-color ($22.08\%$) & gill-color ($14.95\%$) & odor ($12.96\%$) \\
\midrule
CatBoost & odor ($72.43\%$) & spore-print-color ($10.57\%$) & gill-size ($2.71\%$) \\
\midrule
GradientBoostingTree & gill-color ($31.08\%$) & spore-print-color ($19.89\%$) & odor ($17.44\%$) \\
\midrule
Trompt (ours) & odor ($24.93\%$) & gill-size ($8.13\%$) & gill-color ($5.73\%$) \\
\bottomrule
\end{tabular}
\end{small}
\end{center}
\vskip -0.1in
\end{table*}

\subsection{Interpretability}
\label{sec:interpretability}
Besides outstanding performance, tree-based models are well-known for their interpretability.
Here we explore whether Trompt can also provide concise feature importances that highlighted salient features.
To investigate this, we conduct experiments on both synthetic datasets and real-world datasets, following the experimental design of TabNet \cite{arik2021tabnet}.
To derive the feature importances of Trompt for each sample, $\mathbf{M}_\text{importance}\in\mathbb{R}^{B\times{P}\times{C}}$ is reduced to $\mathbf{\hat{M}}_\text{importance}\in\mathbb{R}^{B\times{C}}$ as \cref{eq:interpretability}, where the weight of $\mathbf{M}_\text{importance}$ is the $\mathbf{W}_\text{prompt}$ of \cref{eq:prompt-weight}.

Notice that all Trompt Cells derive separated feature importances.
We demonstrate the averaged results of all cells here and leave the results of each cell in \cref{sec:attn-appendix}.

\begin{equation}
\label{eq:interpretability}
\mathbf{\hat{M}}_\text{importance}=\sum_{i=1}^P(\mathbf{W}_\text{prompt}\odot\mathbf{M}_\text{importance})_{:,i,:}\in\mathbb{R}^{B\times{C}}
\end{equation}

\textbf{Synthetic datasets.} 
The Syn2 and Syn4 datasets are used to study the feature importances learned by each model \cite{chen2018learning}.
A model is trained on oversampled training set (10k to 100k) using default hyperparameters and evaluated on 20 randomly picked testing samples.
The configuration is identical to that in TabNet \cite{arik2021tabnet}.

\begin{figure}[h]
    \centering
    \begin{subfigure}[t]{.5\columnwidth}
        \centering
        \includegraphics[width=.95\linewidth]{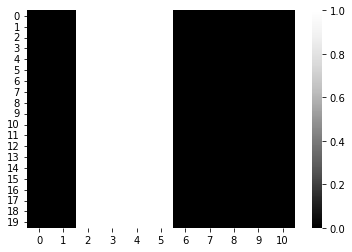}
        \caption{Important features.}
        \label{fig:syn2-real}
    \end{subfigure}%
    \begin{subfigure}[t]{.5\columnwidth}
        \centering
        \includegraphics[width=.95\linewidth]{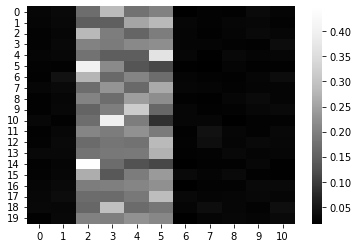}
        \caption{Feature importances of Trompt.}
        \label{fig:syn2-mask}
    \end{subfigure}
    \caption{Attention mask on Syn2 dataset (synthetic).}
    \label{fig:attn-syn2}
\end{figure}

\begin{figure}[h]
    \centering
    \begin{subfigure}[t]{.5\columnwidth}
        \centering
        \includegraphics[width=.95\linewidth]{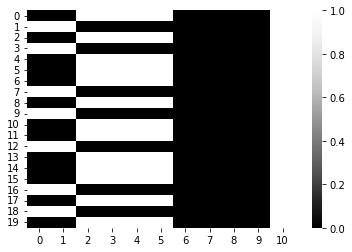}
        \caption{Important features.}
        \label{fig:syn4-real}
    \end{subfigure}%
    \begin{subfigure}[t]{.5\columnwidth}
        \centering
        \includegraphics[width=.95\linewidth]{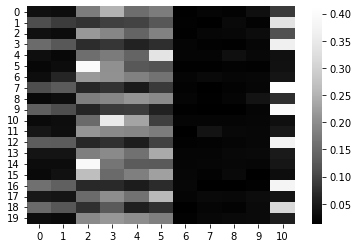}
        \caption{Feature importances of Trompt.}
        \label{fig:syn4-mask}
    \end{subfigure}
    \caption{Attention mask on Syn4 dataset (synthetic).}
    \label{fig:attn-syn4}
\end{figure}

\cref{fig:attn-syn2} and \cref{fig:attn-syn4} compare the important features of the dataset and those learned by Trompt.
In the Syn2 dataset, features 2--5 are important (\cref{fig:syn2-real}) and Trompt excellently focuses on them (\cref{fig:syn2-mask}).
In the Syn4 dataset, either features 0--1 or 2--5 could be important based on the value of feature 10 (\cref{fig:syn4-real}).
As \cref{fig:attn-syn4} shows, Trompt still properly focuses on features 0--5 and discovers the influence of feature 10.

\textbf{Real-world datasets.}
The mushroom dataset \cite{Dua:2019} is used as the real-world dataset for visualization as TabNet \cite{arik2021tabnet}.
With only the \emph{Odor} feature, most machine learning models can achieve $>95\%$ test accuracy \cite{arik2021tabnet}.
As a result, a high feature importance is expected on Odor.

\cref{tab:mushroom} shows the three most important features of Trompt and five tree-based models.
As shown, all models place Odor in their top three.
The second and third places of Trompt, \emph{gill-size} and \emph{gill-color}, also appear in the top three of the other models.
Actually, \emph{cap-color} is selected only by XGBoost.
If it is excluded, the union of the top important features of all models comes down to four features.
The one Trompt missed is \emph{spore-print-color}, which is the fifth place of Trompt.
Overall speaking, the important features selected by Trompt are consistent with those by tree-based models, and can therefore be used in various analyses that are familiar in the field of machine learning.

To further demonstrate that the experimental results were not ad-hoc, we repeat the experiments on additional real-world datasets.
Please see \cref{sec:add-real-data} for the details and experimental results.

\section{Discussion}
In this section, we further explore the “prompt” mechanism of Trompt.
\cref{sec:math-explain} clarifies the underlying hypothesis of how the prompt learning of Trompt fits for tabular data.
In addition, as Trompt is partially inspired by the learning strategy of tree-based models, we further discussed the difference between Trompt and tree-based models in \cref{sec:trompt-and-tree}.

\subsection{Further exploration of the "prompt" mechanism in Trompt}
\label{sec:math-explain}
The "prompt" mechanism in Trompt is realized as \cref{eq:soft-select}.
This equation involves a matrix multiplication of expanded prompt embeddings ($\mathbf{\hat{SE}}_\text{prompt}\in\mathbb{R}^{B\times{P}\times{d}}$) and the transpose of expanded column embeddings ($\mathbf{SE}_\text{column}\in\mathbb{R}^{B\times{C}\times{d}}$).
It results in $\mathbf{M}_\text{importance}\in\mathbb{R}^{P\times{C}}$, which represents prompt-to-column feature importances.
The matrix multiplication calculates the cosine-based distance between $\mathbf{\hat{SE}}_\text{prompt}$ and $\mathbf{SE}_\text{column}$, and favors high similarity between the sample-specific representations and sample-invariant intrinsic properties.

To make it clearer, $\mathbf{\hat{SE}}_\text{prompt}$ consists of $P$ embeddings that are specific to individual samples, except for the first Trompt Cell where $\mathbf{O}_\text{prev}$ is a zero tensor since there is no previous Trompt Cell, as stated in \cref{eq:prompt-stack,eq:prompt-concat}.
On the other hand, $\mathbf{SE}_\text{column}$ consists of $C$ embeddings that represent intrinsic properties specific to a tabular dataset as stated in \cref{eq:column-stack}.

Unlike self-attention, which calculates the distance between queries and keys and derives token-to-token similarity measures, Trompt calculates the distance between $\mathbf{\hat{SE}}_\text{prompt}$ and $\mathbf{SE}_\text{column}$ in \cref{{eq:soft-select}} to derive sample-to-intrinsic-property similarity measures.
The underlying idea of the calculation is to capture the distance between each sample and intrinsic property of a tabular dataset and we hypothesize that incorporating the intrinsic properties into the modeling of a tabular neural network might help making good predictions.

\subsection{The differences between Trompt and Tree-based Models}
As discussed in \cref{sec:discuss}, the idea of using prompt learning to derive feature importances, is inspired by the learning algorithm of tree-based models and the intrinsic properties of tabular data.
As a result, Trompt and tree-based models share a common characteristic in that they enable sample-dependent feature importances.
However, there are two main differences between them.
First, to incorporate the intrinsic properties of tabular data, Trompt uses column embeddings to share the column information across samples, while the learning strategy of tree-based models learn column information in their node-split nature.
Second, Trompt and tree-based models use different techniques to learn feature importance.
Trompt derives feature importances explicitly through prompt learning, while tree-based models vary the feature importances implicitly in the root-to-leaf path.

\label{sec:trompt-and-tree}

\section{Conclusion}
In this study, we introduce Trompt, a novel network architecture for tabular data analysis.
Trompt utilizes prompt learning to determine varying feature importances in individual samples.
Our evaluation shows that Trompt outperforms state-of-the-art deep neural networks (SAINT and FT-Transformer) and closes the performance gap between deep neural networks and tree-based models.

The emergence of prompt learning in deep learning is promising.
While the design of Trompt may not be intuitive or perfect for language model prompts, it demonstrates the potential of leveraging prompts in tabular data analysis.
This work introduces a new strategy for deep neural networks to challenge tree-based models and future research in this direction can explore more prompt-inspired architectures.

\nocite{}

\bibliography{example_paper}
\bibliographystyle{icml2023}

\newpage
\appendix
\onecolumn

\section{Settings of Grinsztajn45}
\label{sec:g45-settings}
In this section, we provide brief summaries with regard to dataset selection criteria in \cref{sec:g45-dataset-selection}, dataset normalization in \cref{sec:g45-dataset-normalization}, baseline models in \cref{sec:g45-models} and hyperparameter search mechanism in \cref{sec:g45-hpo}.

\subsection{Dataset Selection Criteria}
\label{sec:g45-dataset-selection}

Grinsztajn45 \cite{grinsztajn2022why} selects 45 tabular datasets from various domains mainly provided by OpenML \cite{OpenML2013}, which is listed in section A.1 of their paper.

The dataset selection criteria are summarized below.
Please refer to section 3.1 of the original paper for detailed selection criteria.
\begin{itemize}
\item The datasets contain heterogeneous features.
\item They are not high dimensional.
\item They contain I.I.D. data.
\item They contain real-world data.
\item They are not too small.
\item They are not too easy.
\item They are not deterministic.
\end{itemize}

\subsection{Dataset Normalization}
\label{sec:g45-dataset-normalization}

To ensure the homogeneity of the datasets and focus on challenges specific to tabular data, Grinsztajn45 did some modifications to the datasets to make sure that the datasets in the benchmark conform to the following criteria.
Please refer to section 3.2 of the original paper for detailed modification.
\begin{itemize}
\item The training sets are truncated to medium-sized (10,000) or large-sized (50,000).
\item All missing data were removed from the datasets.
\item The classes are balanced.
\item Categorical features with more than 20 items were removed
\item Numerical features with less than 10 unique values were removed.
\item Numerical features with 2 unique values are converted to categorical features.
\end{itemize}

\subsection{Baseline Models}
\label{sec:g45-models}
The paper by Grinsztajn45 presents the performance of four DNN models and four tree-based models. 
The DNN models include MLP \cite{gorishniy2021revisiting}, ResNet \cite{gorishniy2021revisiting}, FT-Transformer \cite{gorishniy2021revisiting}, and SAINT \cite{somepalli2021saint}.
The tree-based models consist of RandomForest \cite{breiman2001random}, GradientBoostingTree \cite{friedman2001greedy}, XGBoost \cite{chen2015xgboost}, and HistGradientBoostingTree \cite{sklearn2023histGBDT}.

However, two models, namely MLP \cite{gorishniy2021revisiting} and HistGradientBoostingTree \cite{sklearn2023histGBDT}, were omitted from the evaluation due to incomplete experimental results in Grinsztajn45 \cite{grinsztajn2022why}.
To provide a comprehensive comparison, we have included LightGBM \cite{ke2017lightgbm} and CatBoost \cite{prokhorenkova2018catboost} as additional models.
These models were selected based on their excellent performance and popularity.

\subsection{Hyperparameter Search Mechanism}
\label{sec:g45-hpo}
Grinsztajn45 evaluates models based on the results of a random search that consumes 20,000 compute-hours, as mentioned in Section 3.3 of the paper \cite{grinsztajn2022why}.
Since different models have varying inference and update times, the number of random search iterations completed within the same compute-hour differs for each model.
For instance, Model A may perform around two hundred iterations, while Model B may perform around three hundred iterations within 20,000 hours.
To ensure a fair evaluation, the iterations are truncated based on the minimum iteration count among all the compared models.

Due to limited computing resources, we have chosen a small search space (\cref{tab:texpert-params}) consisting of 40 parameter combinations.
To avoid unfairly truncating random search results of other models, and compromising the low search iterations of Trompt, we duplicated the grid search results of Trompt to exceed the lowest search iteration count among the models provided by Grinsztajn45.
For instance, if the lowest search iteration of a model was three hundreds, the search results of Trompt will be oversampled to surpass three hundreds and avoid being the lower bound, so other models can retain same search iterations as provided by Grinsztajn45.
As a result, the other models can retain the same search iterations as provided by Grinsztajn45.

Grinsztajn45's suggested evaluation procedure involves an extensive hyperparameter search that explores hundreds of parameter combinations.
However, due to limited computing resources, we have selected a smaller search space of 40 parameter combinations (\cref{tab:texpert-params} in \cref{sec:hyper}) for Trompt.
Please refer to \cref{sec:hyper} for the hyperparameter search spaces of all models.

\section{More Evaluation Results}
\label{sec:more-eval}

In \cref{sec:more-eval-g45}, we present additional evaluation results for Grinsztajn45, which expand upon the findings and analysis presented in the original paper \cite{grinsztajn2022why}.
These additional results provide further insights and contribute to a more comprehensive understanding of the evaluated models.

Furthermore, we include evaluation results on different datasets using the datasets selected by FT-Transformer \cite{gorishniy2021revisiting} and SAINT \cite{somepalli2021saint} in \cref{sec:more-eval-ft} and \cref{sec:more-eval-saint}, respectively.
By applying these datasets to the models, we aim to assess the performance of Trompt in different scenarios and gain a deeper understanding of its capabilities and generalizability.

\subsection{Grinsztajn45}
\label{sec:more-eval-g45}

In main paper, we have discussed the overall performance of Trompt using the learning curves during hyperparameter optimization.
In this section, we present quantitative evaluation results of both default and optimized hyperparameters.
In addition, we provide the figures of individual datasets for reference.

The quantitative evaluation results of classification and regression tasks are discussed in \cref{sec:performance-individual-clf} and \cref{sec:performance-individual-reg} respectively.
For classification datasets, we use \textbf{accuracy} as the evaluation metric.
For regression datasets, we use \textbf{r2-score} as the evaluation metric.
As a result, in both categories, the higher the number, the better the result.
Besides evaluation metrics, the \textbf{ranking} of each model is also provided.
To derive ranking, we calculate the mean and standard deviation of all rankings on datasets of a task.
Notice that since the names of some datasets are long, we first denote each dataset a notation in \cref{tab:not-med-data-1,tab:not-med-data-2,tab:not-large-data} and use them in following tables.

\begin{table}[H]
\caption{Notation of \textbf{medium-sized} datasets (1).}
\vskip 0.15in
\label{tab:not-med-data-1}
\begin{center}
\begin{small}
\begin{tabular}{c | c}
\toprule
\thead{Notation} & \thead{Dataset} \\
\midrule
A1 & KDDCup09\_upselling \\
\midrule
A2 & compass \\
\midrule
A3 & covertype \\
\midrule
A4 & electricity \\
\midrule
A5 & eye\_movements \\
\midrule
A6 & rl \\
\midrule
A7 & road-safety \\
\midrule
B1 & Higgs \\
\midrule
B2 & MagicTelescope \\
\midrule
B3 & MiniBooNE \\
\midrule
B4 & bank-marketing \\
\midrule
B5 & california \\
\midrule
B6 & covertype \\
\midrule
B7 & credit \\
\midrule
B8 & electricity \\
\midrule
B9 & eye\_movements \\
\midrule
B10 & house\_16H \\
\midrule
B11 & jannis \\
\midrule
B12 & kdd\_ipums\_la\_97-small \\
\midrule
B13 & phoneme \\
\midrule
B14 & pol \\
\midrule
B15 & wine \\
\bottomrule
\end{tabular}
\end{small}
\end{center}
\vskip -0.1in
\end{table}

\begin{table}[H]
\caption{Notation of \textbf{medium-sized} datasets (2).}
\vskip 0.15in
\label{tab:not-med-data-2}
\begin{center}
\begin{small}
\begin{tabular}{c | c}
\toprule
\thead{Notation} & \thead{Dataset} \\
\midrule
C1 & Bike\_Sharing\_Demand \\
\midrule
C2 & Brazilian\_houses \\
\midrule
C3 & Mercedes\_Benz\_Greener\_Manufacturing \\
\midrule
C4 & OnlineNewsPopularity \\
\midrule
C5 & SGEMM\_GPU\_kernel\_performance \\
\midrule
C6 & analcatdata\_supreme \\
\midrule
C7 & black\_friday \\
\midrule
C8 & diamonds \\
\midrule
C9 & house\_sales \\
\midrule
C10 & nyc-taxi-green-dec-2016 \\
\midrule
C11 & particulate-matter-ukair-2017 \\
\midrule
C12 & visualizing\_soil \\
\midrule
C13 & yprop\_4\_1 \\
\midrule
D1 & Ailerons \\
\midrule
D2 & Bike\_Sharing\_Demand \\
\midrule
D3 & Brazilian\_houses \\
\midrule
D4 & MiamiHousing2016 \\
\midrule
D5 & california \\
\midrule
D6 & cpu\_act \\
\midrule
D7 & diamonds \\
\midrule
D8 & elevators \\
\midrule
D9 & fifa \\
\midrule
D10 & house\_16H \\
\midrule
D11 & house\_sales \\
\midrule
D12 & houses \\
\midrule
D13 & medical\_charges \\
\midrule
D14 & nyc-taxi-green-dec-2016 \\
\midrule
D15 & pol \\
\midrule
D16 & sulfur \\
\midrule
D17 & superconduct \\
\midrule
D18 & wine\_quality \\
\midrule
D19 & year \\
\bottomrule
\end{tabular}
\end{small}
\end{center}
\vskip -0.1in
\end{table}

\begin{table}[H]
\caption{Notation of \textbf{large-sized} datasets.}
\vskip 0.15in
\label{tab:not-large-data}
\begin{center}
\begin{small}
\begin{tabular}{c | c}
\toprule
\thead{Notation} & \thead{Dataset} \\
\midrule
$\mathbb{A}$1 & covertype \\
\midrule
$\mathbb{A}$2 & road-safety \\
\midrule
$\mathbb{B}$1 & covertype \\
\midrule
$\mathbb{B}$2 & Higgs \\
\midrule
$\mathbb{B}$3 & MiniBooNE \\
\midrule
$\mathbb{B}$4 & jannis \\
\midrule
$\mathbb{C}$1 & black\_friday \\
\midrule
$\mathbb{C}$2 & diamonds \\
\midrule
$\mathbb{C}$3 & nyc-taxi-green-dec-2016 \\
\midrule
$\mathbb{C}$4 & particulate-matter-ukair-2017 \\
\midrule
$\mathbb{C}$5 & SGEMM\_GPU\_kernel\_performance \\
\midrule
$\mathbb{D}$1 & diamonds \\
\midrule
$\mathbb{D}$2 & nyc-taxi-green-dec-2016 \\
\midrule
$\mathbb{D}$3 & year \\
\bottomrule
\end{tabular}
\end{small}
\end{center}
\vskip -0.1in
\end{table}

\subsubsection{Classification}
\label{sec:performance-individual-clf}

The evaluation results for medium-sized classification tasks are presented in \cref{tab:performance-medium-het-clf} for heterogeneous features, and in \cref{tab:performance-medium-num-clf-1,tab:performance-medium-num-clf-2} for numerical features only.

For large-sized classification tasks, the results can be found in \cref{tab:performance-large-het-clf} for heterogeneous features, and in \cref{tab:performance-large-num-clf} for numerical features only.

\begin{figure}[H]
    \centering
    \includegraphics[width=.95\linewidth]{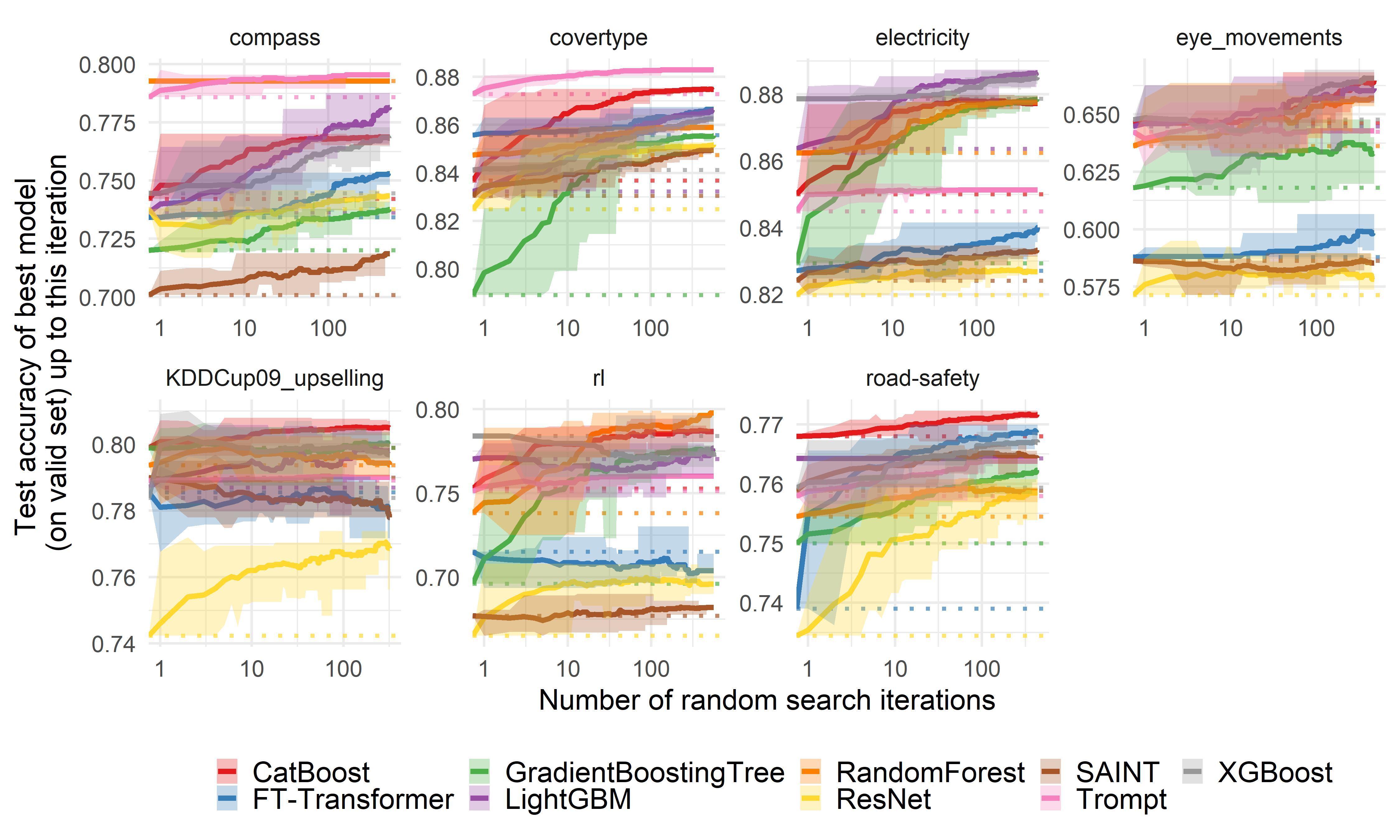}
    \caption{Benchmark on \textbf{\emph{every} medium-sized classification} dataset with \textbf{heterogeneous features}.}
    \label{fig:cat-cls-ind}
\end{figure}

\begin{figure}[H]
    \centering
    \includegraphics[width=.95\linewidth]{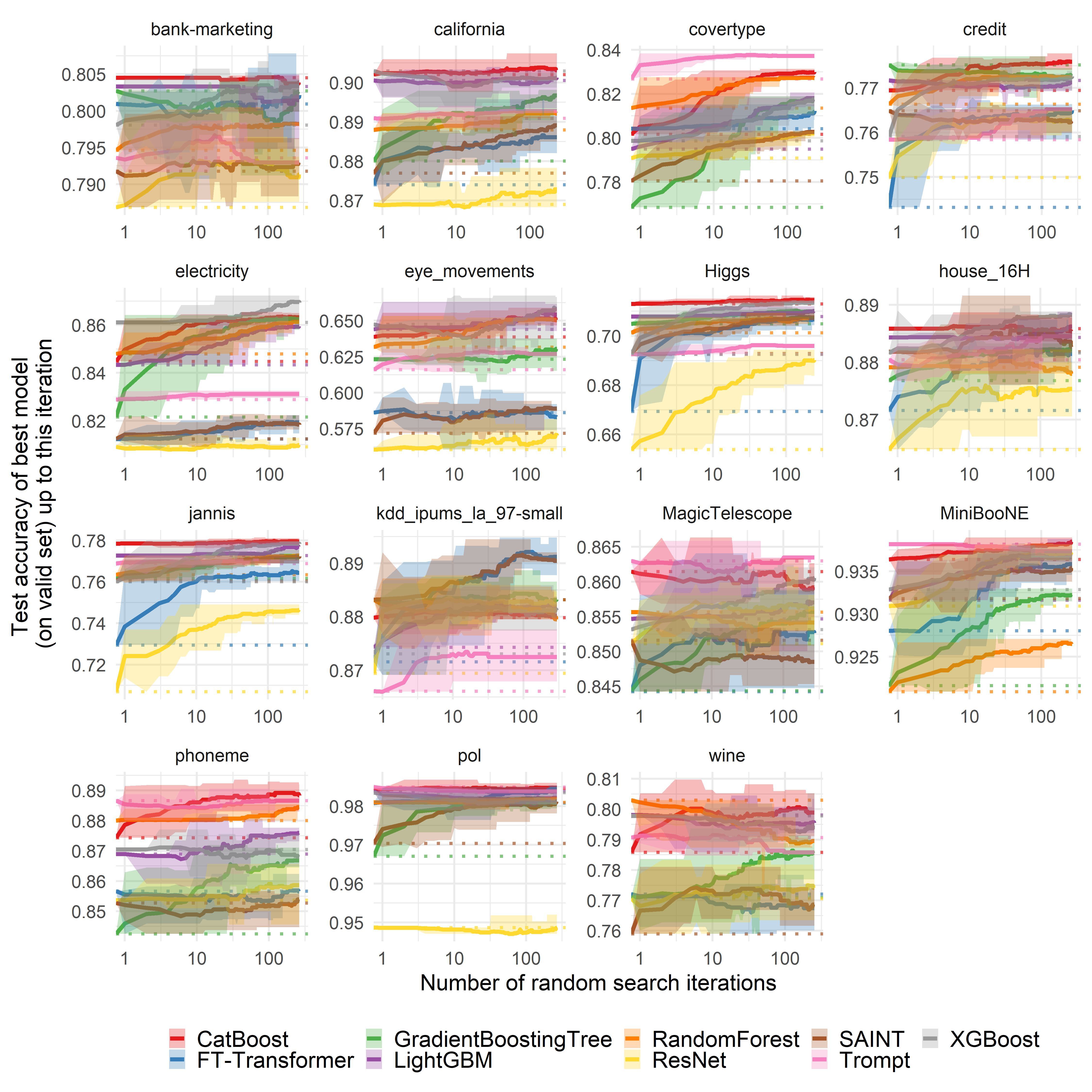}
    \caption{Benchmark on \textbf{\emph{every} medium-sized classification} dataset with \textbf{numerical features only}.}
    \label{fig:num-cls-ind}
\end{figure}

\begin{figure}[H]
    \centering
    \includegraphics[width=.6\linewidth]{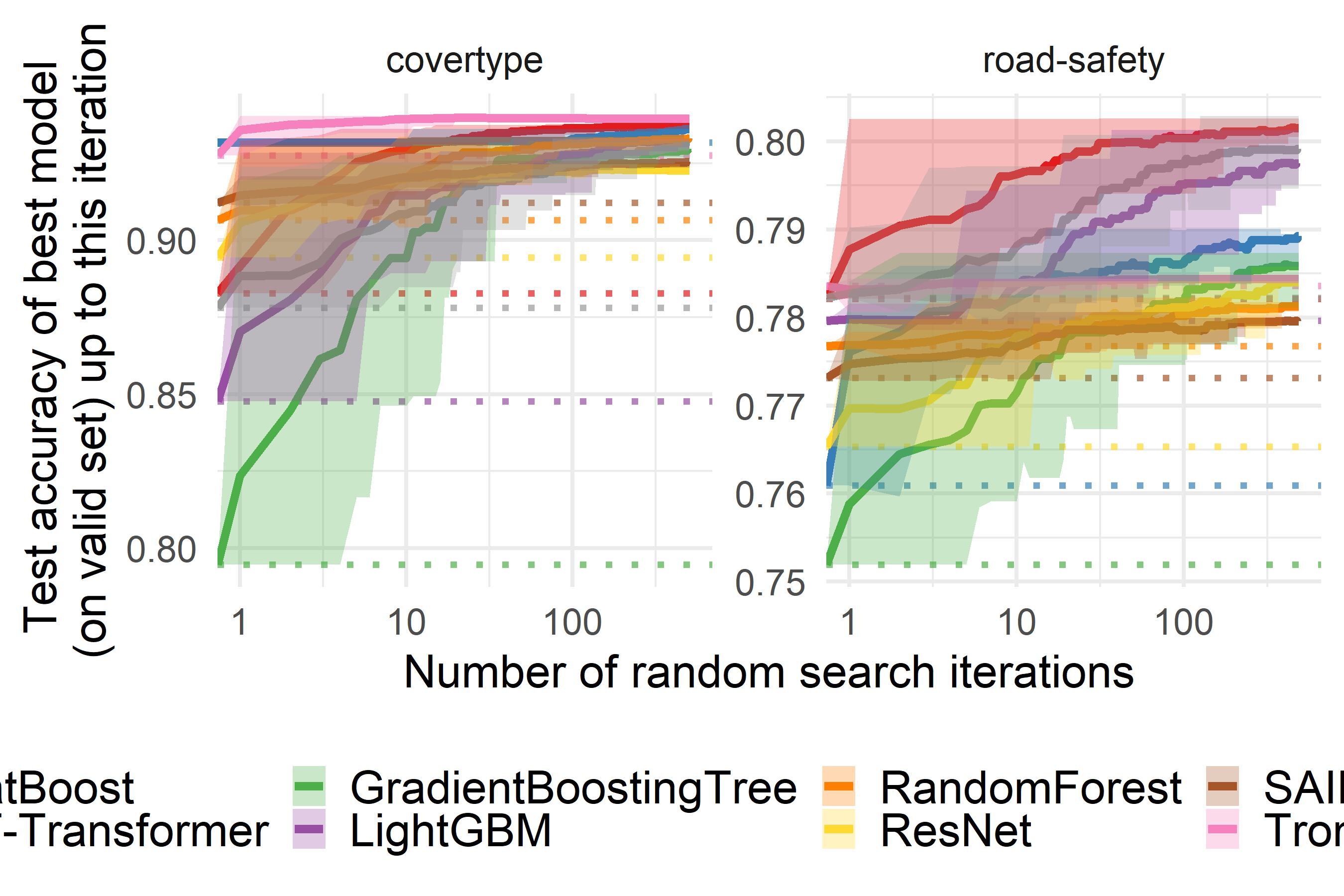}
    \caption{Benchmark on \textbf{\emph{every} large-sized classification} dataset with \textbf{heterogeneous features}.}
    \label{fig:cat-cls-ind-large}
\end{figure}

\begin{figure}[H]
    \centering
    \includegraphics[width=.95\linewidth]{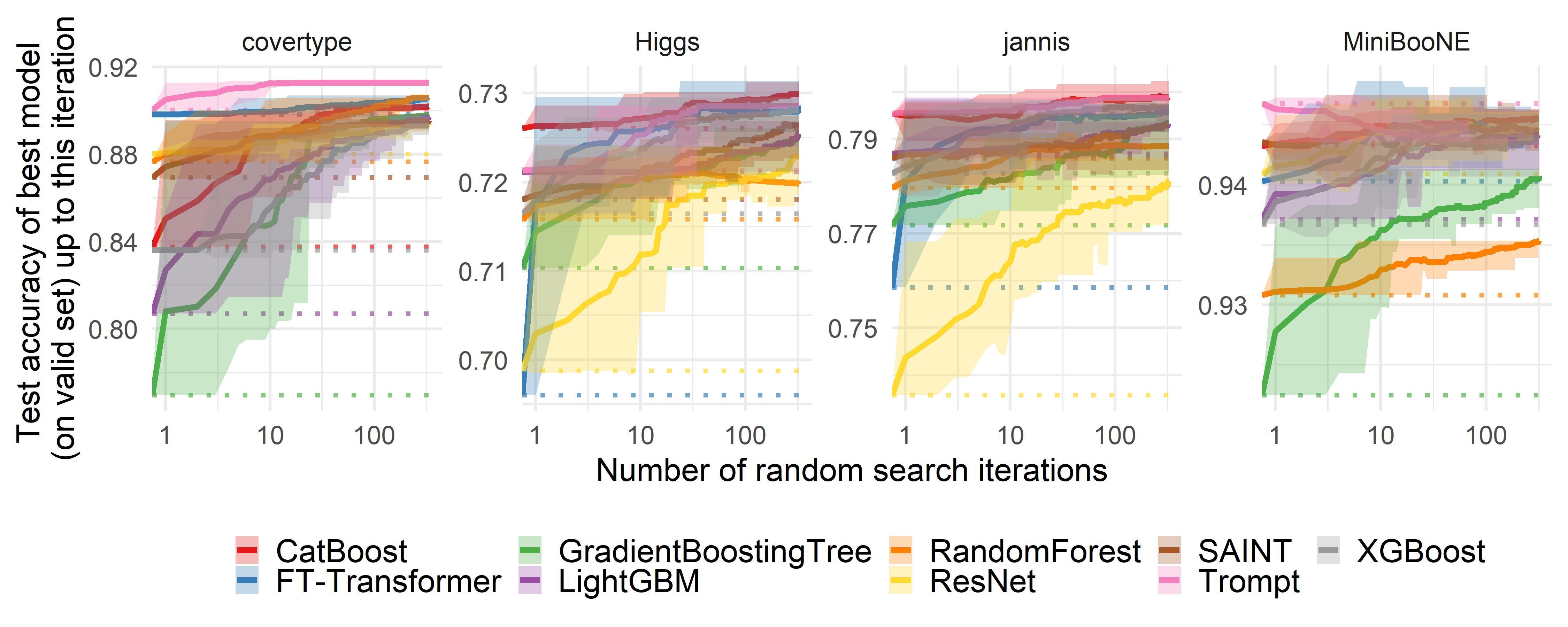}
    \caption{Benchmark on \textbf{\emph{every} large-sized classification} dataset with \textbf{numerical features only}.}
    \label{fig:num-cls-ind-large}
\end{figure}

Furthermore, individual figures illustrating the performance of Trompt on medium-sized tasks are provided in \cref{fig:cat-cls-ind} for heterogeneous features, and in \cref{fig:num-cls-ind} for numerical features only.
The individual figures for large-sized tasks can be found in \cref{fig:cat-cls-ind-large} for heterogeneous features, and in \cref{fig:num-cls-ind-large} for numerical features only.

The evaluation results consistently demonstrate that Trompt outperforms state-of-the-art deep neural networks (FT-Transformer and SAINT) across all classification tasks (refer to \cref{tab:performance-medium-het-clf,tab:performance-medium-num-clf-1,tab:performance-medium-num-clf-2,tab:performance-large-het-clf,tab:performance-large-num-clf}).
Moreover, Trompt's default rankings consistently yield better performance than the searched rankings, indicating its strength in default configurations without tuning.
Remarkably, in a large-sized task with numerical features only, Trompt even surpasses tree-based models (refer to \cref{tab:performance-large-num-clf}).

\begin{table}[H]
\caption{The performance of \textbf{medium-sized classification} task (\emph{heterogeneous features}).}
\vskip 0.15in
\label{tab:performance-medium-het-clf}
\begin{center}
\begin{small}
\begin{tabular}{l | c | c| c | c | c | c | c | c}
\toprule
 & \thead{A1} & \thead{A2} & \thead{A3} & \thead{A4} & \thead{A5} & \thead{A6} & \thead{A7} & \thead{Ranking} \\
\midrule
\multicolumn{9}{c}{Default} \\
\midrule
Trompt (ours) & $78.91\%$ & $78.59\%$ & $87.29\%$ & $84.50\%$ & $64.25\%$ & $75.13\%$ & $75.80\%$ & $3.71\pm1.78$ \\
\midrule
FT-Transformer & $78.56\%$ & $73.43\%$ & $85.57\%$ & $82.71\%$ & $58.79\%$ & $71.52\%$ & $73.90\%$ & $6.29\pm2.00$ \\
\midrule
ResNet & $74.24\%$ & $73.78\%$ & $82.49\%$ & $81.99\%$ & $57.14\%$ & $66.51\%$ & $73.45\%$ & $8.29\pm2.92$ \\
\midrule
SAINT & $79.00\%$ & $70.09\%$ & $83.04\%$ & $82.42\%$ & $58.62\%$ & $67.69\%$ & $75.89\%$ & $6.86\pm2.55$ \\
\midrule
CatBoost & $79.90\%$ & $74.22\%$ & $83.69\%$ & $85.01\%$ & $64.62\%$ & $75.29\%$ & $76.80\%$ & $2.93\pm2.55$ \\
\midrule
LightGBM & $78.70\%$ & $73.63\%$ & $83.23\%$ & $86.37\%$ & $64.48\%$ & $77.04\%$ & $76.43\%$ & $3.86\pm1.93$ \\
\midrule
XGBoost & $78.39\%$ & $74.46\%$ & $84.13\%$ & $87.86\%$ & $64.77\%$ & $78.42\%$ & $75.94\%$ & $3.00\pm2.92$ \\
\midrule
RandomForest & $79.38\%$ & $79.28\%$ & $84.75\%$ & $86.24\%$ & $63.62\%$ & $73.82\%$ & $75.45\%$ & $3.71\pm1.86$ \\
\midrule
GradientBoostingTree & $79.90\%$ & $72.01\%$ & $78.92\%$ & $82.94\%$ & $61.81\%$ & $69.60\%$ & $75.00\%$ & $6.36\pm2.51$ \\
\midrule
\multicolumn{9}{c}{Searched} \\
\midrule
Trompt (ours) & $79.00\%$ & $79.55\%$ & $88.29\%$ & $85.13\%$ & $64.29\%$ & $76.02\%$ & $76.38\%$ & $4.43\pm2.20$ \\
\midrule
FT-Transformer & $78.00\%$ & $75.30\%$ & $86.64\%$ & $84.01\%$ & $59.85\%$ & $70.38\%$ & $76.86\%$ & $5.57\pm2.19$ \\
\midrule
ResNet & $76.87\%$ & $74.35\%$ & $85.17\%$ & $82.68\%$ & $57.82\%$ & $69.59\%$ & $75.85\%$ & $8.43\pm2.73$ \\
\midrule
SAINT & $77.80\%$ & $71.87\%$ & $84.95\%$ & $83.32\%$ & $58.54\%$ & $68.20\%$ & $76.43\%$ & $7.86\pm2.64$ \\
\midrule
CatBoost & $80.50\%$ & $76.87\%$ & $87.48\%$ & $87.73\%$ & $66.48\%$ & $78.67\%$ & $77.16\%$ & $2.43\pm2.71$ \\
\midrule
LightGBM & $79.81\%$ & $78.15\%$ & $86.62\%$ & $88.64\%$ & $66.14\%$ & $77.69\%$ & $76.43\%$ & $3.14\pm2.05$ \\
\midrule
XGBoost & $79.69\%$ & $76.83\%$ & $86.25\%$ & $88.52\%$ & $66.57\%$ & $77.18\%$ & $76.69\%$ & $3.57\pm1.93$ \\
\midrule
RandomForest & $79.38\%$ & $79.28\%$ & $85.89\%$ & $87.76\%$ & $65.70\%$ & $79.79\%$ & $75.88\%$ & $4.29\pm2.27$ \\
\midrule
GradientBoostingTree & $80.01\%$ & $73.77\%$ & $85.55\%$ & $87.85\%$ & $63.30\%$ & $77.58\%$ & $76.23\%$ & $5.29\pm2.17$ \\
\bottomrule
\end{tabular}
\end{small}
\end{center}
\vskip -0.1in
\end{table}

\begin{table}[H]
\caption{The performance of \textbf{medium-sized classification} task (\emph{numerical features only}) (1).}
\vskip 0.15in
\label{tab:performance-medium-num-clf-1}
\begin{center}
\begin{small}
\begin{tabular}{l | c | c| c | c | c | c | c | c}
\toprule
 & \thead{B1} & \thead{B2} & \thead{B3} & \thead{B4} & \thead{B5} & \thead{B6} & \thead{B7} & \thead{B8} \\
 \midrule
\multicolumn{9}{c}{Default} \\
\midrule
Trompt (ours) & $69.26\%$ & $86.30\%$ & $93.82\%$ & $79.36\%$ & $89.09\%$ & $82.68\%$ & $75.84\%$ & $82.89\%$ \\
\midrule
FT-Transformer & $66.94\%$ & $84.42\%$ & $92.80\%$ & $80.09\%$ & $87.40\%$ & $80.42\%$ & $74.32\%$ & $81.24\%$ \\
\midrule
ResNet & $65.39\%$ & $85.11\%$ & $93.10\%$ & $78.68\%$ & $86.90\%$ & $79.09\%$ & $74.99\%$ & $80.91\%$ \\
\midrule
SAINT & $69.29\%$ & $85.16\%$ & $93.18\%$ & $79.18\%$ & $87.69\%$ & $78.05\%$ & $76.49\%$ & $81.25\%$ \\
\midrule
CatBoost & $71.30\%$ & $86.14\%$ & $93.64\%$ & $80.45\%$ & $90.21\%$ & $80.16\%$ & $76.95\%$ & $84.48\%$ \\
\midrule
LightGBM & $70.79\%$ & $85.47\%$ & $93.16\%$ & $80.33\%$ & $90.06\%$ & $79.50\%$ & $77.17\%$ & $84.34\%$ \\
\midrule
XGBoost & $69.25\%$ & $85.31\%$ & $93.29\%$ & $79.81\%$ & $90.30\%$ & $79.87\%$ & $75.91\%$ & $86.11\%$ \\
\midrule
RandomForest & $70.12\%$ & $85.56\%$ & $92.09\%$ & $79.46\%$ & $88.80\%$ & $81.35\%$ & $76.64\%$ & $84.79\%$ \\
\midrule
GradientBoostingTree & $70.49\%$ & $84.44\%$ & $92.16\%$ & $80.27\%$ & $88.00\%$ & $76.85\%$ & $77.52\%$ & $82.16\%$ \\
\midrule
\multicolumn{9}{c}{Searched} \\
\midrule
Trompt (ours) & $69.60\%$ & $86.35\%$ & $93.74\%$ & $79.30\%$ & $89.28\%$ & $83.73\%$ & $76.52\%$ & $83.12\%$ \\
\midrule
FT-Transformer & $70.67\%$ & $85.26\%$ & $93.59\%$ & $80.22\%$ & $88.61\%$ & $81.22\%$ & $76.50\%$ & $81.94\%$ \\
\midrule
ResNet & $69.02\%$ & $85.62\%$ & $93.69\%$ & $79.13\%$ & $87.28\%$ & $80.21\%$ & $76.28\%$ & $80.98\%$ \\
\midrule
SAINT & $70.73\%$ & $84.85\%$ & $93.54\%$ & $79.29\%$ & $88.92\%$ & $80.27\%$ & $76.24\%$ & $81.84\%$ \\
\midrule
CatBoost & $71.46\%$ & $85.92\%$ & $93.84\%$ & $80.39\%$ & $90.32\%$ & $82.98\%$ & $77.59\%$ & $86.33\%$ \\
\midrule
LightGBM & $71.01\%$ & $85.70\%$ & $93.71\%$ & $80.15\%$ & $90.13\%$ & $81.81\%$ & $77.13\%$ & $85.94\%$ \\
\midrule
XGBoost & $71.36\%$ & $86.05\%$ & $93.66\%$ & $80.34\%$ & $90.12\%$ & $81.75\%$ & $77.26\%$ & $86.94\%$ \\
\midrule
RandomForest & $70.76\%$ & $85.41\%$ & $92.65\%$ & $79.82\%$ & $89.21\%$ & $82.73\%$ & $77.25\%$ & $86.14\%$ \\
\midrule
GradientBoostingTree & $71.00\%$ & $85.57\%$ & $93.22\%$ & $80.26\%$ & $89.68\%$ & $81.72\%$ & $77.27\%$ & $86.24\%$ \\
\bottomrule
\end{tabular}
\end{small}
\end{center}
\vskip -0.1in
\end{table}

\begin{table}[H]
\caption{The performance of \textbf{medium-sized classification} task (\emph{numerical features only}) (2).}
\vskip 0.15in
\label{tab:performance-medium-num-clf-2}
\begin{center}
\begin{small}
\begin{tabular}{l | c | c| c | c | c | c | c | c}
\toprule
 & \thead{B9} & \thead{B10} & \thead{B11} & \thead{B12} & \thead{B13} & \thead{B14} & \thead{B15} & \thead{Ranking} \\
\midrule
\multicolumn{9}{c}{Default} \\
\midrule
Trompt (ours) & $61.60\%$ & $88.05\%$ & $76.89\%$ & $86.61\%$ & $88.67\%$ & $98.49\%$ & $79.07\%$ & $4.07\pm2.61$ \\
\midrule
FT-Transformer & $58.62\%$ & $87.16\%$ & $72.94\%$ & $87.16\%$ & $85.67\%$ & $98.08\%$ & $77.21\%$ & $6.93\pm2.06$ \\
\midrule
ResNet & $56.06\%$ & $86.48\%$ & $70.70\%$ & $86.94\%$ & $85.37\%$ & $94.87\%$ & $77.06\%$ & $8.20\pm2.05$ \\
\midrule
SAINT & $57.18\%$ & $88.19\%$ & $76.04\%$ & $88.32\%$ & $85.28\%$ & $97.04\%$ & $75.90\%$ & $6.20\pm2.13$ \\
\midrule
CatBoost & $63.87\%$ & $88.59\%$ & $77.85\%$ & $87.98\%$ & $87.44\%$ & $98.46\%$ & $78.58\%$ & $2.47\pm2.03$ \\
\midrule
LightGBM & $64.39\%$ & $88.43\%$ & $77.27\%$ & $87.43\%$ & $86.90\%$ & $98.38\%$ & $79.81\%$ & $3.27\pm1.82$ \\
\midrule
XGBoost & $64.75\%$ & $88.16\%$ & $76.00\%$ & $87.31\%$ & $87.05\%$ & $98.35\%$ & $79.78\%$ & $4.13\pm1.97$ \\
\midrule
RandomForest & $63.16\%$ & $87.92\%$ & $76.34\%$ & $88.32\%$ & $88.01\%$ & $98.10\%$ & $80.30\%$ & $3.93\pm2.16$ \\
\midrule
GradientBoostingTree & $62.33\%$ & $87.68\%$ & $76.17\%$ & $88.32\%$ & $84.26\%$ & $96.71\%$ & $77.09\%$ & $5.80\pm2.52$ \\
\midrule
\multicolumn{9}{c}{Searched} \\
\midrule
Trompt (ours) & $62.71\%$ & $88.46\%$ & $76.99\%$ & $87.25\%$ & $88.67\%$ & $98.38\%$ & $78.58\%$ & $4.80\pm2.47$ \\
\midrule
FT-Transformer & $58.30\%$ & $88.15\%$ & $76.43\%$ & $89.12\%$ & $85.66\%$ & $98.45\%$ & $76.74\%$ & $6.47\pm2.41$ \\
\midrule
ResNet & $57.03\%$ & $87.54\%$ & $74.63\%$ & $88.23\%$ & $85.87\%$ & $94.86\%$ & $77.41\%$ & $7.73\pm2.50$ \\
\midrule
SAINT & $58.90\%$ & $88.27\%$ & $77.22\%$ & $89.05\%$ & $85.39\%$ & $98.12\%$ & $76.87\%$ & $6.93\pm2.22$ \\
\midrule
CatBoost & $65.07\%$ & $88.54\%$ & $77.95\%$ & $88.02\%$ & $88.83\%$ & $98.47\%$ & $79.89\%$ & $1.93\pm2.36$ \\
\midrule
LightGBM & $65.43\%$ & $88.62\%$ & $77.70\%$ & $88.18\%$ & $87.60\%$ & $98.21\%$ & $79.55\%$ & $3.53\pm1.44$ \\
\midrule
XGBoost & $65.83\%$ & $88.83\%$ & $77.83\%$ & $88.12\%$ & $86.81\%$ & $98.09\%$ & $79.46\%$ & $3.20\pm2.22$ \\
\midrule
RandomForest & $65.04\%$ & $87.80\%$ & $77.27\%$ & $87.95\%$ & $88.45\%$ & $98.20\%$ & $78.96\%$ & $5.33\pm1.88$ \\
\midrule
GradientBoostingTree & $63.04\%$ & $88.22\%$ & $77.17\%$ & $88.32\%$ & $86.68\%$ & $98.06\%$ & $78.56\%$ & $5.07\pm1.77$ \\
\bottomrule
\end{tabular}
\end{small}
\end{center}
\vskip -0.1in
\end{table}

\begin{table}[H]
\caption{The performance of \textbf{large-sized classification} task (\emph{heterogeneous features}).}
\vskip 0.15in
\label{tab:performance-large-het-clf}
\begin{center}
\begin{small}
\begin{tabular}{l | c | c| c}
\toprule
 & \thead{$\mathbb{A}$1} & \thead{$\mathbb{A}$2} & \thead{Ranking} \\
\midrule
\multicolumn{4}{c}{Default} \\
\midrule
Trompt (ours) & $92.76\%$ & $78.36\%$ & $1.50\pm4.36$ \\
\midrule
FT-Transformer & $93.17\%$ & $76.09\%$ & $4.50\pm3.61$ \\
\midrule
ResNet & $89.45\%$ & $76.53\%$ & $6.00\pm2.25$ \\
\midrule
SAINT & $91.23\%$ & $77.31\%$ & $4.50\pm1.73$ \\
\midrule
CatBoost & $88.27\%$ & $78.21\%$ & $4.50\pm1.73$ \\
\midrule
LightGBM & $84.76\%$ & $77.97\%$ & $6.00\pm2.84$ \\
\midrule
XGBoost & $87.81\%$ & $78.22\%$ & $4.50\pm2.65$ \\
\midrule
RandomForest & $90.66\%$ & $77.67\%$ & $4.50\pm1.00$ \\
\midrule
GradientBoostingTree & $79.46\%$ & $75.19\%$ & $9.00\pm4.62$ \\
\midrule
\multicolumn{4}{c}{Searched} \\
\midrule
Trompt (ours) & $93.95\%$ & $78.44\%$ & $3.50\pm3.40$ \\
\midrule
FT-Transformer & $93.61\%$ & $78.92\%$ & $3.50\pm2.36$ \\
\midrule
ResNet & $92.27\%$ & $78.40\%$ & $8.00\pm3.61$ \\
\midrule
SAINT & $92.54\%$ & $77.96\%$ & $8.50\pm4.36$ \\
\midrule
CatBoost & $93.70\%$ & $80.15\%$ & $1.50\pm4.36$ \\
\midrule
LightGBM & $93.25\%$ & $79.75\%$ & $4.00\pm1.32$ \\
\midrule
XGBoost & $93.07\%$ & $79.91\%$ & $4.00\pm2.18$ \\
\midrule
RandomForest & $93.30\%$ & $78.13\%$ & $6.00\pm2.47$ \\
\midrule
GradientBoostingTree & $92.99\%$ & $78.59\%$ & $6.00\pm1.76$ \\
\bottomrule
\end{tabular}
\end{small}
\end{center}
\vskip -0.1in
\end{table}

\begin{table}[H]
\caption{The performance of \textbf{large-sized classification} task (\emph{numerical features only}).}
\vskip 0.15in
\label{tab:performance-large-num-clf}
\begin{center}
\begin{small}
\begin{tabular}{l | c | c| c | c | c}
\toprule
 & \thead{$\mathbb{B}$1} & \thead{$\mathbb{B}$2} & \thead{$\mathbb{B}$3} & \thead{$\mathbb{B}$4} & \thead{Ranking} \\
\midrule
\multicolumn{6}{c}{Default} \\
\midrule
Trompt (ours) & $72.13\%$ & $94.68\%$ & $90.04\%$ & $79.54\%$ & $1.38\pm3.44$ \\
\midrule
FT-Transformer & $69.60\%$ & $94.03\%$ & $89.83\%$ & $75.86\%$ & $6.00\pm2.96$ \\
\midrule
ResNet & $69.88\%$ & $94.09\%$ & $88.01\%$ & $73.58\%$ & $6.00\pm2.78$ \\
\midrule
SAINT & $71.81\%$ & $94.36\%$ & $86.94\%$ & $78.60\%$ & $3.75\pm1.82$ \\
\midrule
CatBoost & $72.61\%$ & $94.32\%$ & $83.77\%$ & $79.54\%$ & $2.88\pm3.01$ \\
\midrule
LightGBM & $72.12\%$ & $93.71\%$ & $80.71\%$ & $78.70\%$ & $5.00\pm2.17$ \\
\midrule
XGBoost & $71.64\%$ & $93.67\%$ & $83.61\%$ & $78.28\%$ & $6.00\pm1.50$ \\
\midrule
RandomForest & $71.58\%$ & $93.08\%$ & $87.67\%$ & $77.97\%$ & $6.00\pm1.80$ \\
\midrule
GradientBoostingTree & $71.03\%$ & $92.25\%$ & $76.98\%$ & $77.18\%$ & $8.00\pm3.29$ \\
\midrule
\multicolumn{6}{c}{Searched} \\
\midrule
Trompt (ours) & $72.86\%$ & $94.36\%$ & $91.27\%$ & $79.88\%$ & $3.25\pm2.97$ \\
\midrule
FT-Transformer & $72.86\%$ & $94.42\%$ & $90.57\%$ & $79.59\%$ & $3.25\pm2.07$ \\
\midrule
ResNet & $72.29\%$ & $94.46\%$ & $89.36\%$ & $78.11\%$ & $6.75\pm3.49$ \\
\midrule
SAINT & $72.65\%$ & $94.45\%$ & $89.53\%$ & $79.30\%$ & $5.50\pm1.67$ \\
\midrule
CatBoost & $72.99\%$ & $94.55\%$ & $90.19\%$ & $79.89\%$ & $1.75\pm3.49$ \\
\midrule
LightGBM & $72.55\%$ & $94.39\%$ & $89.71\%$ & $79.32\%$ & $6.00\pm0.89$ \\
\midrule
XGBoost & $72.81\%$ & $94.40\%$ & $89.32\%$ & $79.67\%$ & $5.25\pm2.30$ \\
\midrule
RandomForest & $71.98\%$ & $93.53\%$ & $90.59\%$ & $78.85\%$ & $7.00\pm3.96$ \\
\midrule
GradientBoostingTree & $72.49\%$ & $94.07\%$ & $89.79\%$ & $79.34\%$ & $6.25\pm1.95$ \\
\bottomrule
\end{tabular}
\end{small}
\end{center}
\vskip -0.1in
\end{table}


\subsubsection{Regression}
\label{sec:performance-individual-reg}

The evaluation results for medium-sized regression datasets are presented in \cref{tab:performance-medium-het-quan-reg-1,tab:performance-medium-het-quan-reg-2} for heterogeneous features, and in \cref{tab:performance-medium-num-quan-reg-1,tab:performance-medium-num-quan-reg-2,tab:performance-medium-num-quan-reg-3} for numerical features only.

For large-sized regression datasets, the results can be found in \cref{tab:performance-large-het-quan-reg} for heterogeneous features, and in \cref{tab:performance-large-num-quan-reg} for numerical features only.

\begin{figure}[H]
    \centering
    \includegraphics[width=.95\linewidth]{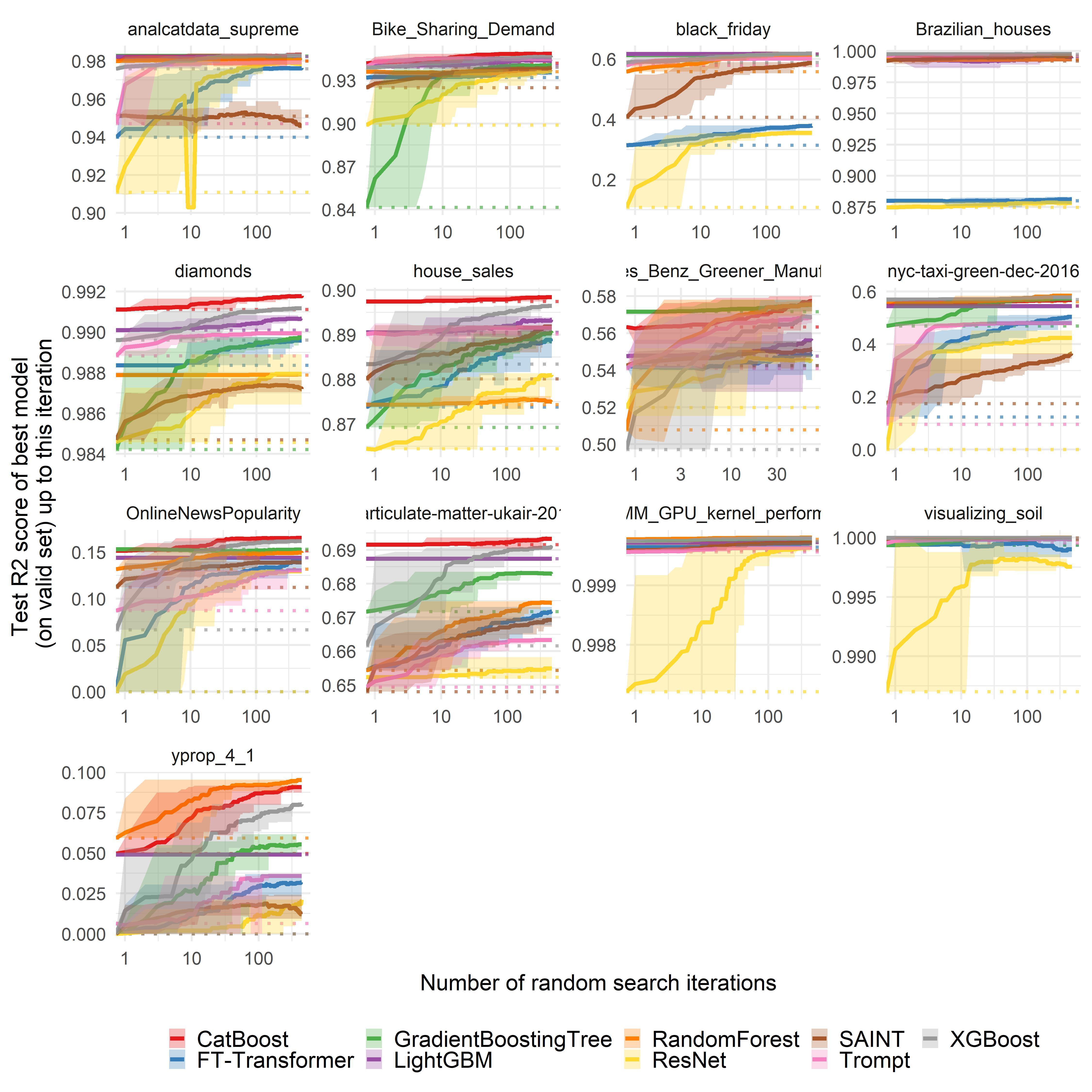}
    \caption{Benchmark on \textbf{\emph{every} medium-sized regression} dataset with \textbf{heterogeneous features}.}
    \label{fig:cat-reg-quan-ind}
\end{figure}

\begin{figure}[H]
    \centering
    \includegraphics[width=.95\linewidth]{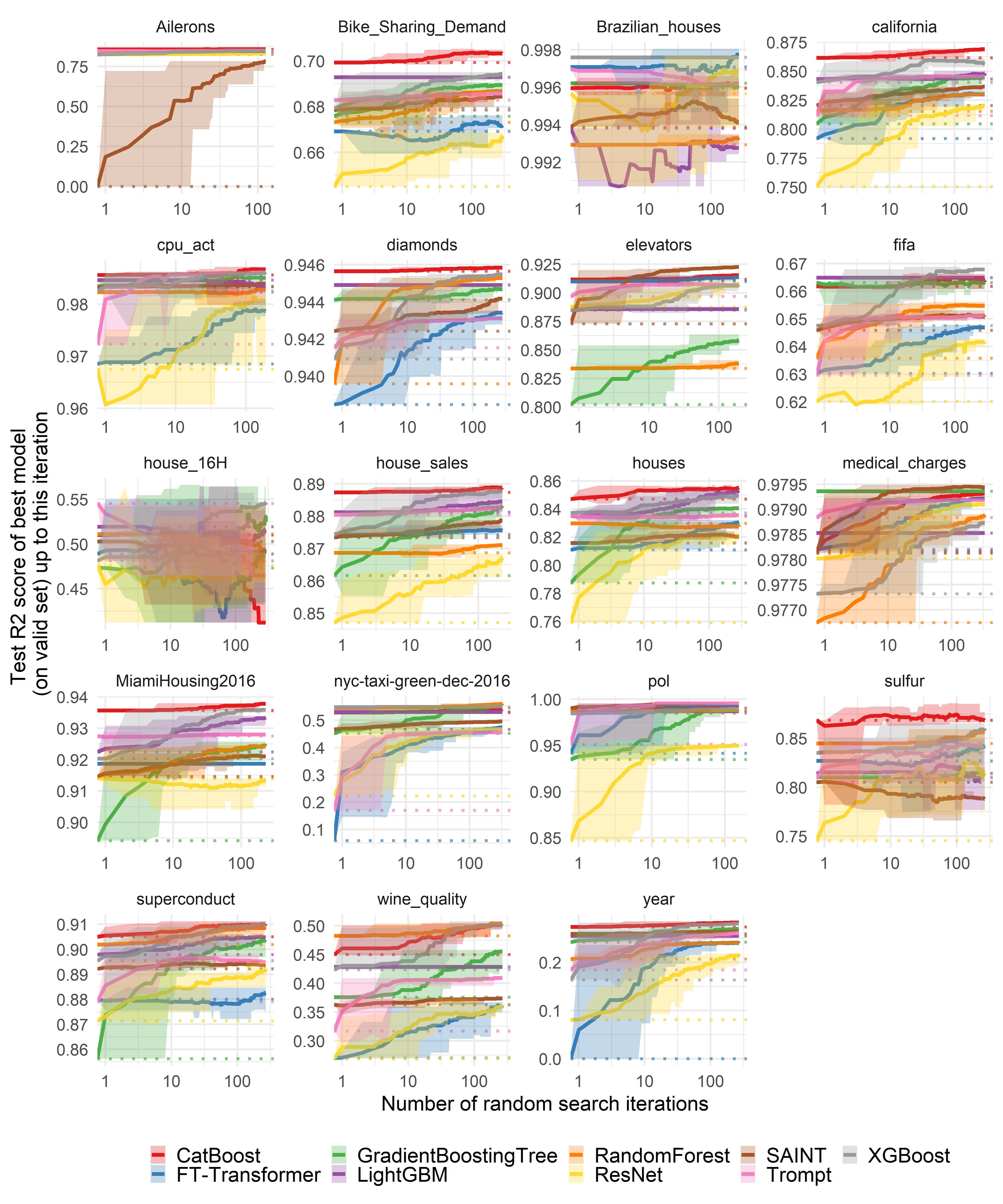}
    \caption{Benchmark on \textbf{\emph{every} medium-sized regression} dataset with \textbf{numerical features only}.}
    \label{fig:num-reg-quan-ind}
\end{figure}

\begin{figure}[H]
    \centering
    \includegraphics[width=.95\linewidth]{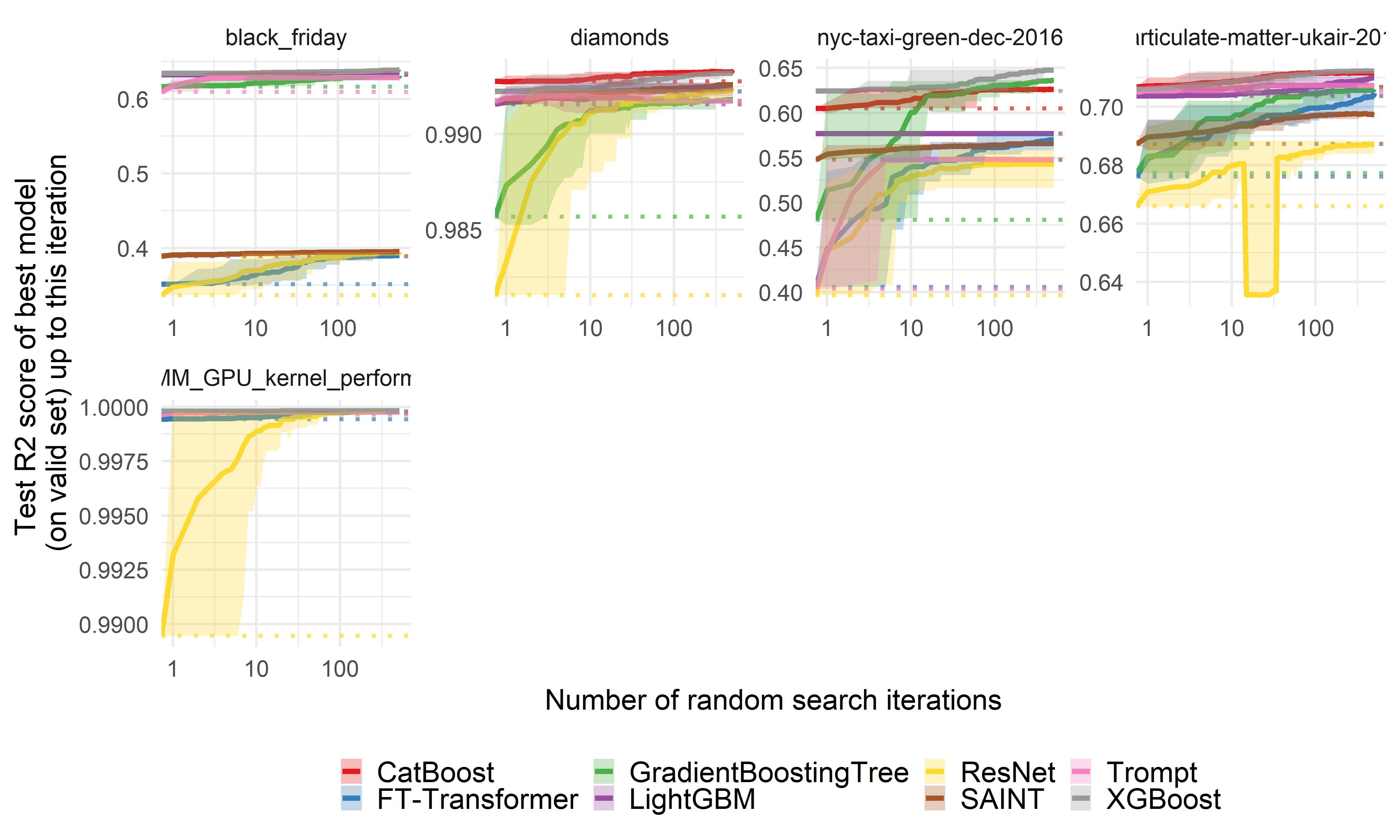}
    \caption{Benchmark on \textbf{\emph{every} large-sized regression} dataset with \textbf{heterogeneous features}.}
    \label{fig:cat-reg-quan-ind-large}
\end{figure}

\begin{figure}[H]
    \centering
    \includegraphics[width=.95\linewidth]{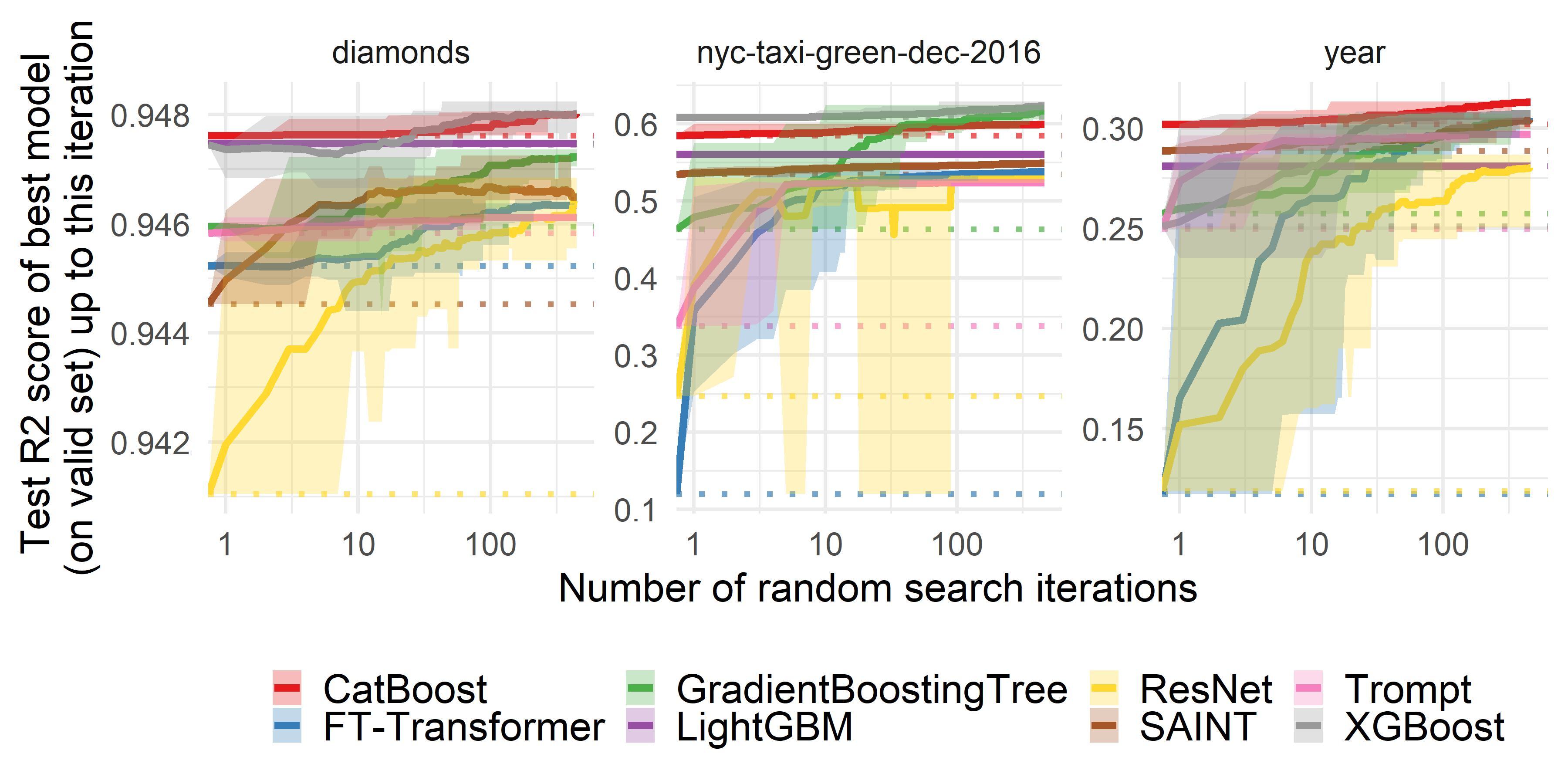}
    \caption{Benchmark on \textbf{\emph{every} large-sized regression} dataset with \textbf{numerical features only}.}
    \label{fig:num-reg-quan-ind-large}
\end{figure}

Furthermore, individual figures illustrating the performance of Trompt on medium-sized regression tasks are provided in \cref{fig:cat-reg-quan-ind} for heterogeneous features, and in \cref{fig:num-reg-quan-ind} for numerical features only.
The individual figures for large-sized tasks can be found in \cref{fig:cat-reg-quan-ind-large} for heterogeneous features, and in \cref{fig:num-reg-quan-ind-large} for numerical features only.

The evaluation results consistently demonstrate that Trompt outperforms state-of-the-art deep neural networks (SAINT and FT-Transformer) on medium-sized regression tasks (refer to \cref{tab:performance-medium-het-quan-reg-1,tab:performance-medium-het-quan-reg-2,tab:performance-medium-num-quan-reg-1,tab:performance-medium-num-quan-reg-2,tab:performance-medium-num-quan-reg-3}).
However, Trompt's performance is slightly inferior to other deep neural networks on large-sized datasets (refer to \cref{tab:performance-large-het-quan-reg,tab:performance-large-num-quan-reg}).
Nevertheless, it is worth noting that the performance of Trompt remains consistently competitive when considering all benchmark results.

\begin{table}[H]
\caption{The performance of \textbf{medium-sized regression} task (\emph{heterogeneous features}) (1).}
\vskip 0.15in
\label{tab:performance-medium-het-quan-reg-1}
\begin{center}
\begin{small}
\begin{tabular}{l | c | c | c | c | c | c | c | c}
\toprule
 & \thead{C1} & \thead{C2} & \thead{C3} & \thead{C4} & \thead{C5} & \thead{C6} & \thead{C7} & \thead{C8} \\
\midrule
\multicolumn{9}{c}{Default} \\
\midrule
Trompt (ours) & $93.93\%$ & $99.63\%$ & $54.09\%$ & $8.71\%$ & $99.96\%$ & $94.70\%$ & $57.94\%$ & $98.88\%$ \\
\midrule
FT-Transformer & $93.21\%$ & $88.00\%$ & $54.24\%$ & $0.00\%$ & $99.96\%$ & $93.99\%$ & $31.46\%$ & $98.84\%$ \\
\midrule
ResNet & $89.90\%$ & $87.47\%$ & $51.99\%$ & $0.00\%$ & $99.72\%$ & $91.09\%$ & $10.79\%$ & $98.46\%$ \\
\midrule
SAINT & $92.50\%$ & $99.20\%$ & $54.25\%$ & $11.23\%$ & $99.96\%$ & $95.10\%$ & $40.72\%$ & $98.47\%$ \\
\midrule
CatBoost & $94.21\%$ & $99.59\%$ & $56.33\%$ & $15.16\%$ & $99.97\%$ & $98.01\%$ & $61.70\%$ & $99.11\%$ \\
\midrule
LightGBM & $94.02\%$ & $99.38\%$ & $54.77\%$ & $14.41\%$ & $99.97\%$ & $98.23\%$ & $61.68\%$ & $99.01\%$ \\
\midrule
XGBoost & $93.93\%$ & $99.76\%$ & $49.71\%$ & $6.64\%$ & $99.97\%$ & $97.59\%$ & $58.93\%$ & $98.96\%$ \\
\midrule
RandomForest & $93.61\%$ & $99.30\%$ & $50.78\%$ & $13.16\%$ & $99.98\%$ & $98.00\%$ & $55.85\%$ & $98.79\%$ \\
\midrule
GradientBoostingTree & $84.15\%$ & $99.62\%$ & $57.17\%$ & $15.30\%$ & $99.97\%$ & $98.27\%$ & $61.34\%$ & $98.42\%$ \\
\midrule
\multicolumn{9}{c}{Searched} \\
\midrule
Trompt (ours) & $94.50\%$ & $99.75\%$ & $56.87\%$ & $13.05\%$ & $99.96\%$ & $97.93\%$ & $60.17\%$ & $98.99\%$ \\
\midrule
FT-Transformer & $93.58\%$ & $88.12\%$ & $54.90\%$ & $14.05\%$ & $99.97\%$ & $97.63\%$ & $37.93\%$ & $98.96\%$ \\
\midrule
ResNet & $93.65\%$ & $87.83\%$ & $54.47\%$ & $12.95\%$ & $99.96\%$ & $97.83\%$ & $35.56\%$ & $98.79\%$ \\
\midrule
SAINT & $93.89\%$ & $99.51\%$ & $55.14\%$ & $13.90\%$ & $99.97\%$ & $94.59\%$ & $58.72\%$ & $98.72\%$ \\
\midrule
CatBoost & $94.87\%$ & $99.60\%$ & $57.74\%$ & $16.54\%$ & $99.97\%$ & $98.33\%$ & $61.79\%$ & $99.18\%$ \\
\midrule
LightGBM & $94.37\%$ & $99.42\%$ & $55.58\%$ & $14.41\%$ & $99.97\%$ & $98.23\%$ & $61.68\%$ & $99.07\%$ \\
\midrule
XGBoost & $94.62\%$ & $99.76\%$ & $56.87\%$ & $16.21\%$ & $99.98\%$ & $98.30\%$ & $61.88\%$ & $99.12\%$ \\
\midrule
RandomForest & $93.79\%$ & $99.34\%$ & $57.55\%$ & $14.94\%$ & $99.98\%$ & $98.07\%$ & $60.91\%$ & $98.79\%$ \\
\midrule
GradientBoostingTree & $94.07\%$ & $99.46\%$ & $57.53\%$ & $15.27\%$ & $99.98\%$ & $98.13\%$ & $61.54\%$ & $98.98\%$ \\
\bottomrule
\end{tabular}
\end{small}
\end{center}
\vskip -0.1in
\end{table}

\begin{table}[H]
\caption{The performance of \textbf{medium-sized regression} task (\emph{heterogeneous features}) (2).}
\vskip 0.15in
\label{tab:performance-medium-het-quan-reg-2}
\begin{center}
\begin{small}
\begin{tabular}{l | c | c | c | c | c | c | c}
\toprule
 & \thead{C9} & \thead{C10} & \thead{C11} & \thead{C12} & \thead{C13} & \thead{Ranking} \\
\midrule
\multicolumn{7}{c}{Default} \\
\midrule
Trompt (ours) & $89.02\%$ & $9.61\%$ & $64.94\%$ & $99.95\%$ & $0.64\%$ & $5.38\pm2.02$ \\
\midrule
FT-Transformer & $87.38\%$ & $12.38\%$ & $65.43\%$ & $99.94\%$ & $0.00\%$ & $6.88\pm1.74$ \\
\midrule
ResNet & $86.45\%$ & $0.00\%$ & $65.23\%$ & $98.70\%$ & $0.00\%$ & $8.35\pm2.13$ \\
\midrule
SAINT & $88.01\%$ & $17.48\%$ & $64.80\%$ & $99.98\%$ & $0.00\%$ & $6.31\pm1.54$ \\
\midrule
CatBoost & $89.75\%$ & $54.63\%$ & $69.16\%$ & $99.99\%$ & $4.97\%$ & $2.15\pm2.13$ \\
\midrule
LightGBM & $89.05\%$ & $54.48\%$ & $68.74\%$ & $99.99\%$ & $4.91\%$ & $3.00\pm1.74$ \\
\midrule
XGBoost & $88.34\%$ & $56.99\%$ & $66.16\%$ & $100.00\%$ & $0.00\%$ & $4.08\pm2.46$ \\
\midrule
RandomForest & $87.44\%$ & $56.18\%$ & $65.44\%$ & $100.00\%$ & $5.92\%$ & $4.23\pm2.27$ \\
\midrule
GradientBoostingTree & $86.93\%$ & $46.90\%$ & $67.17\%$ & $99.94\%$ & $0.00\%$ & $4.62\pm2.92$ \\
\midrule
\multicolumn{7}{c}{Searched} \\
\midrule
Trompt (ours) & $89.16\%$ & $48.04\%$ & $66.33\%$ & $99.99\%$ & $3.59\%$ & $5.77\pm1.98$ \\
\midrule
FT-Transformer & $88.85\%$ & $50.44\%$ & $67.18\%$ & $99.90\%$ & $3.18\%$ & $7.23\pm1.70$ \\
\midrule
ResNet & $88.10\%$ & $42.42\%$ & $65.50\%$ & $99.76\%$ & $2.11\%$ & $8.31\pm2.08$ \\
\midrule
SAINT & $89.18\%$ & $36.42\%$ & $66.93\%$ & $99.99\%$ & $1.21\%$ & $7.00\pm1.98$ \\
\midrule
CatBoost & $89.84\%$ & $56.79\%$ & $69.33\%$ & $100.00\%$ & $9.08\%$ & $2.00\pm2.24$ \\
\midrule
LightGBM & $89.33\%$ & $54.48\%$ & $68.74\%$ & $100.00\%$ & $4.91\%$ & $4.31\pm1.18$ \\
\midrule
XGBoost & $89.65\%$ & $57.82\%$ & $69.08\%$ & $100.00\%$ & $8.01\%$ & $2.15\pm1.74$ \\
\midrule
RandomForest & $87.50\%$ & $58.48\%$ & $67.44\%$ & $100.00\%$ & $9.52\%$ & $4.31\pm2.80$ \\
\midrule
GradientBoostingTree & $89.05\%$ & $57.29\%$ & $68.30\%$ & $100.00\%$ & $5.54\%$ & $3.92\pm1.35$ \\
\bottomrule
\end{tabular}
\end{small}
\end{center}
\vskip -0.1in
\end{table}

\begin{table}[H]
\caption{The performance of \textbf{medium-sized regression} task (\emph{numerical features only}) (1).}
\vskip 0.15in
\label{tab:performance-medium-num-quan-reg-1}
\begin{center}
\begin{small}
\begin{tabular}{l | c | c | c | c | c | c | c}
\toprule
 & \thead{D1} & \thead{D2} & \thead{D3} & \thead{D4} & \thead{D5} & \thead{D6} & \thead{D7}\\
\midrule
\multicolumn{8}{c}{Default} \\
\midrule
Trompt (ours) & $84.80\%$ & $68.29\%$ & $99.70\%$ & $92.75\%$ & $81.17\%$ & $97.23\%$ & $94.15\%$ \\
\midrule
FT-Transformer & $83.80\%$ & $66.92\%$ & $99.71\%$ & $91.87\%$ & $79.20\%$ & $96.85\%$ & $93.85\%$ \\
\midrule
ResNet & $82.54\%$ & $64.52\%$ & $99.57\%$ & $91.41\%$ & $75.06\%$ & $96.75\%$ & $nan$ \\
\midrule
SAINT & $0.00\%$ & $67.85\%$ & $99.39\%$ & $91.46\%$ & $82.04\%$ & $98.33\%$ & $94.24\%$ \\
\midrule
CatBoost & $85.76\%$ & $69.93\%$ & $99.60\%$ & $93.56\%$ & $86.16\%$ & $98.56\%$ & $94.57\%$ \\
\midrule
LightGBM & $84.68\%$ & $69.28\%$ & $99.38\%$ & $92.25\%$ & $84.33\%$ & $98.46\%$ & $94.49\%$ \\
\midrule
XGBoost & $82.58\%$ & $67.93\%$ & $99.76\%$ & $92.03\%$ & $84.04\%$ & $98.25\%$ & $94.09\%$ \\
\midrule
RandomForest & $83.71\%$ & $67.32\%$ & $99.29\%$ & $91.41\%$ & $81.54\%$ & $98.23\%$ & $93.96\%$ \\
\midrule
GradientBoostingTree & $83.95\%$ & $67.58\%$ & $99.62\%$ & $89.42\%$ & $80.46\%$ & $98.34\%$ & $94.41\%$ \\
\midrule
\multicolumn{8}{c}{Searched} \\
\midrule
Trompt (ours) & $85.08\%$ & $68.57\%$ & $99.62\%$ & $92.80\%$ & $84.53\%$ & $98.61\%$ & $94.31\%$ \\
\midrule
FT-Transformer & $83.90\%$ & $67.17\%$ & $99.77\%$ & $91.87\%$ & $83.00\%$ & $97.87\%$ & $94.34\%$ \\
\midrule
ResNet & $83.21\%$ & $66.71\%$ & $99.69\%$ & $91.36\%$ & $82.03\%$ & $98.07\%$ & $nan$ \\
\midrule
SAINT & $78.31\%$ & $68.44\%$ & $99.41\%$ & $92.10\%$ & $83.67\%$ & $98.39\%$ & $94.42\%$ \\
\midrule
CatBoost & $85.92\%$ & $70.31\%$ & $99.62\%$ & $93.78\%$ & $86.90\%$ & $98.67\%$ & $94.59\%$ \\
\midrule
LightGBM & $84.68\%$ & $69.28\%$ & $99.28\%$ & $93.33\%$ & $84.80\%$ & $98.31\%$ & $94.49\%$ \\
\midrule
XGBoost & $84.58\%$ & $69.43\%$ & $99.76\%$ & $93.59\%$ & $85.64\%$ & $98.61\%$ & $94.55\%$ \\
\midrule
RandomForest & $83.75\%$ & $68.69\%$ & $99.33\%$ & $92.42\%$ & $83.02\%$ & $98.28\%$ & $94.53\%$ \\
\midrule
GradientBoostingTree & $84.25\%$ & $68.94\%$ & $99.60\%$ & $92.43\%$ & $84.48\%$ & $98.51\%$ & $94.47\%$ \\
\bottomrule
\end{tabular}
\end{small}
\end{center}
\vskip -0.1in
\end{table}

\begin{table}[H]
\caption{The performance of \textbf{medium-sized regression} task (\emph{numerical features only}) (2).}
\vskip 0.15in
\label{tab:performance-medium-num-quan-reg-2}
\begin{center}
\begin{small}
\begin{tabular}{l | c | c | c | c | c | c | c}
\toprule
 & \thead{D8} & \thead{D9} & \thead{D10} & \thead{D11} & \thead{D12} & \thead{D13} & \thead{D14}  \\
\midrule
\multicolumn{8}{c}{Default} \\
\midrule
Trompt (ours) & $89.69\%$ & $62.96\%$ & $54.53\%$ & $88.04\%$ & $83.52\%$ & $97.88\%$ & $16.99\%$ \\
\midrule
FT-Transformer & $91.01\%$ & $63.03\%$ & $48.90\%$ & $87.42\%$ & $81.10\%$ & $97.82\%$ & $5.86\%$ \\
\midrule
ResNet & $88.77\%$ & $62.01\%$ & $47.62\%$ & $84.71\%$ & $75.92\%$ & $97.80\%$ & $22.34\%$ \\
\midrule
SAINT & $87.30\%$ & $64.59\%$ & $50.30\%$ & $87.34\%$ & $81.59\%$ & $97.81\%$ & $46.65\%$ \\
\midrule
CatBoost & $91.17\%$ & $66.18\%$ & $51.01\%$ & $88.73\%$ & $84.72\%$ & $97.82\%$ & $52.91\%$ \\
\midrule
LightGBM & $88.59\%$ & $66.49\%$ & $51.95\%$ & $88.12\%$ & $83.51\%$ & $97.85\%$ & $53.06\%$ \\
\midrule
XGBoost & $88.48\%$ & $64.75\%$ & $48.14\%$ & $87.43\%$ & $83.74\%$ & $97.73\%$ & $54.87\%$ \\
\midrule
RandomForest & $83.37\%$ & $63.58\%$ & $51.12\%$ & $86.87\%$ & $82.99\%$ & $97.67\%$ & $54.54\%$ \\
\midrule
GradientBoostingTree & $80.22\%$ & $66.31\%$ & $47.33\%$ & $86.16\%$ & $78.74\%$ & $97.94\%$ & $45.15\%$ \\
\midrule
\multicolumn{8}{c}{Searched} \\
\midrule
Trompt (ours) & $90.69\%$ & $65.13\%$ & $46.50\%$ & $88.27\%$ & $83.57\%$ & $97.92\%$ & $45.57\%$ \\
\midrule
FT-Transformer & $91.37\%$ & $64.69\%$ & $48.67\%$ & $87.56\%$ & $83.05\%$ & $97.92\%$ & $47.43\%$ \\
\midrule
ResNet & $90.82\%$ & $64.19\%$ & $48.16\%$ & $86.72\%$ & $82.08\%$ & $97.91\%$ & $46.78\%$ \\
\midrule
SAINT & $92.27\%$ & $65.06\%$ & $49.40\%$ & $87.87\%$ & $82.03\%$ & $97.94\%$ & $49.58\%$ \\
\midrule
CatBoost & $91.56\%$ & $66.39\%$ & $41.22\%$ & $88.89\%$ & $85.53\%$ & $97.93\%$ & $54.06\%$ \\
\midrule
LightGBM & $88.59\%$ & $66.49\%$ & $51.60\%$ & $88.45\%$ & $85.33\%$ & $97.85\%$ & $53.06\%$ \\
\midrule
XGBoost & $90.67\%$ & $66.79\%$ & $54.63\%$ & $88.76\%$ & $84.95\%$ & $97.87\%$ & $55.23\%$ \\
\midrule
RandomForest & $83.82\%$ & $65.47\%$ & $49.15\%$ & $87.10\%$ & $82.77\%$ & $97.89\%$ & $56.04\%$ \\
\midrule
GradientBoostingTree & $85.84\%$ & $66.32\%$ & $52.49\%$ & $88.32\%$ & $84.07\%$ & $97.94\%$ & $55.21\%$ \\
\bottomrule
\end{tabular}
\end{small}
\end{center}
\vskip -0.1in
\end{table}

\begin{table}[H]
\caption{The performance of \textbf{medium-sized regression} task (\emph{numerical features only}) (3).}
\vskip 0.15in
\label{tab:performance-medium-num-quan-reg-3}
\begin{center}
\begin{small}
\begin{tabular}{l | c | c | c | c | c | c}
\toprule
 & \thead{D15} & \thead{D16} & \thead{D17} & \thead{D18} & \thead{D19} & \thead{Ranking}\\
\midrule
\multicolumn{7}{c}{Default} \\
\midrule
Trompt (ours) & $95.13\%$ & $80.96\%$ & $87.91\%$ & $31.68\%$ & $18.41\%$ & $4.68\pm2.29$ \\
\midrule
FT-Transformer & $94.16\%$ & $82.70\%$ & $88.01\%$ & $26.98\%$ & $0.00\%$ & $6.21\pm2.29$ \\
\midrule
ResNet & $84.68\%$ & $74.54\%$ & $87.14\%$ & $26.86\%$ & $8.13\%$ & $8.06\pm2.08$ \\
\midrule
SAINT & $99.04\%$ & $80.52\%$ & $89.22\%$ & $36.25\%$ & $25.92\%$ & $5.32\pm1.86$ \\
\midrule
CatBoost & $98.63\%$ & $86.85\%$ & $90.51\%$ & $45.00\%$ & $27.34\%$ & $2.05\pm2.16$ \\
\midrule
LightGBM & $98.70\%$ & $81.43\%$ & $89.79\%$ & $42.86\%$ & $25.50\%$ & $3.05\pm1.89$ \\
\midrule
XGBoost & $98.50\%$ & $83.49\%$ & $89.55\%$ & $42.37\%$ & $16.33\%$ & $4.47\pm2.04$ \\
\midrule
RandomForest & $98.67\%$ & $84.47\%$ & $90.20\%$ & $48.28\%$ & $20.69\%$ & $5.26\pm2.40$ \\
\midrule
GradientBoostingTree & $93.49\%$ & $81.04\%$ & $85.62\%$ & $37.57\%$ & $24.21\%$ & $5.84\pm2.60$ \\
\midrule
\multicolumn{7}{c}{Searched} \\
\midrule
Trompt (ours) & $99.58\%$ & $84.15\%$ & $89.49\%$ & $40.91\%$ & $26.03\%$ & $5.11\pm1.97$ \\
\midrule
FT-Transformer & $99.44\%$ & $84.26\%$ & $88.26\%$ & $36.07\%$ & $23.96\%$ & $6.37\pm2.48$ \\
\midrule
ResNet & $94.99\%$ & $81.45\%$ & $89.22\%$ & $36.11\%$ & $21.73\%$ & $7.61\pm2.31$ \\
\midrule
SAINT & $99.56\%$ & $78.81\%$ & $89.37\%$ & $37.38\%$ & $26.45\%$ & $5.79\pm2.46$ \\
\midrule
CatBoost & $99.24\%$ & $86.84\%$ & $90.94\%$ & $50.11\%$ & $28.26\%$ & $2.26\pm2.46$ \\
\midrule
LightGBM & $98.70\%$ & $81.31\%$ & $90.48\%$ & $42.86\%$ & $25.50\%$ & $4.84\pm2.25$ \\
\midrule
XGBoost & $98.97\%$ & $86.03\%$ & $91.02\%$ & $50.06\%$ & $28.04\%$ & $2.79\pm2.11$ \\
\midrule
RandomForest & $98.87\%$ & $85.64\%$ & $90.89\%$ & $50.43\%$ & $24.09\%$ & $5.58\pm2.33$ \\
\midrule
GradientBoostingTree & $98.91\%$ & $81.31\%$ & $90.36\%$ & $45.55\%$ & $26.94\%$ & $4.58\pm1.69$ \\
\bottomrule
\end{tabular}
\end{small}
\end{center}
\vskip -0.1in
\end{table}

\begin{table}[H]
\caption{The performance of \textbf{large-sized regression} task (\emph{heterogeneous features}).}
\vskip 0.15in
\label{tab:performance-large-het-quan-reg}
\begin{center}
\begin{small}
\begin{tabular}{l | c | c | c | c | c | c}
\toprule
 & \thead{$\mathbb{C}$1} & \thead{$\mathbb{C}$2} & \thead{$\mathbb{C}$3} & \thead{$\mathbb{C}$4} & \thead{$\mathbb{C}$5} & \thead{Ranking} \\
\midrule
\multicolumn{7}{c}{Default} \\
\midrule
Trompt (ours) & $99.96\%$ & $60.97\%$ & $99.17\%$ & $40.35\%$ & $70.48\%$ & $5.20\pm1.50$ \\
\midrule
FT-Transformer & $99.94\%$ & $35.14\%$ & $99.23\%$ & $40.61\%$ & $67.61\%$ & $5.80\pm2.48$ \\
\midrule
ResNet & $98.95\%$ & $33.70\%$ & $98.16\%$ & $39.71\%$ & $66.60\%$ & $8.00\pm2.86$ \\
\midrule
SAINT & $99.97\%$ & $38.91\%$ & $99.18\%$ & $54.80\%$ & $68.74\%$ & $4.80\pm0.75$ \\
\midrule
CatBoost & $99.98\%$ & $63.32\%$ & $99.28\%$ & $60.50\%$ & $70.68\%$ & $1.80\pm2.25$ \\
\midrule
LightGBM & $99.98\%$ & $63.24\%$ & $99.16\%$ & $57.69\%$ & $70.37\%$ & $3.60\pm1.67$ \\
\midrule
XGBoost & $99.98\%$ & $63.45\%$ & $99.22\%$ & $62.44\%$ & $70.60\%$ & $1.60\pm2.73$ \\
\midrule
RandomForest & $-$ & $-$ & $-$ & $-$ & $-$ & $-$ \\
\midrule
GradientBoostingTree & $99.98\%$ & $61.65\%$ & $98.57\%$ & $48.09\%$ & $67.73\%$ & $5.20\pm1.36$ \\
\midrule
\multicolumn{7}{c}{Searched} \\
\midrule
Trompt (ours) & $99.98\%$ & $62.86\%$ & $99.18\%$ & $54.79\%$ & $70.73\%$ & $6.20\pm2.26$ \\
\midrule
FT-Transformer & $99.98\%$ & $39.00\%$ & $99.26\%$ & $57.02\%$ & $70.45\%$ & $5.80\pm1.75$ \\
\midrule
ResNet & $99.98\%$ & $39.38\%$ & $99.23\%$ & $54.30\%$ & $68.71\%$ & $6.40\pm2.88$ \\
\midrule
SAINT & $99.98\%$ & $39.53\%$ & $99.26\%$ & $56.58\%$ & $69.73\%$ & $5.20\pm1.55$ \\
\midrule
CatBoost & $99.98\%$ & $63.62\%$ & $99.33\%$ & $62.64\%$ & $71.17\%$ & $2.60\pm2.25$ \\
\midrule
LightGBM & $99.98\%$ & $63.24\%$ & $99.24\%$ & $57.69\%$ & $70.99\%$ & $4.60\pm1.86$ \\
\midrule
XGBoost & $99.98\%$ & $63.90\%$ & $99.32\%$ & $64.79\%$ & $71.22\%$ & $1.20\pm2.80$ \\
\midrule
RandomForest & $-$ & $-$ & $-$ & $-$ & $-$ & $-$ \\
\midrule
GradientBoostingTree & $99.98\%$ & $63.06\%$ & $99.18\%$ & $63.62\%$ & $70.58\%$ & $4.00\pm2.07$ \\
\bottomrule
\end{tabular}
\end{small}
\end{center}
\vskip -0.1in
\end{table}

\begin{table}[H]
\caption{The performance of \textbf{large-sized regression} task (\emph{numerical features only}).}
\vskip 0.15in
\label{tab:performance-large-num-quan-reg}
\begin{center}
\begin{small}
\begin{tabular}{l | c | c | c | c | c}
\toprule
 & \thead{$\mathbb{D}$1} & \thead{$\mathbb{D}$2} & \thead{$\mathbb{D}$3} & \thead{Ranking}\\
\midrule
\multicolumn{5}{c}{Default} \\
\midrule
Trompt (ours) & $94.58\%$ & $33.79\%$ & $24.98\%$ & $5.67\pm1.41$ \\
\midrule
FT-Transformer & $94.52\%$ & $11.98\%$ & $11.72\%$ & $7.33\pm3.07$ \\
\midrule
ResNet & $94.10\%$ & $24.69\%$ & $11.88\%$ & $7.33\pm2.95$ \\
\midrule
SAINT & $94.45\%$ & $53.44\%$ & $28.87\%$ & $4.33\pm2.06$ \\
\midrule
CatBoost & $94.76\%$ & $58.47\%$ & $30.20\%$ & $1.33\pm3.37$ \\
\midrule
LightGBM & $94.75\%$ & $56.07\%$ & $28.10\%$ & $2.67\pm2.22$ \\
\midrule
XGBoost & $94.74\%$ & $60.87\%$ & $25.12\%$ & $3.00\pm2.22$ \\
\midrule
RandomForest & $-$ & $-$ & $-$ & $-$ \\
\midrule
GradientBoostingTree & $94.59\%$ & $46.35\%$ & $25.74\%$ & $4.33\pm0.48$ \\
\midrule
\multicolumn{5}{c}{Searched} \\
\midrule
Trompt (ours) & $94.61\%$ & $52.42\%$ & $29.71\%$ & $7.33\pm3.30$ \\
\midrule
FT-Transformer & $94.63\%$ & $53.82\%$ & $30.51\%$ & $5.67\pm1.83$ \\
\midrule
ResNet & $94.64\%$ & $52.84\%$ & $28.01\%$ & $7.00\pm2.63$ \\
\midrule
SAINT & $94.65\%$ & $54.94\%$ & $30.46\%$ & $5.00\pm0.50$ \\
\midrule
CatBoost & $94.80\%$ & $59.97\%$ & $31.30\%$ & $2.00\pm2.63$ \\
\midrule
LightGBM & $94.75\%$ & $56.07\%$ & $28.10\%$ & $4.67\pm1.71$ \\
\midrule
XGBoost & $94.80\%$ & $62.36\%$ & $30.75\%$ & $1.33\pm3.37$ \\
\midrule
RandomForest & $-$ & $-$ & $-$ & $-$ \\
\midrule
GradientBoostingTree & $94.72\%$ & $61.72\%$ & $30.73\%$ & $3.00\pm1.71$ \\
\bottomrule
\end{tabular}
\end{small}
\end{center}
\vskip -0.1in
\end{table}

\subsection{Datasets chosen by FT-Transformer}
\label{sec:more-eval-ft}

In this section, we further investigate the performance of Trompt on datasets selected by FT-Transformer \cite{gorishniy2021revisiting}, which encompass different domains, task types, and sizes.
To ensure a fair comparison, we adjust the model sizes of Trompt to match those of FT-Transformer by reducing the dimensions of its hidden layers.

It's important to note that due to limited computing resources, Trompt did not undergo hyperparameter search.
Instead, we obtained the performance of FT-Transformer from its original paper.
In terms of the learning strategy, Trompt was trained for 100 epochs, and the performance was evaluated using the checkpoint at the 100th epoch.
This approach was adopted as we observed that the datasets chosen by FT-Transformer are often large, making overfitting less likely.

As shown in \cref{tab:performance-more-eval-ft}, Trompt generally achieves comparable or slightly inferior performance when compared to the default hyperparameter settings of FT-Transformer on the datasets specifically chosen by FT-Transformer.
It is important to note that the reported performance is an average result based on three random seeds.

\begin{table}[H]
\caption{The performance on datasets chosen by FT-Transformer.}
\vskip 0.15in
\label{tab:performance-more-eval-ft}
\begin{center}
\begin{small}
\begin{tabular}{c | c | c| c | c | c | c}
\toprule
\thead{Dataset} & \thead{Metric} & \thead{Trompt (ours)} & \thead{FT (Default)} & \thead{FT (Tune)} & \thead{\#Parameters (Trompt)} & \thead{\#Parameters (FT)} \\
\midrule
CA & RMSE & $0.474$ & $0.469$ & $0.459$ & $850,852$ & $894,913$ \\
\midrule
AD & Acc. & $0.8629$ & $0.857$ & $0.859$ & $863,509$ & $915,458$ \\
\midrule
HE & Acc. & $0.3690$ & $0.381$ & $0.391$ & $873,883$ & $921,316$ \\
\midrule
JA & Acc. & $0.7269$ & $0.725$ & $0.732$ & $876,079$ & $913,156$ \\
\midrule
HI & Acc. & $0.7279$ & $0.725$ & $0.729$ & $861,781$ & $902,786$ \\
\midrule
AL & Acc. & $0.9317$ & $0.953$ & $0.96$ & $1,044,523$ & $1,133,800$ \\
\midrule
EP & Acc. & $0.8932$ & $0.8959$ & $0.8982$ & $1,638,931$ & $1,659,841$ \\
\midrule
YE & RMSE & $8.8218$ & $8.889$ & $8.855$ & $895,132$ & $926,401$ \\
\midrule
CO & Acc. & $0.9048$ & $0.967$ & $0.970$ & $876,466$ & $913,735$ \\
\midrule
YA & RMSE & $0.7537$ & $0.756$ & $0.756$ & $1,223,992$ & $1,160,257$ \\
\midrule
MI & RMSE & $0.7468$ & $0.747$ & $0.746$ & $919,972$ & $944,065$ \\
\bottomrule
\end{tabular}
\end{small}
\end{center}
\vskip -0.1in
\end{table}

\subsection{Datasets chosen by SAINT}
\label{sec:more-eval-saint}

In this section, we conducted further evaluation of Trompt on datasets selected by SAINT \cite{somepalli2021saint}, which cover various domains, task types, and sizes.
To ensure fair comparison, we adjusted the model sizes of Trompt to match those of SAINT by reducing the dimensions of its hidden layers.

It is important to note that due to limited computing resources, Trompt did not undergo hyperparameter search.
Instead, we obtained the performances of SAINT from its original paper. In terms of the learning strategy, Trompt was trained for 100 epochs, and the performance was evaluated using the checkpoint with the lowest validation loss.
This approach was adopted as we observed that some datasets chosen by SAINT are often small, and models are more prone to overfitting.

As shown in \cref{tab:performance-more-eval-saint}, Trompt achieves comparable performance to SAINT on the datasets specifically chosen by SAINT.
It is worth mentioning that the reported performance is based on a single random seed.

\begin{table}[H]
\caption{The performance on datasets chosen by SAINT.}
\vskip 0.15in
\label{tab:performance-more-eval-saint}
\begin{center}
\begin{small}
\begin{tabular}{c | c | c| c | c | c}
\toprule
\thead{OpenML ID} & \thead{Metric} & \thead{Trompt (ours)} & \thead{SAINT} & \thead{\#Parameters (Trompt)} & \thead{\#Parameters (SAINT)} \\
\midrule
31 & AUC & $0.8265$ & $0.7900$ & $7,578,619$ & $8,233,739$ \\
\midrule
1017 & AUC & $0.8933$ & $0.8430$ & $39,521,539$ & $84,093,615$ \\
\midrule
44 & AUC & $0.9835$ & $0.9910$ & $38,675,971$ & $58,399,221$ \\
\midrule
1111 & AUC & $0.8114$ & $0.8080$ & $60,085,567$ & $61,716,420$ \\
\midrule
1487 & AUC & $0.9230$ & $0.9190$ & $38,733,571$ & $91,681,626$ \\
\midrule
1494 & AUC & $0.9258$ & $0.9370$ & $29,659,027$ & $31,136,311$ \\
\midrule
1590 & AUC & $0.9165$ & $0.9210$ & $3,945,643$ & $4,420,452$ \\
\midrule
4134 & AUC & $0.8419$ & $0.8530$ & $45,276,931$ & $3,296,373,186$ \\
\midrule
42178 & AUC & $0.8454$ & $0.8570$ & $65,51,239$ & $7,500,881$ \\
\midrule
42733 & AUC & $0.6820$ & $0.6760$ & $29,743,735$ & $30,585,898$ \\
\midrule
1596 & Acc. & $0.960281$ & $0.9460$ & $38,665,096$ & $52,507,599$ \\
\midrule
4541 & Acc. & $0.6071$ & $0.6060$ & $40,478,596$ & $44,131,471$ \\
\midrule
40664 & Acc. & $0.9913$ & $1.0000$ & $8,664,841$ & $8,960,176$ \\
\midrule
40685 & Acc. & $0.9997$ & $0.9990$ & $1,969,996$ & $2,142,668$ \\
\midrule
188 & Acc. & $0.6622$ & $0.6800$ & $6,569,098$ & $7,547,934$ \\
\midrule
40687 & Acc. & $0.7463$ & $0.7350$ & $3,203,035$ & $3,381,200$ \\
\midrule
40975 & Acc. & $0.9884$ & $0.9970$ & $1,037,761$ & $1,147,867$ \\
\midrule
41166 & Acc. & $0.7064$ & $0.7010$ & $34,490,755$ & $35,807,954$ \\
\midrule
41169 & Acc. & $0.3839$ & $0.3770$ & $13,802,953$ & $14,361,949$ \\
\midrule
42734 & Acc. & $0.7495$ & $0.7520$ & $8,922,568$ & $9,205,592$ \\
\midrule
422 & RMSE & $0.0272$ & $0.0270$ & $39,478,402$ & $76,649,015$ \\
\midrule
541 & RMSE & $7.9160$ & $11.6610$ & $684,082$ & $897,840$ \\
\midrule
42563 & RMSE & $23094.4130$ & $33112.3870$ & $38,900,098$ & $109,678,283$ \\
\midrule
42571 & RMSE & $1918.3982$ & $1953.3910$ & $17,456,806$ & $19,048,879$ \\
\midrule
42705 & RMSE & $8.9351$ & $10.2820$ & $38,840,962$ & $173,809,579$ \\
\midrule
42724 & RMSE & $12144.9121$ & $11577.6780$ & $38,683,522$ & $62,405,052$ \\
\midrule
42726 & RMSE & $2.0735$ & $2.1130$ & $1,466,218$ & $1,775,189$ \\
\midrule
42727 & RMSE & $0.1502$ & $0.1450$ & $35,502,610$ & $37,517,460$ \\
\midrule
42728 & RMSE & $16.3780$ & $12.5780$ & $2,049,022$ & $2,234,102$ \\
\midrule
42729 & RMSE & $1.9436$ & $1.8820$ & $6,682,150$ & $6,922,958$ \\
\bottomrule
\end{tabular}
\end{small}
\end{center}
\vskip -0.1in
\end{table}

\section{Settings of Ablation Study}
\label{sec:ablation-setting}
In the ablation study, we explored different approaches to normalize the regression targets for regression tasks.
Specifically, we compared standardization (mean subtraction and scaling) with the quantile transformation used in Grinsztajn45 \cite{grinsztajn2022why}, which relies on the Scikit-learn library's quantile transformation \cite{sklearn2023quantile}.

Based on our experiments, we found that standardization generally leads to better performance compared to quantile transformation, as demonstrated in \cref{tab:target-transform}. To ensure a fair comparison, all results in \cref{sec:eval} were obtained using the configurations specified in Grinsztajn45.

In the ablation study, we simply selected the better normalization approach based on its performance.
We provide these details here to explain the performance differences observed in the regression tasks discussed in \cref{sec:eval}, as well as those in \cref{sec:ablation} and \cref{sec:more-ablation}.

\begin{table}[H]
\caption{Average r2-score of Trompt using different target normalizations on Grinsztajn45 regression tasks.}
\vskip 0.15in
\label{tab:target-transform}
\begin{center}
\begin{small}
\begin{tabular}{c | c}
\toprule
\thead{Target Normalization} & \thead{r2-score} \\
\midrule
Quantile Transformation& $70.55\%$ \\
\midrule
Standardization & $74.15\%$ \\
\bottomrule
\end{tabular}
\end{small}
\end{center}
\vskip -0.1in
\end{table}

\section{More Ablation Studies}
\label{sec:more-ablation}
In \cref{sec:more-ablation-params}, we present additional ablation studies focusing on different values of various hyperparameters. We investigate the impact of varying these hyperparameters on the performance of Trompt.

Furthermore, in \cref{sec:more-ablation-archi}, we delve into the necessity of key components in the architecture of Trompt. We conduct ablation experiments to examine the effect of removing or modifying these components on the overall performance of Trompt.

These additional ablation studies aim to provide further insights into the role and importance of different hyperparameters and architectural components in Trompt.

\subsection{Hyperparameters}
\label{sec:more-ablation-params}
\textbf{Ablations on the size of hidden dimension.}

The hidden dimension ($d$) parameter in Trompt plays a crucial role in configuring various parts of the model, such as the size of dense layers and embeddings. To evaluate the impact of different values of $d$, we conducted experiments using Trompt with six different values of $d$.

The results presented in \cref{tab:ablation-dimension} demonstrate that Trompt achieves good performance when an adequate amount of hidden dimension is used, particularly when $d$ is larger than 32. This suggests that a larger hidden dimension allows Trompt to capture and represent more complex patterns and relationships in the data, leading to improved performance.

\begin{table}[H]
\caption{The performance of different number of hidden dimension.}
\vskip 0.15in
\label{tab:ablation-dimension}
\begin{center}
\begin{small}
\begin{tabular}{l | c | c | c | c | c | c}
\toprule
 & \thead{8} & \thead{16} & \thead{32} & \thead{64} & \thead{128 (Default)} & \thead{256}\\
\midrule
Classification & $79.53\%$ & $80.49\%$ & $81.16\%$ & $81.62\%$ & $81.81\%$ & $81.69\%$ \\
\midrule
Regression & $72.63\%$ & $73.61\%$ & $74.22\%$ & $74.30\%$ & $74.15\%$ & $74.47\%$ \\
\bottomrule
\end{tabular}
\end{small}
\end{center}
\vskip -0.1in
\end{table}

\textbf{Ablations on the number of Trompt Cells.}

The number of Trompt Cells ($L$) has a significant impact on the model capacity of Trompt. As shown in \cref{tab:ablation-layers}, the evaluation results indicate that increasing the number of cells leads to better performance.

In particular, Trompt performs poorly when $L=1$. This can be attributed to the design of the Trompt Cell, as depicted in the first part of \cref{fig:cell}, which relies on the output from the previous cell ($\mathbf{O}_\text{prev}$) to absorb input-dependent information.

When $L=1$, the first Trompt Cell lacks the previous cell's output, resulting in feature importances that are irrelevant to the input and becoming deterministic feature importances for all samples. This degradation in performance can be observed in the evaluation results.

Therefore, it is evident that a larger number of Trompt Cells is necessary to effectively capture and leverage input-dependent information and achieve better performance in Trompt.

\begin{table}[H]
\caption{The performance of different number of Trompt Cells.}
\vskip 0.15in
\label{tab:ablation-layers}
\begin{center}
\begin{small}
\begin{tabular}{l | c | c| c | c}
\toprule
 & \thead{1} & \thead{3} & \thead{6 (default)} & \thead{12}\\
\midrule
Classification & $79.70\%$ & $81.36\%$ & $81.81\%$ & $82.10\%$ \\
\midrule
Regression & $70.47\%$ & $73.57\%$ & $74.15\%$ & $74.61\%$  \\
\bottomrule
\end{tabular}
\end{small}
\end{center}
\vskip -0.1in
\end{table}

\subsection{Architecture}
\label{sec:more-ablation-archi}

\textbf{Ablations on whether the output of previous Trompt Cell is connected to current Trompt Cell.}

The connection between the output of the previous Trompt Cell and the current Trompt Cell is crucial, as it allows for the fusion of prompt embeddings with input-related representations.
This fusion results in sample-wise feature importances, providing valuable insights into the importance of each feature.
Without this connection, the feature importances of each Trompt Cell would become deterministic and lose their variability.
As illustrated in \cref{tab:ablation-output-fusion}, connecting the output of the previous Trompt Cell yields improved performance in both regression and classification tasks.

\begin{table}[H]
\caption{The performance of whether the output of previous Trompt Cell is connected to current Trompt Cell.}
\vskip 0.15in
\label{tab:ablation-output-fusion}
\begin{center}
\begin{small}
\begin{tabular}{l | c | c}
\toprule
 & \thead{True (default)} & \thead{False}\\
\midrule
Classification & $81.81\%$ & $81.68\%$ \\
\midrule
Regression & $74.15\%$ & $73.82\%$ \\
\bottomrule
\end{tabular}
\end{small}
\end{center}
\vskip -0.1in
\end{table}

\textbf{Ablations on whether column embeddings are input independent.}

When constructing column embeddings, we deliberately design them to be independent of the input and to capture the intrinsic properties of the tabular dataset through end-to-end training.
In this particular experiment, we examined the impact of sharing the column embeddings ($\mathbf{E}_\text{prompt}$) and input embeddings ($\mathbf{E}_\text{feature}$), which compromises the input-independent nature of column embeddings.
The results in \cref{tab:ablation-column-input} demonstrate that maintaining input-independent column embeddings leads to improved performance in both regression and classification tasks.

\begin{table}[H]
\caption{The performance of whether column embeddings are input independent.}
\vskip 0.15in
\label{tab:ablation-column-input}
\begin{center}
\begin{small}
\begin{tabular}{l | c | c}
\toprule
 & \thead{True} & \thead{False (default)}\\
\midrule
Classification & $81.66\%$ & $81.81\%$ \\
\midrule
Regression & $74.03\%$ & $74.15\%$ \\
\bottomrule
\end{tabular}
\end{small}
\end{center}
\vskip -0.1in
\end{table}

\section{More Interpretability Experiments}
In the main paper, we presented the average of $\mathbf{\hat{M}}_\text{importance}$ for each Trompt Cell.
In \cref{sec:attn-appendix}, we provide the individual $\mathbf{\hat{M}}_\text{importance}$ values for each Trompt Cell.
Furthermore, in \cref{sec:add-real-data}, we offer additional results on real-world datasets.

\subsection{Feature Importances of Each Layer}
\label{sec:attn-appendix}
\begin{figure}[H]
    \centering
    \begin{subfigure}[t]{.34\columnwidth}
        \centering
        \includegraphics[width=.95\linewidth]{figures/syn2-real.png}
        \caption{Important Features.}
        \label{fig:syn2-real-layer}
    \end{subfigure}%
    \begin{subfigure}[t]{.66\columnwidth}
        \centering
        \includegraphics[width=.95\linewidth]{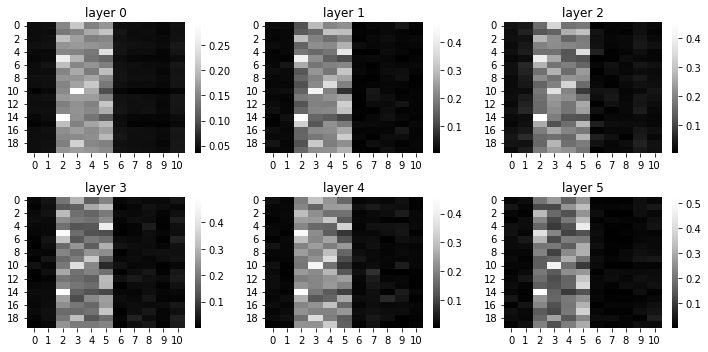}
        \caption{Masks of Trompt.}
        \label{fig:syn2-mask-layer}
    \end{subfigure}
    \caption{Attention masks of each layer on Syn2 dataset.}
    \label{fig:attn-syn2-layer}
\end{figure}

\begin{figure}[H]
    \centering
    \begin{subfigure}[t]{.34\columnwidth}
        \centering
        \includegraphics[width=.95\linewidth]{figures/syn4-real.png}
        \caption{Important Features.}
        \label{fig:syn4-real-layer}
    \end{subfigure}%
    \begin{subfigure}[t]{.66\columnwidth}
        \centering
        \includegraphics[width=.95\linewidth]{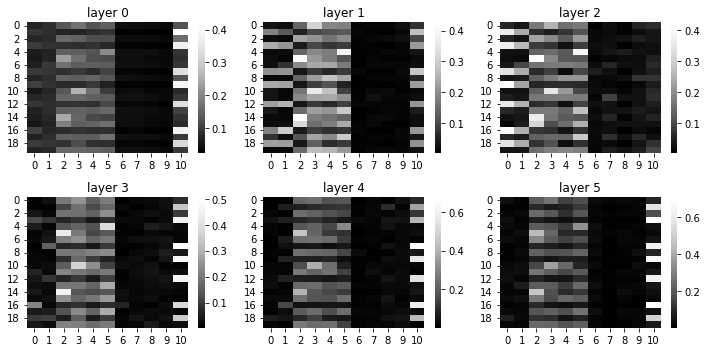}
        \caption{Masks of Trompt.}
        \label{fig:syn4-mask-layer}
    \end{subfigure}
    \caption{Attention masks of each layer on Syn4 dataset.}
    \label{fig:attn-syn4-layer}
\end{figure}

As evident from the attention visualization in \cref{fig:attn-syn2-layer,fig:attn-syn4-layer}, Trompt effectively directs its attention towards important features in both the Syn2 and Syn4 datasets.
It is worth noting that in our experiments, we employed default hyperparameters, as outlined in \cref{tab:params}, resulting in Trompt being composed of six Trompt Cells.

\subsection{Additional Real-world Datasets}
\label{sec:add-real-data}
The additional interpretability experiments were conducted on the red wine quality dataset and white wine quality dataset \cite{cortez2009modeling}.
According to the descriptions of dataset, feature selections are required since there are noisy columns in both datasets.
The experimental results are presented in \cref{tab:red-wine,tab:white-wine}.
The results indicate that both Trompt and tree-based models yielded comparable feature importances.
Specifically, Trompt assigned higher scores to the alcohol and sulphates columns in the red wine quality dataset, and the volatile acidity column in the white wine quality dataset.

\begin{table}[H]
\caption{The top-3 importance score ratio on the red wine quality dataset.}
\vskip 0.15in
\label{tab:red-wine}
\begin{center}
\begin{small}
\begin{tabular}{l | c | c | c}
\toprule
 & \thead{1st} & \thead{2nd} & \thead{3rd} \\
\midrule
RandomForest & alcohol ($27.17\%$) & sulphates ($15.44\%$) & volatile acidity ($10.92\%$) \\
\midrule
XGBoost & alcohol ($35.42\%$) & sulphates ($15.44\%$) & volatile acidity ($7.56\%$) \\
\midrule
LightGBM & alcohol ($26.08\%$) & sulphates ($15.75\%$) & volatile acidity ($10.63\%$) \\
\midrule
CatBoost & sulphates ($16.29\%$) & alcohol ($15.67\%$) & volatile acidity ($10.40\%$) \\
\midrule
GradientBoostingTree & alcohol ($26.27\%$) & sulphates ($16.24\%$) & volatile acidity ($11.12\%$) \\
\midrule
Trompt (ours) & alcohol ($11.83\%$) & sulphates ($10.94\%$) & total sulfur dioxide ($9.78\%$) \\
\bottomrule
\end{tabular}
\end{small}
\end{center}
\vskip -0.1in
\end{table}

\begin{table}[H]
\caption{The top-3 importance score ratio on the white wine quality dataset.}
\vskip 0.15in
\label{tab:white-wine}
\begin{center}
\begin{small}
\begin{tabular}{l | c | c | c}
\toprule
 & \thead{1st} & \thead{2nd} & \thead{3rd} \\
\midrule
RandomForest & alcohol ($24.22\%$) & volatile acidity ($12.44\%$) & free sulfur dioxide ($11.78\%$) \\
\midrule
XGBoost & alcohol ($31.87\%$) & free sulfur dioxide ($11.38\%$) & volatile acidity ($10.05\%$) \\
\midrule
LightGBM & alcohol ($24.02\%$) & volatile acidity ($12.47\%$) & free sulfur dioxide ($11.45\%$) \\
\midrule
CatBoost & alcohol ($17.34\%$) & volatile acidity ($12.07\%$) & free sulfur dioxide ($11.47\%$) \\
\midrule
GradientBoostingTree & alcohol ($27.84\%$) & volatile acidity ($13.59\%$) & free sulfur dioxide ($12.87\%$) \\
\midrule
Trompt (ours) & fixed acidity ($10.91\%$) & volatile acidity ($10.47\%$) & pH ($10.37\%$) \\
\bottomrule
\end{tabular}
\end{small}
\end{center}
\vskip -0.1in
\end{table}

\section{Hyperparameter Search Spaces} 
\label{sec:hyper}

The hyperparameter search space of all models is defined in \cref{tab:texpert-params,tab:ft-params,tab:resnet-params,tab:mlp-params,tab:saint-params,tab:catboost-params,tab:lightgbm-params,tab:xgboost-params,tab:forest-params,tab:gbt-params}.
We use the same search spaces for the models tested in Grinsztajn45 and additionally define the search spaces for CatBoost, LightGBM, and Trompt since they are newly added.
For CatBoost, we followed the search spaces declared by FT-Transformer \cite{gorishniy2021revisiting}.
For LightGBM, we followed the search spaces suggested by practitioners \cite{kaggle2019lightgbm,neptune.ai2022lightgbm}.

Notice that for the hyperparameter search space of Trompt, we focus on the variation of deriving feature importances (part one of \cref{fig:cell}).
In the default design, we apply concatenation on $\mathbf{SE}_\text{prompt}$ and $\mathbf{O}_\text{prev}$.
Here, we explore the possibility of summation.
Additionally, if we applied summation, the following dense layer is not necessary.
Here, we explore the possibility of removing the dense layer.
As for dense, we explore the variation of sharing weight among all prompts.
Lastly, removing residual connections of Equation \cref{eq:prompt-concat} is also explored.
Besides the variation of deriving feature importances, we also explore removing the residual connection of expanding feature embeddings (part three of \cref{fig:cell}).
In addition, we adjust the minimal batch ratio so that Trompt can be trained using different batch sizes.

To clarify, since the dense layer must be applied if concatenation was applied, and sharing dense must be false if the dense layer was not applied, the effective parameter combinations of \cref{tab:texpert-params} amount to 40.

\begin{table}[H]
\caption{Hyperparameter space of Trompt.}
\vskip 0.15in
\label{tab:texpert-params}
\begin{center}
\begin{small}
\begin{tabular}{ll}
\toprule
\thead{Parameter} & \thead{Distribution}\\
\midrule
Feature Importances Type & $[\mathtt{concat},\mathtt{add}]$\\
Feature Importances Dense & $[\mathtt{true},\mathtt{false}]$\\
Feature Importances Residual Connection & $[\mathtt{true},\mathtt{false}]$\\
Feature Importances Sharing Dense & $[\mathtt{true},\mathtt{false}]$\\
Feature Embeddings Residual Connection & $[\mathtt{true},\mathtt{false}]$\\
Minimal Batch Ratio & $[0.1,0.01]$\\
\bottomrule
\end{tabular}
\end{small}
\end{center}
\vskip -0.1in
\end{table}

\begin{table}[H]
\caption{Hyperparameter space of FT-Transformer.}
\label{tab:ft-params}
\vskip 0.15in
\begin{center}
\begin{small}
\begin{tabular}{ll}
\toprule
\thead{Parameter} & \thead{Distribution}\\
\midrule
Num Layers & $\mathtt{uniform\_int}[1,6]$\\
Feature Embedding Size & $\mathtt{uniform\_int}[64,512]$\\
Residual Dropout  & $\mathtt{uniform}[0,0.5]$\\
Attention Dropout & $\mathtt{uniform}[0,0.5]$\\
FFN Dropout & $\mathtt{uniform}[0,0.5]$\\
FFN Factor & $\mathtt{uniform}[2/3,8/3]$\\
Learning Rate & $\mathtt{log\_uniform}[1e-5,1e-3]$\\
Weight Decay & $\mathtt{log\_uniform}[1e-6,1e-3]$\\
KV Compression & $[\mathtt{true},\mathtt{false}]$\\
LKV Compression Sharing & $[\mathtt{headwise},\mathtt{key\_value}]$\\
Learning Rate Scheduler & $[\mathtt{true},\mathtt{false}]$\\
Batch Size & $[256, 512, 1024]$\\
\bottomrule
\end{tabular}
\end{small}
\end{center}
\vskip -0.1in
\end{table}

\begin{table}[H]
\caption{Hyperparameter space of ResNet.}
\label{tab:resnet-params}
\vskip 0.15in
\begin{center}
\begin{small}
\begin{tabular}{ll}
\toprule
\thead{Parameter} & \thead{Distribution}\\
\midrule
Num Layers & $\mathtt{uniform\_int}[1,16]$\\
Layers Size & $\mathtt{uniform\_int}[64,1024]$\\
Hidden Factor & $\mathtt{uniform}[1,4]$\\
Hidden Dropout & $[0,0.5]$\\
Residual Dropout & $\mathtt{uniform}[0,0.5]$\\
Learning Rate & $\mathtt{log\_uniform}[1e-5,1e-2]$\\
Weight Decay & $\mathtt{log\_uniform}[1e-8,1e-3]$\\
Category Embedding Size & $\mathtt{uniform\_int}[64,512]$\\
Normalization & $[\mathtt{batch\_norm},\mathtt{layer\_norm}]$\\
Learning Rate Scheduler & $[\mathtt{true},\mathtt{false}]$\\
Batch Size & $[256,512,1024]$\\
\bottomrule
\end{tabular}
\end{small}
\end{center}
\vskip -0.1in
\end{table}

\begin{table}[H]
\caption{Hyperparameter space of MLP.}
\label{tab:mlp-params}
\vskip 0.15in
\begin{center}
\begin{small}
\begin{tabular}{ll}
\toprule
\thead{Parameter} & \thead{Distribution}\\
\midrule
Num Layers & $\mathtt{uniform\_int}[1,8]$\\
Layer Size & $\mathtt{uniform\_int}[16,1024]$\\
Dropout & $[0,0.5]$\\
Learning Rate & $\mathtt{log\_uniform}[1e-5,1e-2]$\\
Category Embedding Size & $\mathtt{uniform\_int}[64,512]$\\
Learning Rate Scheduler & $[\mathtt{true},\mathtt{false}]$\\
Batch Size & $[256,512,1024]$\\
\bottomrule
\end{tabular}
\end{small}
\end{center}
\vskip -0.1in
\end{table}

\begin{table}[H]
\caption{Hyperparameter space of SAINT.}
\label{tab:saint-params}
\vskip 0.15in
\begin{center}
\begin{small}
\begin{tabular}{ll}
\toprule
\thead{Parameter} & \thead{Distribution}\\
\midrule
Num Layers & $\mathtt{uniform\_int}[1,2,3,6,12]$\\
Num Heads & $[2,4,8]$\\
Layer Size & $\mathtt{uniform\_int}[32,64,128]$\\
Dropout & $[0,0.1,0.2,0.3,0.4,0.5,0.6,0.7,0.8]$\\
Learning Rate & $\mathtt{log\_uniform}[1e-5,1e-3]$\\
Batch Size & $[128,256]$\\
\bottomrule
\end{tabular}
\end{small}
\end{center}
\vskip -0.1in
\end{table}

\begin{table}[H]
\caption{Hyperparameter space of CatBoost.}
\label{tab:catboost-params}
\vskip 0.15in
\begin{center}
\begin{small}
\begin{tabular}{ll}
\toprule
\thead{Parameter} & \thead{Distribution}\\
\midrule
Max Depth & $[3,4,5,6,7,8,9,10]$\\
Learning Rate & $\mathtt{log\_uniform}[1e-5,1]$\\
Iterations & $\mathtt{quantile\_uniform}[100,6000]$\\
Bagging Temperature & $\mathtt{uniform}[0,1]$\\
L2 Leaf Reg & $\mathtt{log\_uniform}[1,10]$\\
Leaf Estimation Iteration & $[1,2,3,4,5,6,7,8,9,10]$\\
\bottomrule
\end{tabular}
\end{small}
\end{center}
\vskip -0.1in
\end{table}

\begin{table}[H]
\caption{Hyperparameter space of LightGBM.}
\label{tab:lightgbm-params}
\vskip 0.15in
\begin{center}
\begin{small}
\begin{tabular}{ll}
\toprule
\thead{Parameter} & \thead{Distribution}\\
\midrule
Learning Rate & $\mathtt{uniform}[0.001,1]$\\
Max Depth & $[1,2,3,4,5,6,7,8,9,10,11]$\\
Bagging Fraction & $\mathtt{uniform}[0.1,1.0]$\\
Bagging Frequency & $[1,2,3,4,5]$\\
Num Leaves & $\mathtt{quantile\_uniform}[30,150]$\\
Feature Fraction & $\mathtt{uniform}[0.1,1.0]$\\
Num Estimators & $1000$\\
Boosting & $[\mathtt{gbdt},\mathtt{rf},\mathtt{dart}]$\\
\bottomrule
\end{tabular}
\end{small}
\end{center}
\vskip -0.1in
\end{table}

\begin{table}[H]
\caption{Hyperparameter space of XGBoost.}
\label{tab:xgboost-params}
\vskip 0.15in
\begin{center}
\begin{small}
\begin{tabular}{ll}
\toprule
\thead{Parameter} & \thead{Distribution}\\
\midrule
Max Depth & $\mathtt{uniform\_int}[1,11]$\\
Num Estimators & $1000$\\
Min Child Weight & $\mathtt{log\_uniform\_int}[1,1e2]$\\
Subsample & $\mathtt{unifrom}[0.5,1]$\\
Learning Rate & $\mathtt{log\_unifrom}[1e-5,0.7]$\\
Col Sample by Level & $\mathtt{uniform}[0.5,1]$\\
Col Sample by Tree & $\mathtt{uniform}[0.5,1]$\\
Gamma & $\mathtt{log\_uniform}[1e-8,7]$\\
Lambda & $\mathtt{log\_uniform}[1,4]$\\
Alpha & $\mathtt{log\_uniform}[1e-8,1e2]$\\
\bottomrule
\end{tabular}
\end{small}
\end{center}
\vskip -0.1in
\end{table}

\begin{table}[H]
\caption{Hyperparameter space of RandomForest.}
\label{tab:forest-params}
\vskip 0.15in
\begin{center}
\begin{small}
\begin{tabular}{ll}
\toprule
\thead{Parameter} & \thead{Distribution}\\
\midrule
Max Depth & $[\mathtt{none},2,3,4]([0.7,0.1,0.1,0.1])$\\
Num Estimators & $250$\\
Criterion & $[\mathtt{gini},\mathtt{entropy}]([\mathtt{squared\_error},\mathtt{absolute\_error}])$\\
Max Features & $[\mathtt{sqrt},\mathtt{log2},\mathtt{none},0.1,0.2,0.3,0.4,0.5,0.6,0.7,0.8,0.9]$\\
Min Samples Split & $[2,3]([0.95,0.05])$\\
Min Samples Leaf & $\mathtt{log\_uniform\_int}[1.5,50.5]$\\
Bootstrap & $[\mathtt{true},\mathtt{false}]$\\
Min Impurity Decrease & $[0.0,0.01,0.02,0.05]([0.85,0.05,0.05,0.05])$\\
\bottomrule
\end{tabular}
\end{small}
\end{center}
\vskip -0.1in
\end{table}

\begin{table}[H]
\caption{Hyperparameter space of GradientBoostingTree.}
\label{tab:gbt-params}
\vskip 0.15in
\begin{center}
\begin{small}
\begin{tabular}{ll}
\toprule
\thead{Parameter} & \thead{Distribution}\\
\midrule
Loss & $[\mathtt{deviance},\mathtt{exponential}](classif)([\mathtt{squared\_error},\mathtt{absolute\_error},\mathtt{huber}])(regression)$\\
Learning Rate & $\mathtt{log\_normal}[\mathtt{log}(0.01),\mathtt{log}(10)]$\\
Subsample & $\mathtt{uniform}[0.5,1]$\\
Num Estimators & $1000$\\
Criterion & $[\mathtt{friedman\_mse},\mathtt{squared\_error}]$\\
Max Depth & $[\mathtt{none},2,3,4,5]([0.1,0.1,0.5,0.1,0.1])$\\
Min Samples Split & $[2.3]([0.95,0.05])$\\
Min Impurity Decrease & $[0.0,0.01,0.02,0.05]([0.85,0.05])$\\
Max Leaf Nodes & $[\mathtt{none},5,10,15]([0.85,0.5])$\\
\bottomrule
\end{tabular}
\end{small}
\end{center}
\vskip -0.1in
\end{table}

\begin{table}[H]
\caption{Hyperparameter space of HistGradientBoosting.}
\label{tab:hgbt-params}
\vskip 0.15in
\begin{center}
\begin{small}
\begin{tabular}{ll}
\toprule
\thead{Parameter} & \thead{Distribution}\\
\midrule
Loss & $[\mathtt{squared\_error},\mathtt{absolute\_error},\mathtt{huber}](regression)$\\
Learning Rate & $\mathtt{log\_normal}[\mathtt{log}(0.01),\mathtt{log}(10)]$\\
Max Iteration & $1000$\\
Min Depth & $[\mathtt{none},2,3,4]$\\
Min Samples Leaf & $\mathtt{normal\_int}[20,2]$\\
Max Leaf Nodes & $\mathtt{normal\_int}[31,5]$\\
\bottomrule
\end{tabular}
\end{small}
\end{center}
\vskip -0.1in
\end{table}


\end{document}